# A One-Sided Classification Toolkit with Applications in the Analysis of Spectroscopy Data

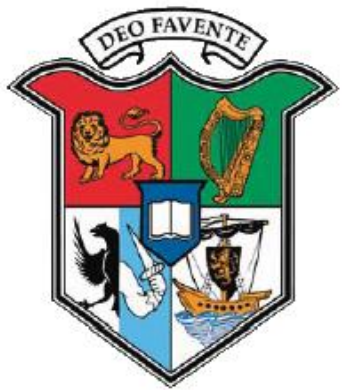

Frank G. Glavin

College of Engineering and Informatics,

National University of Ireland, Galway

A thesis submitted in partial fulfilment of
the requirements for the degree of

*Master of Science in
Applied Computing and Information Technology*

December 2009

Research Supervisor: Dr. Michael G. Madden
Research Director: Professor Gerard J. Lyons

# Table of Contents

# List of Figures

# List of Tables

# Dedication

I would like to dedicate this thesis to the memory of a great friend,

*Keith O'Reilly*,

who sadly passed away during the summer (2009).

# Acknowledgements

First and foremost, I would like to thank my supervisor, Dr. Michael Madden, for his constant support, quality feedback and excellent guidance throughout my research. I cannot emphasise enough how truly grateful I am to him for everything he has done.

I would like to thank Dr. Abdenour Bonsiar, with whom I shared an office with for the most part of this research, for all of the helpful discussions and useful recommendations that he made which greatly improved my work.

I am also grateful to Shehroz Khan for helping me to settle in when I first arrived by having some useful discussions.

I would like to thank Cormac Duffy for taking the time out of his busy schedule to help me with the lengthy process of proof-reading. I really appreciate it!

I'd like to thank Daniel, Colin, Deirdre, Cathal, Bob, Rory and Duff for being there when times were tough and things weren't going so well.

The financial support of Enterprise Ireland, under Project CFTD/05/222a, is gratefully acknowledged.

Last, but certainly not least, I would like to thank my family. My brother, John, and sister, Paula, have always been encouraging and supportive in everything that I do. I don't get enough opportunities to thank my mother, Ann, for everything that she has done for me over the years. She is certainly the biggest influence in my life and the reason that I am where I am today.

**Thanks for everything!!**

# Abstract

# A One-Sided Classification Toolkit with Applications in the Analysis of Spectroscopy Data

Frank Glavin

College of Engineering and Informatics,

National University of Ireland, Galway

This dissertation investigates the use of one-sided classification algorithms in the application of separating hazardous chlorinated solvents from other materials, based on their Raman spectra. The experimentation is carried out using a new one-sided classification toolkit that was designed and developed from the ground up.

In the one-sided classification paradigm, the objective is to separate elements of the target class from all outliers. These one-sided classifiers are generally chosen, in practice, when there is a deficiency of some sort in the training examples. Sometimes outlier examples can be rare, expensive to label, or even entirely absent. However, this author would like to note that they can be equally applicable when outlier examples are plentiful but nonetheless not statistically representative of the complete outlier concept. It is this scenario that is explicitly dealt with in this research work. In these circumstances, one-sided classifiers have been found to be more robust that conventional multi-class classifiers.

The term "unexpected" outliers is introduced to represent outlier examples, encountered in the test set, that have been taken from a different distribution to the training set examples. These are examples that are a result of an inadequate representation of all possible outliers in the training set. It can often be impossible to fully characterise outlier examples given the fact that they can represent the immeasurable quantity of "*everything else*" that is not a target.

The findings from this research have shown the potential drawbacks of using conventional multi-class classification algorithms when the test data come from a completely different distribution to that of the training samples. The experiments carried out using the chlorinated spectral data were designed to mimic the effect that these "unexpected" outliers would have on both one-sided and multi-class classification algorithms in a real-world setting.

# 1. Introduction

This thesis is concerned with the development and application of a one-sided classification toolkit. The functionality of the software is demonstrated by carrying out experimentation on high dimensional spectral data. This research work also introduces the term "unexpected" outliers and describes the effect that these have on the classification performance of both one-sided and multi-class classifiers.

This section will introduce and briefly describe the main components of the work carried out. This will be followed by a description of the motivations for undertaking such research. The structure and layout of the chapters will then be outlined and this chapter will conclude with a summary list of principal contributions.

## 1.1 Overview

At a high level, there are four core aspects addressed in this body of work. These will now be briefly summarised, and will be described in full detail in later chapters. Firstly, we will look at one-sided classification. This is concerned with making classification predictions based on a single, well-characterised class that is usually referred to as the positive or target class. The objective of a one-sided classifier is to distinguish this target class from *"everything else"* that is not a target. Non-target examples are often referred to as negative examples or outliers. One-sided classifiers differ from the conventional multi-class approach in the manner in which they are trained. Training examples for outliers can often be rare or even non-existent in the one-sided approach. However, even when such examples are plentiful, but nonetheless not fully representative of the complete concept, one-sided classifiers can still be equally applicable.

Secondly, as part of this research, a toolkit for carrying out one-sided classification experiments was designed and developed using the Java programming language. This is a software package which is run from a Command Line Interface (CLI) where the user can choose one-sided classification algorithms and adjust settings to run comprehensive experiments with a variety of options. This software was built from the ground up and was not merely an extension of existing work. For this reason, a whole chapter of this thesis is dedicated to describing its design and implementation (Chapter Three).

The third aspect of this research is spectroscopic analysis. This involves the compilation and study of the chemical spectra of a substance or mixture of substances. In particular, this thesis is concerned with Raman spectroscopy which acquires the spectra using an instrument that is known as a Raman spectrometer. The resulting data obtained is high dimensional with attributes consisting of individual points on a Raman spectrum for the particular substance. The application of one-sided algorithms to the classification of Ramon spectroscopy is novel research, to the best of this author's knowledge, and has not been found elsewhere in the literature.

Fourthly, and finally, the term "unexpected" outliers will be introduced and explained in detail. These are examples that can be encountered in practice but which have not been well described by the training examples used by the classification algorithm. It is sometimes not possible to fully characterise specific concepts. For example, if a target class consists of Shakespearean literature works, then the outlier concept is made up of *everything* else that is not his work. Such a concept is far too vast to statistically represent in the training set which leads to the inevitable scenario of meeting these "unexpected" outliers in practice.

## 1.2 Motivations

The motivation behind using one-sided classification algorithms stems from the fact that there many practical problems in which it would appear that they are better suited. Such problems include ones containing open-ended concepts that are impossible to fully characterise at the time of training. This leads to the classifier being exposed to out-of-sample examples, of which one-sided classifiers can be more robust at rejecting. Villalba and Cunningham (2009) note that the motivating factor for using one-sided classifiers is based on the absence of necessary data at the time of training and not simply the prospect of providing a better performance than multi-class classifiers in general. They note that the deficiency of data, which can be faced by multi-class classifiers, includes when the negative example space is too vast, when labelling negative examples is too expensive or when negative examples have not yet been encountered in practice (unknown negative examples). The following are some of the core motivations for this research work.

### 1.2.1   Experimentation using Spectroscopy Data

Results from the literature have shown that conventional multi-class Machine Learning techniques have been successfully used in the classification of spectroscopy data in a variety of different domains. This work will be presented later in Section 2.7.2. As previously mentioned, no reported research from the literature has been found concerning the classification of Raman spectroscopy data using one-sided classification techniques. It is for this reason, and the open-ended nature of the data in question, that these novel experiments were chosen to be carried out.

### 1.2.2 Explicitly Considering the Effects of Out-Of-Sample Data

Multi-class classifiers are often deployed in real-world applications even though they may not be the ideal choice in certain circumstances. The limitations of such an approach can be seen when the training samples are not fully representative of everything that the classifier is likely to meet in practice. Such scenarios are plentiful and, by explicitly considering out-of-sample data in the experimentation, the aim is to show the performance trends that occur in both multi-class and one-sided classifiers. The experiments were set up to imitate the effect of encountering such outliers in a real world application.

## 1.3 Thesis Structure

This thesis consists of six chapters, the first of which introduces the research which was undertaken. The following is a brief outline of the remaining chapters.

Chapter Two provides a comprehensive look at the state-of-the-art in the specified research areas. In the first section, an explanation is given for the concepts of Machine Learning and data mining, followed by differentiating between the topics of multi-class and one-sided classification. Relevant one-sided classification algorithms are described and a variety of different applications from the literature are then reviewed. The section that follows provides details about existing classification software that is both freely and commercially available. The final section introduces the technique of Raman spectroscopy and proceeds to describe examples from the literature where datasets, compiled using this technique, were classified using Machine Learning methods.

Chapter Three begins by providing an overview of the software developed during this research project. This software is called OSCAIL. The specific software requirements are then listed and a description of the OSCAIL architecture is provided. This includes early and improved control flow diagrams and explanations. This is followed by descriptions of each of the core components of the software. This chapter concludes with a description of the testing process and the identification of possible improvements for the toolkit.

Chapter Four is concerned with the effects of "unexpected" outliers on the classification performance of both one-sided and multi-class classifiers. Working with closed datasets is discussed and this is followed by a definition of the term "unexpected" outliers. A graphical illustration is used to portray the concept and some practical examples are then outlined. The relevance of topics such as transfer learning and concept drift to this problem are then briefly described. A short discussion of appropriate performance measures to use in one-sided classification is then given. Finally, some experimentation with "unexpected" outliers is carried out on a hand-written digit dataset.

Chapter Five provides details of the research experimentation carried out on the spectroscopy data. A description of the problem domain is first presented, and this is followed by details of the spectral datasets used. An overview of the experiments carried out and the results found are then listed. The chapter concludes with an analysis of the experiments that were carried out.

In the final chapter, Chapter Six, all the research work carried out is summarised and the key research contributions are presented. A brief discussion of the directions for future work is then provided followed by some concluding remarks.

## 1.4 Principal Contributions

The principal contributions of this body of work are as follows:

- The design, implementation and application of a comprehensive one-sided classification toolkit written in Java.
- A thorough review of the state-of-the-art, including a broad summary of specific applications and existing software available.
- Carrying out novel experimentation and subsequent analyses of spectroscopy data by using the toolkit and determining the relative usefulness of such an approach.
- Introducing and discussing the concept of "unexpected" outliers and their effect on the classification performance of both multi-class and one-sided classifiers.
- A paper that was accepted for oral presentation in the Irish Conference on Artificial Intelligence and Cognitive Science (AICS09). The title of the paper is "Analysis of the Effect of Unexpected Outliers in the Classification of Spectroscopy Data". The paper was written by this author (Frank G. Glavin) and Dr. Michael G. Madden. It should also be noted that the paper won the "Best Paper Award" at the conference[1] and will also be published in a special AICS volume in Springer's Lecture Notes in Artificial Intelligence series. The paper can be found in Appendix B.
- An abstract that has been accepted as a poster presentation in the upcoming Pittsburgh Conference on Analytical Chemistry and Applied Spectroscopy (Pittcon) to be held in Orlando, Florida, USA in February 2010. The abstract is entitled "Qualitative Analysis of

---

[1] http://aics.ucd.ie/welcome

Chlorinated Solvents using Raman Spectroscopy and One-Sided Classification Machine Learning Techniques" and was written, once again, by this author and Dr. Madden.

# 2. Literature Review

This chapter will begin by introducing the concepts of Machine Learning and data mining. A brief discussion of each will be followed by explanations of both multi-class and one-sided classification. Some one-sided classification algorithms will then be detailed and a selection of work from this application domain will be listed and described. This will be followed by a short overview of existing software that is available for the purposes of classification. The final section will take a look at a spectroscopic technique, called Raman spectroscopy, and some of the recent work using classification methods with this technique will then be outlined.

## 2.1 Machine Learning and Data Mining

Machine Learning is a branch of Artificial Intelligence that deals with the creation of computer algorithms that aim to improve by learning automatically through experience (Mitchell, 1997, Section 1). Mitchell notes that learning, in this case, involves searching through several possible hypotheses spaces in order to identify which best fits the training examples used. He also provides a broad definition of what learning is in terms of computing:

> "*A computer program is said to learn from experience E with respect to some class of tasks T and performance measure P, if its performance at tasks in T, as measured by P, improves with experience.*" - (Mitchell, 1997, Section 1.1)

A learning problem that is well-defined should have a task that is clear and well described, some form of metric for measuring the quality of performance and a source that can provide experience through training. The training data used can be either labelled or unlabelled. If the labels are provided, this process is called *supervised learning*. Both the inputs and outputs are known, and the learning phase involves learning from examples. Some form of feedback is required in order to measure the performance of the results. If the labels are categorical it becomes a problem of *classification*. This thesis is primarily concerned with this problem. In the experimentation carried out in Chapter Five, labelled spectroscopy data are used in the training set to build the discriminating *one-sided classifiers*[2].

---

[2] Both multi-class and one-sided classification are explained in Sections 2.2 and 2.3 respectively

It is worth noting at this point that there is an underlying assumption that the training examples and testing examples are both drawn from the same distribution. This is a necessary assumption for obtaining theoretical results but it is important to highlight that this assumption does not always hold true and is often violated in practice (Mitchell, 1997, Section 1.2.2).

Data mining is concerned with discovering patterns in data. This process can be automatic but is more commonly carried out in a semi-automatic manner. It essentially involves analyzing a dataset with the aim of extracting significant and helpful patterns that will aid us in making future predictions on new, previously unseen, data. As an example of data mining, Witten and Frank (2005, Section 1.1) considered a database of customer profiles and choices. They noted that businesses aim to target customers that have questionable loyalty, and act speedily to rectify this potential problem. There are usually large amounts of customer data available so the behaviour of former customers can be analysed and distinguishing characteristics of those who remained loyal and those who did not can be "mined". This information can then be used to target those who are likely to switch to other products or services. Such people of interest could be given special priority in a direct attempt to ensure that they remain loyal, based on this customer data. Useful patterns that are unearthed by data mining can be extremely advantageous in the task of classifying new, unknown examples. As can be seen, it can also help us gain knowledge from the data, and not just be helpful in making predictions. This discovered knowledge can become valuable information, as seen in the customer loyalty example above.

## 2.2 Conventional Multi-Class Classification

The multi-class classification paradigm aims to classify unknown examples based on a series of predefined classes (or two classes in the simple case of a binary classifier). Its objective is to automate the process of learning to make correct predictions that are based on past observations. The classification algorithm generates a *discriminating* classifier which uses a classification rule to assign new examples to one of the predetermined classes. This rule is based on the positive and negative examples that are made available in the training set.

### 2.2.1  A Simple Example

As an example, let us assume that we have some vegetables that we would like to generate a discriminating classifier for. Keeping it simple, we will train a binary classifier to classify unknown examples as either potatoes or carrots. For humans, the task of distinguishing between these vegetables is trivial. It can be accomplished simply by looking at the shape and colour, or identifying the feel, smell, and taste which makes each of them unique. Unfortunately, it is not as straightforward to computationally automate this procedure. The training process iterates over a set of labelled training instances that describe a variety of potato and carrot examples and their individual characteristics. Based on this information, the classifier can make a prediction when a new, previously unseen, example is presented. The features that are used to represent each vegetable are extremely important. For instance, if only weight and density were used in our example, classification performance could be very poor as these are clearly not the best distinguishing features. It could be quite likely that a carrot and a potato would have very similar weight ranges.

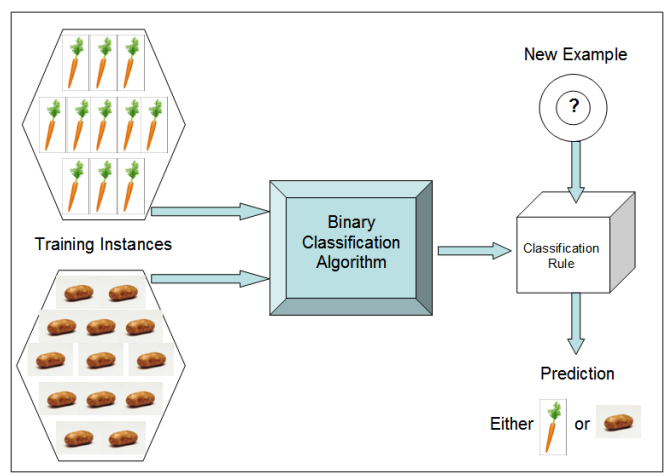

**Figure 1: Binary classification of carrots and potatoes**

The objective of the classification algorithm, in this case, is to create a classification rule that will correctly assign new unknown examples to their corresponding class; either carrot or potato.

### 2.2.2 Drawbacks of this Approach

What happens if the binary classifier, as described above, is presented with a turnip as the test example? The turnip will be incorrectly assigned to either the carrot or the potato class. The classifier is confined to assigning all the new test examples to one of the predefined classes that were used for training. Monroe and Madden (2005) point out the two key assumptions made in multi-class classification. The first is what is known as the *closed set assumption* which states that all possible cases must fall into one of the specified classes. The second assumption is that of a *good distribution* of the data. The training set is expected to contain good representative examples of all the specified classes.

For a large amount of real world practical problems, counter-examples (the direct opposite of a particular class) used for training can be extremely rare, not well represented statistically or

- 8 -

completely absent altogether (Villalba and Cunningham, 2007). Some problems of this nature, as outlined in the OSCAIL project proposal[3], include industrial process control, analysis of chemical spectra, and the classification of textual documents. The task of industrial process inspection will generally have a wealth of data available to describe each process working correctly. However, it is not feasible to expect to have representative counter-examples containing the large selection of behaviours that would indicate that the process is running incorrectly. Identifying illicit materials from chemical spectra consists of creating large comprehensive databases of various mixtures, which are likely to be encountered in practice, for use in the training set. These would include a variety of different narcotic substances, as well as various different cutting agents that could be used. The large size of such a database and its relative incompleteness (it is impossible to represent all mixtures) indicates a drawback of such a method using multi-class classification. Sometimes negative examples can exist but the nature of their distribution makes them impossible to characterise. An example of such a situation, also described in the OSCAIL proposal, would be attempting to classify textual documents to their respective authors. Training examples for each specific author's work would be available but how do we gather examples that are typical of the work *not* belonging to a specific author? When one is met with the obvious drawbacks of such examples, one-sided classification can be analysed as a possible alternative. One of the main objectives of this thesis is to investigate how robust and capable the performance of one-sided classification algorithms can be in such challenging circumstances.

## 2.3  One-Sided Classification

One-sided classification (OSC), also commonly referred to as single-class or one-class classification, differs in one vital aspect to multi-class classification in that it is only concerned with a single, well-characterised class that is known as the target or positive class. Objects of this class are distinguished from all others, referred to as outliers. Outliers consist of *all* other objects that are not targets. In one-sided classification, training data for the outliers may be rare, entirely unavailable, or statistically unrepresentative. Tax (2001) notes that the problem of one-sided classification is more difficult than that of multi-class classification. The decision boundary in the multi-class approach has the benefit of being well described from both sides with appropriate examples from each class being available, whereas the one-sided approach can only support one side of the decision boundary fully, in the absence of a comprehensive set of counter-examples. To accurately fit a separating boundary around a single class, without counter-examples, is certainly a more difficult task than if there were some available (Tax, 2001).

---

[3] The project proposal for OSCAIL was developed by Dr. M. Madden and Prof. P. Cunningham and a summary overview of it can be found here: http://datamining.it.nuigalway.ie/content/view/17/29/

## 2.4 One-Sided Classification Algorithms

In recent years, several well-known algorithms have been extended to work with the one-sided paradigm. Various authors have proposed their own unique methods for implementing this extension. The most relevant algorithms to this thesis will now be discussed in detail, followed by brief explanations of some other one-sided approaches.

### 2.4.1 One-Sided Support Vector Machine

In order to put the One-Sided Support Vector Machine algorithm approaches into perspective, it is important to first briefly describe how the original Support Vector Machine (SVM) (Vapnik, 1995; Vapnik, 1999) paradigm operates.

**2.4.1.1 Original Support Vector Machine**

There are two separate cases, one in which the classification data are linearly separable and the other in which the data can not presently be linearly separated. In the linear case, the best separating hyper-plane (maximum margin function) of the training samples that linearly separates the two classes is found.

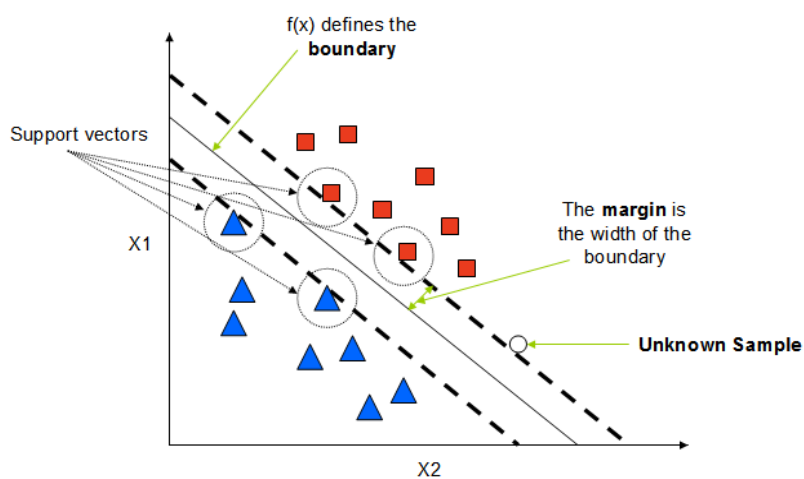

**Figure 2: An example of a linear Support Vector Machine[4].**

---

[4] Based on lecture notes written by T. Howley & M. Madden, 2005

The widest possible separating margin will lead to maximal generalisation. The data vectors on each side of the hyper-plane are called the *support vectors*. The hyper-plane, used to classify new *unknown samples*, is a quadratic programming optimisation problem as described, for example, by Russell and Norvig (2003, Section 20.6). In the case where the data cannot be completely linearly separated due to a small number of examples, slack variables can be introduced. When we deal with completely non-linear data, as illustrated in Figure 3 below, we must transform the data into a higher dimensional space in order to achieve the linear separation.

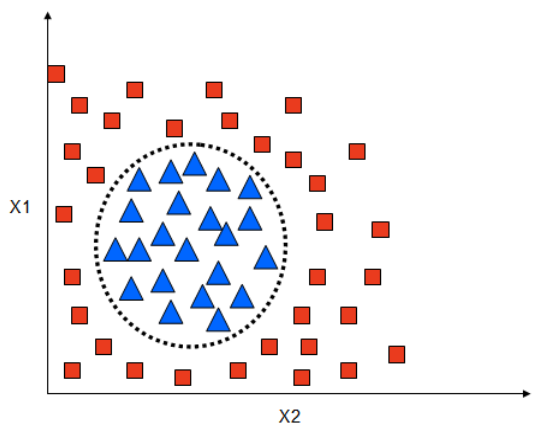

**Figure 3: Non-linearly separable data**
**(Based on Figure 20.27 from Russell and Norvig, 2003)**

This transformation or mapping of the data is achieved by using some function ϕ(*x*). Each sample is mapped to the new feature space and then the separating hyperplane can be computed.

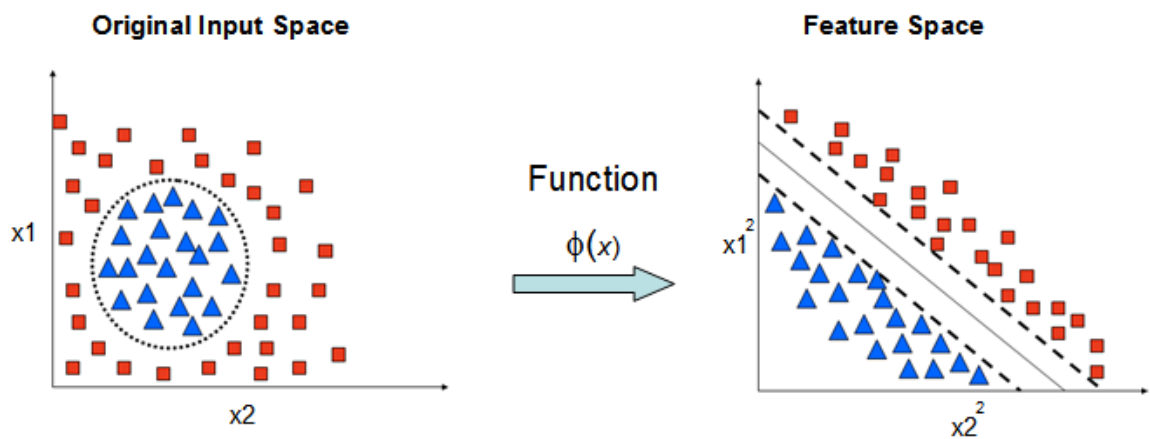

**Figure 4: Mapping data to a higher dimensional feature space**
**(Based on Figures 20.27 and 20.28 from Russell and Norvig, 2003)**

This process can be computationally expensive but it is possible to carry it out in an efficient manner when using what is known as the *Kernel Trick*. This method uses a kernel function to calculate the dot product in the new feature space as opposed to mapping each individual sample to a new set of features and then calculating the dot product. The separating hyper-plane in the new feature space can then be calculated without explicitly mapping the data (Schölkopf, 2001). Two keys aspects of SVMs, as described by Burges (1998), are generalisation theory, to aid with the choice of hypothesis, and kernel functions to achieve non-linearity in the hypothesis space without the requirement of having a non-linearity algorithm.

**2.4.1.2 One-Sided Approaches to Support Vector Machines**

In the last decade, there have been several advances in the literature that extend the Support Vector Machine algorithm to work with one-sided data. Tax and Duin (1999; 1999; 2004) propose a method that they call Support Vector Data Description (SVDD) to solve the multi-dimensional outlier detection problem. Their method involves generating a hyper-sphere boundary, as opposed to a hyper-plane, around a dataset while keeping a minimum radius. The authors note that a fraction of the training objects can be rejected if doing so significantly reduces the volume of the sphere. SVDD is made more flexible by using kernel functions and the description can be tightened when there are some negative examples for use in the training set. Their experimental results found that SVDD performed either comparable to, or better than, other outlier-detection methods. The alternative methods that were used for comparison were Parzen density estimation, Normal density estimation, and a Nearest Neighbour approach. Using a Gaussian kernel appeared the most promising whereas their results showed that a polynomial kernel had suffered from the influence of the norms of the object vectors. The necessity of requiring a large dataset to ensure a good performance was also discussed as a shortcoming of the SVDD approach.

Tax and Duin (2002) also proposed a method for generating artificial outliers distributed in and around the hyper-sphere in order to prevent both over-fitting and under-fitting. These outliers are distributed using a *d*-dimensional Gaussian distribution. The authors stated that the object's direction remains constant from the origin but the norm of the object vectors is rescaled. It was found that for very large dimensionalities this method was not feasible. Experimentation conducted showed that promising results could be achieved up to 30 dimensions. It is also emphasised in their work that there is no guarantee of good performance on "real" outliers when the target class is scattered over the complete feature space. It is only when there were available outlier examples that this could be checked.

Schölkopf *et al.* (2000; 2001) propose an approach where an attempt is made to separate the surface region that contains data from the region without any data. A hyper-plane is constructed that keeps a maximal distance from the origin with all data points lying on the positive side of the origin, as in Figure 5 below.

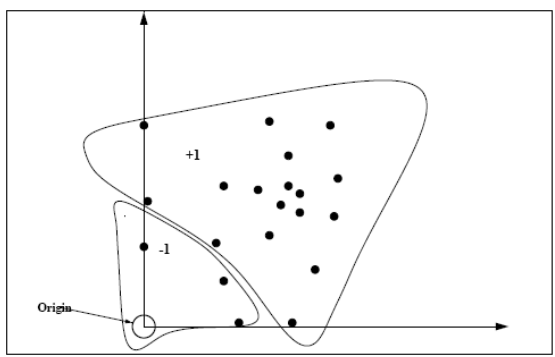

**Figure 5: One-Class SVM Classifier. Origin is only original member of second class**
**(Source: Manevitz and Yousef, 2002)**

The algorithm put forward by the authors computes a binary function returning +1 in "small" regions containing data and -1 everywhere else. The data are first mapped into feature space using a kernel and then separated from the origin with a maximum margin by solving a quadratic program (QP). The main drawback of this approach, as conceded by the authors, is the crucial role played by the origin. This origin acts as a prior for where the class abnormal instances are assumed to lie. This is what is known as the *problem of origin*. The authors tested their approach on both synthetic and real-world data. The U.S. postal service dataset of handwritten digits (Bottou, C. Cortes et al., 1994) was used and a Gaussian kernel was used in training. The experimentation results for this method looked promising but the authors noted that this was merely an indication of its true potential, brought about by using small training sets.

### 2.4.2 One-Sided k-Nearest Neighbour

It makes sense to briefly describe the original k-Nearest Neighbour (kNN) algorithm before discussing the one-sided approach.

### 2.4.2.1 Original k-Nearest Neighbour

The kNN algorithm is amongst the simplest of all Machine Learning algorithms and is an instance-based learning approach for approximating real-valued or discrete-valued target functions (See, for example, Mitchell (1997, Section 8.2) ). An assumption is made that all instances correspond to points from an *n*-dimensional space. Training data for the algorithm are simply stored and when a new instance is presented, the "nearest" or most similar instances from the neighbourhood are retrieved and used for classifying this new instance. The similarity between instances is measured using a distance metric such as Euclidean[5] distance. An unknown instance *x* is classified by assigning the most common value of the *k* neighbours in the training set that are closest to it. This is known as a *majority vote.* The amount of neighbours to take into consideration is obviously a very important parameter. We can observe from Figure 6 that when *k* = 3, the test instance *x* would be classified as a red square. However, when *k* = 9, the majority vote would classify *x* as a blue triangle.

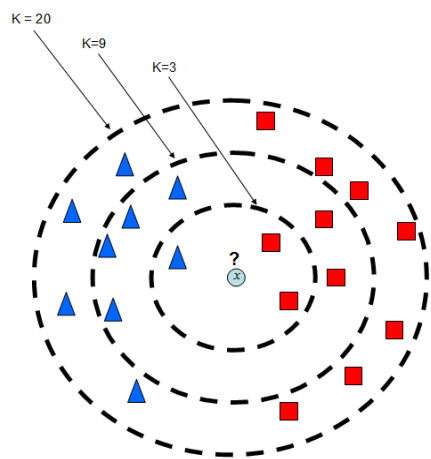

**Figure 6: Classifying a new example using the k-Nearest Neighbour algorithm[6]**

The best value for *k* will always depend on the structure of the data in question. A large value of *k* can reduce the effect that noise has on the classification but, as a result, the boundary between the classes could be made less distinctive. Various heuristic techniques, such as an internal cross validation[7] on the training data, are available to aid the process of choosing the best value of *k*. The performance of the algorithm can be severely affected by the presence of irrelevant or noisy features in the dataset. For this reason, a great deal of research effort has focussed on the selection and scaling of features to improve the performance. An introduction to the process of feature selection can be found in Guyon and Elisseeff (2003).

---

[5] The distance metrics are explained in Section A.2 of the Appendix.

[6] The figure is based on a kNN example from: http://en.wikipedia.org/wiki/File:KnnClassification.svg

[7] Cross-validation is explained in the Section A.3 of the Appendix

**2.4.2.2 One-Sided Approaches to K-Nearest Neighbour**

Datta (1997) extended the k-Nearest Neighbour algorithm to learn just a single class called the "positive" class. This modified approach, which the author calls NN-PC (Nearest Neighbour Positive Class), involves only storing training examples from a single class. In addition to these examples, the NN-PC must learn a constant δ, described by the author as being the maximum distance that a test example can be from a stored training example while it is still deemed to be from the positive class. Datta (1997, Section 5.3.2) define δ as follows:

$$\delta = Max\{\forall_x \; Min\{\forall_y \neq x \; distance(x, y)\}\}$$

Here, *x* and *y* are both examples of the positive class, and *distance(x, y)* is the distance function that is used to provide a value describing the distance between them. For example, Euclidean distance could be used for this purpose. δ is essentially measuring the variance of the training examples and then using this value as a threshold for classifying new test examples. If the distance of a test example to its nearest training example is less than the maximum distance between two training examples, then it is classified as belonging to the target class.

Tax (2001) proposed a One-Sided k-Nearest Neighbour approach, NN-*d*, where a test example is only accepted if its local density is larger than, or equal to, the local density of its *first* nearest neighbour in the training set. The distance from the test object, *te,* to its nearest neighbour in the training set, *tr,* is compared to the distance of *tr* to its nearest neighbour. The fact that the author uses *k* = 1, and uses the quotient of the densities as the distance measure makes it possible to apply several simplifications. An explicit calculation of the densities is avoided and the boundary is only approximated. This method does not have any free parameters to optimise as *k* = 1 has been determined already, and therefore it is completely reliant on the training examples. The method is scale sensitive as a direct result of the use of distances in the evaluation.

Monroe and Madden (2005) used a One-Class Nearest Neighbour algorithm in their vehicle model recognition research. The proposed method involves choosing an appropriate threshold and the amount of *k* neighbours to use. The distance from a test example *A* to the nearest training example *B* is found and this is called *D1*. Then, if *k* = 1, the distance from *B* to its nearest training example is found and this is called *D2*. Otherwise, the average distance to the nearest neighbours of *B* is found and called *D2*. If *D1* divided by *D2* is greater than the threshold value, the test example *A* is rejected as being an outlier. If it is less than the threshold than it is accepted as being part of the target class. This process is illustrated in Figure 7 below.

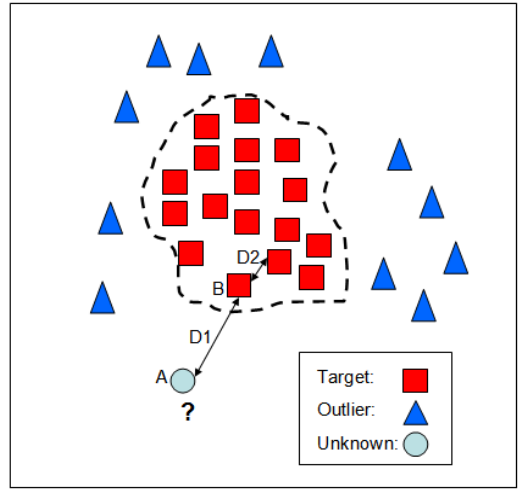

**Figure 7: An example of a One-Sided k-Nearest Neighbour algorithm
(Based on Monroe and Madden, 2005)**

In the example above (Figure 7), *A* would be rejected as an outlier as the distance from *A* to its nearest neighbour is much greater than the distance from that neighbour to its closest training neighbour.

### 2.4.3 One-Sided k-Means

The k-Means clustering algorithm (Bishop, 1995) is a simple reconstruction method in which it is assumed that the data are clustered and can be characterised by a few prototype objects. These prototypes are optimally placed by minimising the following error (Tax, 2001):

$$\varepsilon_{k-m} = a_0 + \sum_i \left( \min_k \|x_i - \mu_k\|^2 \right)$$

The distances to the prototypes of all the objects, in the k-Means algorithm, are averaged which makes the method robust to remote outliers. For one-sided classification using k-means, an amount of clusters and classification threshold must be decided upon by the user. Then, when the test examples are presented, the Euclidean distance to the nearest prototype in the training data is found and the sample is classified as either a target or an outlier based on a comparison of the distance with the specified threshold. This process is illustrated in Figure 8 below. The prototype objects consist of target training examples. The distance from the unknown object to its nearest prototype is found. If this distance is less than the classification threshold, the example is classified as a target. Otherwise it will be classified as an outlier.

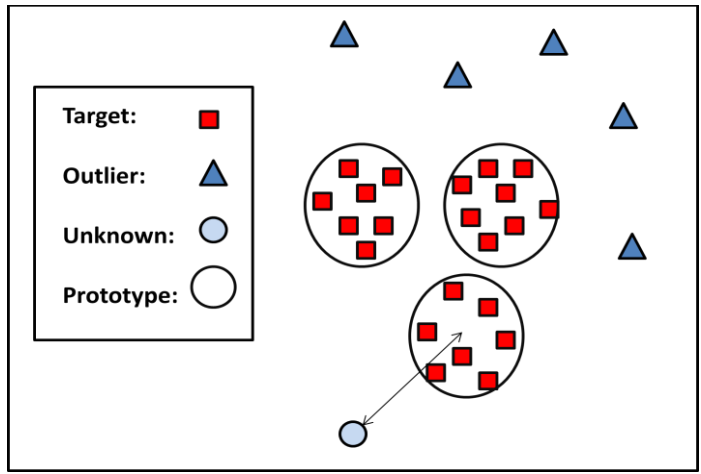

Figure 8: An Example of One-Sided Classification using k-Means

### 2.4.4 One-Sided Neural Networks

The processing power of the human brain is provided by complex networks of what are known as neurons. A *neuron* is a cell in the brain that has the functionality of collecting, processing and sending electrical signals. Some of the earliest work in Artificial Intelligence involved attempts to mimic this process by creating artificial neural networks (ANN) as noted by Russell and Norvig (2003, Section 20.5). The two main types of neural network are feed-forward (acyclic) and recurrent (cyclic). Feed-forward neural networks can be either single layer, where the inputs are directly connected to the outputs, or multilayer where there are one or more hidden layers present.

Skabar (2003) described the use of a feed-forward neural network with a single output neuron that uses only labelled positive examples and unlabelled (both positive and negative) examples in the application of predicting mineral deposits. The author notes that the only difference in the derivation of the error function in the one-sided approach is in the labelling of the examples. In the conventional two class approach, positive examples would be assigned a target value of 1 and negative examples would be assigned a target value of 0. After training, the network would represent the probability of an example belonging to the positive class and a classification threshold of 0.5 could be set. However, in the approach described by the author, examples with an assigned target value of 0 (unlabelled examples) would consist of both positives and negatives. Under the assumption that the labelled positives are representative of the positive class, the

- 17 -

author hypothesised that the networks function could still successfully discriminate between the positives and negatives. The experimental results showed that, using the neural network approach, the prediction model provided a better fit to the observed data in comparison to the density estimation based model that was used.

Another example of using a feed-forward network, only using positive examples, can be found in Manevitz and Yousef (2007). Furthermore, a review of novelty detection methods using neural networks can be found in Markou and Singh (2003).

### 2.4.5 One-Sided Naïve Bayes

The traditional Naïve Bayes algorithm (Duda and Hart, 1973) aims to calculate the probability of a certain class from an unlabelled example based on its attribute values. This can be written formally as follows:

$$p(C_i \mid A_1 = v_1 \& ... \& A_n = v_n)$$

Here, $C_i$ is a specific class, $A_1$ to $A_n$ are attributes of this class and $v_1$ to $v_n$ are values for the attributes. An assumption is made that all attributes are independent and Bayes theorem[8] can be applied to the above calculation in order to calculate the probability that a test example is from a specified class. Datta (1997) adjusted the Naïve Bayes algorithm for one-sided classification. The author remarked that when only the positive class is available, trying to calculate $p(C_i)$ becomes infeasible and this value can be assumed close to 1. Datta's approach, called NB-PC (Naïve Bayes Positive Class) used the probabilities of the attribute values in order to calculate a threshold. This threshold is calculated as follows (Datta, 1997, Section 5.3.3):

$$t = Min[A_x \prod_{j}^{attributes} p(A_j = v_i)]$$

An example consists of various different attributes, each of which can have several different values. A single attribute value for an example $x$ is described as $A_j = v_i$ and the probability of that attribute value is $p(A_j = v_i)$. The author explained that the probabilities of all the values for each of the attributes are essentially normalised by the probability of the $v$ value that occurs most frequently. Classification is carried out as follows: For a test example, if $\prod_{j}^{attributes} (A_j = v_i)$ is greater than or equal to the threshold, $t$, then the example is classified as being a member of the positive class (Datta, 1997).

---

[8] A comprehensive explanation of Bayes theorem can be found in Mitchell (1997, Section 6.2)

### 2.4.6 One-Sided Decision Trees

Conventional decision tree learning involves creating a tree-like structure in which each node of the tree corresponds to some attribute of the example. There is a single node at the top of the tree, called the "root" node that expands to lower level nodes via a conditional branch construction based on the different attribute values. An illustrated example from Mitchell (1997, Figure 3.1) shows a tree used to classify the decision of "whether or not to play tennis":

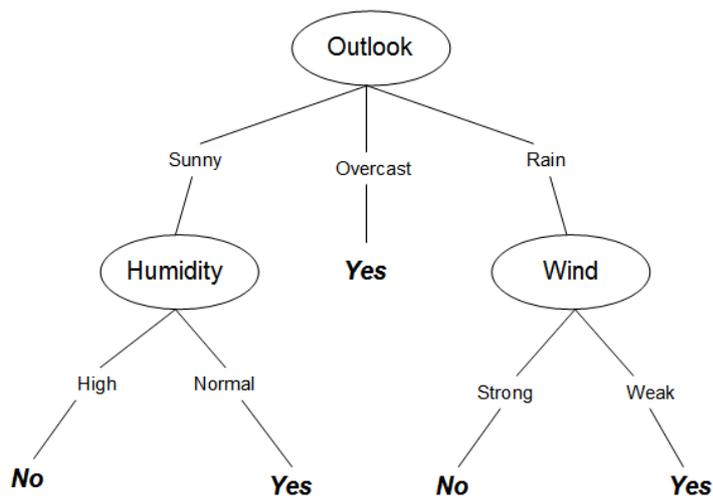

**Figure 9: A decision tree for the concept of *PlayTennis***
**Source: (Mitchell, 1997, Figure 3.1)**

Each branch protruding from a node corresponds to one of the possible values for that attribute. For instance, in Figure 9 above, the attribute Outlook can have the value of Sunny, Overcast or Rain. The algorithm works its way down the tree, dependent on the attribute values, and then arrives at a classification decision when it reaches the "leaf" node at the bottom. An algorithm developed by Quinlan (1986), called ID3, is used to generate decision trees from a training set. To begin, a subset of the training set called a "window" is chosen at random and used in the generation of the decision tree. The resulting tree is then applied to the remainder of the training set. If some have been misclassified, these are then added to the "window" and the decision tree is updated in an iterative process. Some of the issues associated with ID3 such as computational inefficiency, over-fitting and dealing with missing attributes were addressed in an updated algorithm called C4.5 (Quinlan, 1993).

An algorithm called POSC4.5, based on C4.5, was developed by Letouzey *et al*. (2000) for dealing with only positive and unlabelled examples in the training set. The algorithm takes a set of target examples, referred to as POS, and a set of unlabelled examples, referred to as UNL, as its input. The POS set is then split into two sets as follows: A new set POS$_L$ comprises two thirds of the original POS and another set POS$_T$ making up the remaining third. UNL is then split in the same fashion creating both UNL$_L$ and UNL$_T$. There are two stages. The first involves using the POS$_L$ and UNL$_L$ sets to construct the hypothesis set. The authors stated that these sets are used to "simulate the positive statistical oracle and the instance statistical oracle". The second stage uses the POS$_T$ and UNL$_T$ sets in a hypothesis testing algorithm. The authors concluded that the target concept weight is a key parameter when learning from positive and unlabelled examples. Their experiments showed that a lower bound on the target concept weight is sufficient when learning from positive and unlabelled data, though they conceded that further experiments would be required.

### 2.4.7 Nonparametric Density Estimation: Parzen Windows

One-sided classifiers can be obtained by explicitly estimating the density of a given training set and then setting a corresponding threshold on this density (Tax, 2001). Fundamental techniques for estimating an unknown probability density function are based on the fact that the probability *P* of a vector *x,* falling in a region *R,* is given by:

$$P = \int_R P(x')dx'$$

There are two general classes of distribution and density models. These are parametric and nonparametric models. The parametric model assumes that the forms of the underlying density functions are known and the function is often characterised by a *location* parameter and a *scale* parameter. Such models are simplistic but can suffer from a relatively high bias when the data does not obey the assumed functional form (Hand *et al.*, 2001). Classical parametric densities are unimodal, that is, they have a single local maximum; however many practical problems can involve multimodal densities (Duda and Hart, 1973). Nonparametric methods can be used when the density estimate is data driven and without assuming any a priori knowledge about the functional form. Parzen-window density estimation (Parzen, 1962) is an example of a non-parametric model. Each example, in this model, contributes a Gaussian density with peaks around it and then drops off quickly. The training of this model involves the determination of a single parameter, h, which is the optimal width of the kernel and which controls the drop off of the Gaussian densities. Tax (2001) notes that the computational training cost of a Parzen density estimator is minimal, however, the testing phase is very expensive. This is due to the fact that all of the training objects have to be stored with their distances to each test example being calculated and then sorted. Such a characteristic can limit how applicable this method could be when faced with large datasets in high dimensional feature space (Tax, 2001).

## 2.5 Applications of One-Sided Classification

In the following subsections, a variety of applications for one-sided classification from the literature, as applied to many different domains, will be described. The majority of this work has been carried out over the last decade, in a period of time when one-sided classification has received noticeable attention.

### 2.5.1 Document Classification and Retrieval

Document retrieval is concerned with matching documents against user queries and is generally based on relevance feedback from the user. Typically, the user types a query of keywords or a full sentence and is then presented with several web or text documents that are deemed to be relevant to the query provided. The user then provides some information concerning how relevant the results were and the system is updated accordingly.

Manevitz and Yousef (2002) implemented two different versions of a One-Class SVM for document classification, one which was based on the identification of "outlier" data as a representation of the second class, and the other which was a version proposed by Schölkopf *et al.* (2001). They carried out several experiments using a variety of different kernels and different representations of the data (tf-idf[9], binary vectors etc.). The results were then compared with several other one-class versions of well-known algorithms such as Nearest Neighbour, Naïve Bayes and a One-Class Neural Network algorithm. They used Hamming distance to decide how far a point can be from the origin before it is classified as an outlier. Their overall results showed that the One-Class SVM, proposed by Schölkopf *et al.* (2001), gave the best results. This was made very clear upon viewing the comparisons with all the contesting algorithms, with the exception of the Compression Neural Network, which was comparable. However, it should be noted that the One-Class SVM is less computationally expensive than the Compression Neural Network. It should also be noted that the authors found that the One-Class SVM was very sensitive to parameter selection and the choice of kernel. For this reason, they reported that the neural network approach was the preferred method.

---

[9] Term Frequency, Inverse Document Frequency (td-idf) is a weight used to evaluate how important a word is to a document.

Onoda *et al.* (2005) proposed a new feedback method that only uses non-relevant documents termed *non-relevance feedback document retrieval*. Their method is based on the One-Class SVM and SVDD algorithms. They observed that the initial retrieved documents, which are displayed to the user, can sometimes not contain any relevant material at all. In this case, most relevance feedback document retrieval systems would fail as they need both relevant and non-relevant documents to create a binary classification problem. Their proposed technique works in several steps. Firstly, the documents are prepared for the first feedback. They are then shown to the user and classified by the user as being either relevant or non-relevant. If none of the documents are relevant to the user they are labelled as "-1" and the process goes to the next step. However, if there are some relevant and some non-relevant present, then regular relevance feedback techniques are used. The next step, assuming that all are non-relevant documents, is to determine the area of these documents by using a discriminant function that is generated by either One-Class SVM or SVDD. Once this has been determined, the classifier can verify which results reside in the "non-relevant" area and which are in the "*not* non-relevant" area and classify them accordingly. Documents are ranked in increasing order of distance from the non-relevant area and then shown to the user. The step of classifying the documents as relevant or not is then repeated by the user. The authors' results showed consistently better performance than the well-known techniques of Vector Space Model (VSM) (Salton *et al.*, 1975) and the Rocchio method (Rocchio, 1971) when only non-relevant documents were present.

### 2.5.2 Content-Based Image Retrieval

Content-Based Image Retrieval (CBIR) has generated a lot of interest in recent years due to the large amounts of digital storage now available for images online. Ease of access to these images has been greatly improved as a result of faster internet connections.

Seo (2007) presented a One-Class Support Vector Machine based classification method for categorising a large image database by colour and texture, for CBIR. In order to achieve an accurate detection and classification of the images, features need to be extracted from the image patterns. In this paper, the author decided to use colour and texture information to represent the image features. This is a large, complex pattern-processing task that the author suggests suits the One-Class SVM paradigm. Experimentation was carried out using 1000 real-world sample images that were divided up into 14 categories. An Artificial Neural Network (ANN) was used as an alternative CBIR algorithm for performance comparison with One-Class SVM. The experiments showed that One-Class SVM out-performed ANN on the 1000 real-world sample image data. The importance of good kernel and parameter selection for One-Class SVM was stressed by the author. They also stated that this work illustrates the ability of kernel-based learning methods to be highly competitive and efficient on tasks such as CBIR.

Chen *et al.* (2001) developed a technique based on One-Class SVM that fits a tight hyper-sphere in the non-linearly transformed feature space. This hyper-sphere, based on target examples, includes most of the target images. Good generalisation is achieved from the regularisation term of the SVM. They used two variations: One for the linear case, termed *LOC-SVM,* and one for non-linear multi-mode distributions called *KOC-SVM*. Tests were carried out using synthesised and real-world data to verify the effectiveness of these approaches. They were compared to a relevance feedback technique called Whitening Transform (WT) (Fukunaga, 1990). *LOC-SVM* performed the worst due to its lack of flexibility. The accuracy of *KOC-SVM* and WT were quite similar although the former performed marginally better.

Zhang *et al.* (2005) proposed a method, based on One-Class SVM, to solve the Multiple Instance Learning (MIL) problem in single object CBIR. In MIL, the label of an individual instance is not known. Only the label of a set of instances is known. This is called the label of the bag. The MIL problem is to map an instance to its label according to information that is learned from the bag labels. The authors suggested that their method has the advantage of being based only on a semantic region of the image as opposed to the whole image. They believe this to be more reasonable as users are sometimes only interested in a certain section of the image. The experiments that they ran were based on 9,800 images and they compared their results with a neural network based MIL algorithm and a general feature re-weighting algorithm. The results of their experiments showed that their method out-performed the other two. The authors believe that the robustness of the One-Class SVM method can meet user needs in this domain more closely.

### 2.5.3 Anomaly and Outlier Detection

Anomaly and outlier detection is concerned with detecting abnormal patterns that do not conform to a normal known behaviour of a specified system. This is applicable in a variety of domains such as intrusion detection, fault detection, and fraud detection. The two main categories of this type of technique, described by Lazarevic (2004), are supervised and unsupervised learning. Supervised anomaly detection involves a classifier being trained using examples from *both* a normal and an anomaly class, and then classifying the test examples based on this information. Unsupervised anomaly detection is concerned with detecting anomalies in an unlabelled test set while assuming that the majority of the examples are normal.

Li *et al.* (2003) investigated the usefulness of using an One-Class SVM for intrusion detection. They aimed to improve upon the One-Class SVM proposed by Schölkopf *et al.* (2001) by not only assuming that the origin is the second class but to also assume that points that are close enough to the origin are outlier points. If the input data correspond to these points, these are deemed to belong to the anomaly class. Their experimentation was carried out on data from the 1999 DARPA Intrusion Detection Evaluation Program which contains approximately four gigabytes of compressed tcpdump gathered from over seven weeks of network traffic. The experiments involved differentiating between intrusions and normal activities. As a comparison, they also tested a cluster-based method, Naïve Bayes, k-Nearest Neighbour and the standard SVM. The best accuracy rate of 96% from their results was achieved by the One-Class SVM. The closest accuracy to this was the clustering algorithm with 93%. The remaining algorithms averaged at around 90% accuracy.

Ma and Perkins (2003) proposed a new algorithm for time-series novelty detection based on a One-Class SVM. The authors explained how a *projected phase space* allows One-Class SVM to be applied to time-series data. They also explained that they combined One-Class SVM outputs for different phase spaces to achieve better novelty detection results. Experiments were carried out based on both synthetic and measured data and these showed "promising" results according to the authors.

Wang (2004) introduced a new family of kernel methods to deal with intrusion detection. They combined the STIDE kernel and the Markov chain kernel with a One-Class SVM for use with anomaly detection. For their experimentation, they used a sendmail dataset from the University of New Mexico which includes hundreds of normal executions of sendmail daemons and several which have been attacked. The results showed that the combination of these kernels and the One-Class SVM provided more accurate results than that of conventional anomaly detection algorithms, that is, conventional STIDE and Markov chain methods.

Gardner *et al.* (2006) applied One-Class SVM novelty detection methods to detect epileptic seizures in humans. They proposed to improve the state-of-the-art in seizure detection by changing the problem into a time-series novelty detection problem. The authors noted the downfalls of traditional approaches to seizure detection that use binary classification. Their technique was said to overcome three main limitations that were evident in competing algorithms. The technique does not require training on seizures; patient specific tuning is not necessary; and the patient's state of consciousness does not need to be known. A simple benchmark novelty detection algorithm was developed by the authors in order to draw a direct comparison with their One-Class SVM approach. They reported that both techniques were effective in detecting seizures but the One-Class SVM method was shown to out-perform the benchmark approach, especially concerning false positive rate, in a consistent manner.

Tax *et al.* (1999) used Support Vector Data Description (SVDD) in order to distinguish normal from abnormal conditions based on measurements taken from a water pump. This application domain is called machine fault detection. The author's aim was to find the best representation of the dataset so that the target class could be best distinguished from the outlier class. They applied SVDD to a normal pump in a working station and also to a pump in an experimental setup. The importance of a pre-processing step for good classification performance was emphasised by the authors, especially that of feature selection. Feature selection involves finding the optimal features from a given set of features. Often times, simple criteria can be used as the basis for this judgement. They recorded data from several vibrating sensors on a rotating machine and then created three different subsets of the sensor channels. These experiments compared several methods of feature extraction using a relatively low feature dimensionality.

### 2.5.4  Vehicle Model Recognition

Monroe and Madden (2005) proposed the application of a One-Sided k-Nearest Neighbour (kNN) algorithm to the task of recognising the make and model of vehicles from sample images. The authors used specific feature extraction techniques from a dataset of 150 frontal view images of vehicles. The dataset was broken down into five classes, each containing 30 example images. They implemented the system using Matlab software. All the images were converted to greyscale and the top half of each image was automatically cropped off, as it was determined to contain no unique, distinguishing features. The important features of each image were found using Canny edge detection (Canny, 1986) and a fixed-length numerical vector was derived for each. Some multi-class techniques were also implemented to draw a comparison with the performance of the One-Sided kNN approach. These included the C4.5 decision tree and a feed-forward neural network. The experiment results showed that the neural network and One-Class kNN had comparable results and both out-performed the decision tree method. The authors note that a limitation of the One-Class kNN was in calculating the threshold to optimise the performance levels.

### 2.5.5 Facial Expression Analysis

Zeng *et al.* (2006) used a method of one-class classification for recognising emotional and non-emotional facial expressions in a human conversation setting that was based in an Adult Attachment Interview[10] (AAI) (George, Kaplan et al., 1985). Emotional facial expressions were defined in terms of facial action units in the psychological study of AAI, however there were no distinct descriptions available for non-emotional facial expressions. Due to the nature of facial expressions, it would be quite difficult and expensive to characterise these non-emotional expressions. For this reason, the authors treated the expression recognition problem as a one-classed problem where the target class contained the emotional facial expressions and the outliers contained all of the emotionless data. They first used Kernel Whitening (Tax and Juszczak, 2002) to map the emotional data into a kernel subspace and then applied the SVDD (Tax and Duin, 2004) algorithm. This approach was then compared to Principal Component Analysis (PCA) and SVDD as well as PCA and a single Gaussian method. The experimental results showed that Kernel W\hitening significantly improved upon the results of just SVDD alone. It was also suggested by the authors that the Gaussian method could not correctly estimate the density of the target data.

### 2.5.6 Personal Identification Verification

Bicego *et al.* (2005) proposed the use of a One-Class Support Vector Machine for personal identification verification using facial images. The experimentation conducted included 40 test subjects with 10 facial images for each. The training set comprised 5 of the images from each subject. A direct comparison was carried out between binary SVM and One-Class SVM using the same dataset and features for both tests. The results showed that the one-class approach retained the attractive features of binary SVM, such as good generalisation, while also omitting the need for an "imposter" class in the training set. In the binary method, the choice of the "imposter class" is paramount and essentially dictates how well it will perform. Eliminating the need for representing the counter-class in the training set, while maintaining equivalent performance results shows just how useful a one-class approach can be.

Brew *et al.* (2007) applied a variety of state-of-the-art one-class classification techniques to solve the problem of determining if segments of speech come from a specific individual. This is known as the speaker verification problem. The target class can be well characterised by speech examples from the user but the set of counter-examples remains completely open. The authors suggested one-class classifiers to be an appropriate solution to this task, as labelled data exist for just one of the classes in a two class problem. They carried out feature extraction on the voice samples in an effort to distinguish the target "*speaker*" class from everything else (the outlier "*imposter*" class). The algorithms used were Gaussian Mixture Model (GMM), Gaussian Model, SVDD, and One-Class kNN. The experimental results showed that the GMM and Gaussian Model best fitted their target distribution between the training and test sets. On average, the best performance that was produced came from the GMM. It was also noted by the authors that different classification algorithms performed better on different speakers.

---

[10] The AAI is a widely used and validated technique in the developmental research of identifying adult attachment representations. It is a semi-structured interview that characterises a participant's current state of mind based on past parent-child experiences.

### 2.5.7 Other Applications

The following are brief explanations of some other unique techniques and applications for one-sided classifiers.

Sanchez-Hernandez *et al.* (2007) applied a range of one-class classifiers to the task of classifying fenland [11] from Landsat Enhanced Thematic Mapper Plus [12] imagery. The experimental results showed that the use of SVDD, in particular, provided a much better classification performance than standard multi-class classification techniques from their benchmark analysis. The size of the training set was also significantly reduced due to the fact that only positive examples were used.

Skabar *et al.* (2000) defined a classifier evaluation function in which Bayesian likelihoods of necessity and sufficiency were used to measure the performance of classifiers when only labelled positive examples and unlabelled (both positive and negative) examples were present in the training set. A neural network system was implemented which made use of this evaluation function as a heuristic for guiding the search process. Experimentation was carried out which compared the Bayesian likelihoods approach with a back-propagation approach. The results showed that the Bayesian likelihoods approach achieved a comparable performance to the back-propagation method. This is noteworthy considering the back-propagation method required *both* labelled positive and labelled negative examples for training. The author believes that this classifier evaluation function could be applicable in a broad range of domains where it is difficult or not feasible to collect counter-examples.

---

[11] Fenland is a habitat of considerable conservation value.

[12] http://eros.usgs.gov/products/satellite/landsat7.php

Kruengkrai and Jaruskulchai (2003) used a One-Class SVM approach to extract relevant sentences from a provided source. Such a technique can be useful in automatically creating a summarisation of works of text. The experimentation was carried out on 6942 newspaper articles using LIBSVM (Chang and Lin, 2001). Results found that reasonable performance can be achieved when the appropriate parameters were selected but the authors also stated that this One-Class SVM had distinct advantages, especially when the training set was unbalanced.

Hardoon and Manevitz (2005) applied one-class Machine Learning algorithms to the problem of classifying whether or not a person is performing a specific task based on fMRI[13] slices of their brain. The image data being used were very high dimensional and could be extremely noisy due to the complex nature of the brain. A compression neural network and One-Class SVM were used and their performances were compared against each other. They both achieved similar results and the authors suggested that combining these techniques in future work could prove beneficial.

Cohen *et al.* (2004) attempted to detect patients in a hospital who had one or more nosocomial[14] infections by applying a One-Class SVM to clinical and survey data from the patients. The findings revealed that the One-Class SVM achieved good sensitivity (amount of sick patients correctly identified with the infections) but at the cost of poor specificity (amount of well patients identified as not having the infections). And so, experts would need to decide if such a high detection rate was worth the cost of treating the "false positive" patients.

Skabar (2003) proposed a technique for learning symbolic concept descriptions when using a dataset that consisted of only labelled positive examples as well as other unlabelled examples. The unlabelled examples comprised both positive and negative examples. The technique used an evolutionary search of the concept description space. The concept descriptions were described using the VL1 language (Michalski *et al.*, 1986). One of the features of this technique, as mentioned by the author, is that it uses an "empirical determination of a bias weighting factor" which essentially controls the balance of specialisation and generalisation of the hypotheses. This is useful when one is searching for a desired hypothesis to specifically have either a low false-negative rate or low false-positive rate. Experimentation carried out used the Australian Credit Dataset (Quinlan, 1987) and compared the proposed technique to the C4.5 algorithm

---

[13] fMRI stands for functional Magnetic Resonance Imaging and is an imaging technique used to map different sensory and cognitive functions to specific regions of the brain. (Hardoon and Manevitz, 2005)

[14] This type of infection develops during a patient's hospitalisation and was not present or developing at the time of their admission. (Cohen *et al.*, 2004)

(Quinlan, 1993). The proposed technique was shown to perform comparably even though the C4.5 algorithm used both labelled positives and labelled negatives in the training set.

Rabaoui *et al.* (2008) have developed a method for recognising environmental sounds that can be used with surveillance and security applications. Their method was based on One-Class SVMs in conjunction with what they termed "a sophisticated dissimilarity measure". They use wavelet based feature vectors that were produced from the coefficients of the wavelets. These coefficients could capture localised time and frequency information about the sound waveform better than some of their standard counterparts and were noted to provide better discrimination performance. The experimentation was carried out on 1000 sounds, each of which belongs to one of nine classes. They tested various different feature combinations and stated that their method provided consistently low error rates and good classification performance.

## 2.6 Relevant Classification Software Currently Available

There are many software packages and workbenches for carrying out both multi-class and one-sided classification tasks. Some of these are open source and freely available while others are commercially available. A few relevant examples of such software will now be detailed below.

### 2.6.1 Waikato Environment for Knowledge Analysis (WEKA)

WEKA is open-source software, distributed under the GNU[15] general public license that is essentially a collection of Machine Learning algorithms for data mining tasks. The software, which was developed at the University of Waikato, is written in Java code and contains tools for classification, regression, clustering, data pre-processing, association rules and visualisation. The main user interface of WEKA is called *Explorer* and contains several different panels to give access to the main components of the workbench.

---

[15] The GNU General Public License is a free, copyleft license for software and other kinds of works. More information can be found at: http://www.gnu.org/copyleft/gpl.html

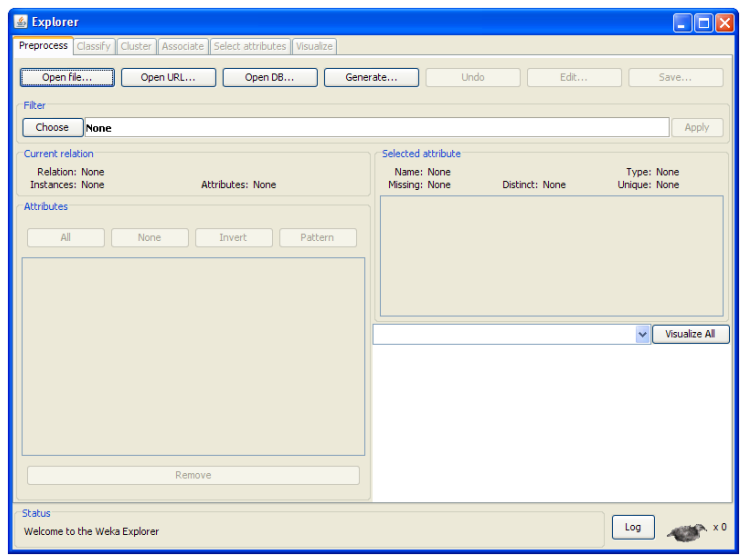

**Figure 10: WEKA Explorer graphical user interface**

The *preprocess* panel allows users to import data from CSV [16] files, databases etc. A preprocessing filter can then be applied to transform these data if required. An example of this would be normalising numerical features into specific ranges. The *classify* panel allows the user to select either a classification or regression algorithm to be applied to the loaded data which then displays the result of the predictive model. The *cluster* panel gives access to the clustering techniques that are provided by WEKA. The next panel is called *associate*. This gives access to the association rule learners that attempt to find relationships between attributes in the dataset. This panel is followed by the *select attributes* panel, which contains algorithms for the identification of the most predictable attributes in the dataset. Finally, the *visualise* panel has a display of several different scatter plot visualisations of the data that can be resized as well as having several other selection parameters. The WEKA workbench also contains three other interfaces; namely, *Experimenter*, *KnowlegeFlow* and *SimpleCLI*. The *Experimenter* interface allows the user to compare a variety of learning techniques. The experimentation process can be automated enabling users to set up large-scale statistical experiments. *KnowledgeFlow* allows users to drag boxes around the screen that represent algorithms and data sources. These can also be joined together into the configuration that is required. The final interface is a simple Command Line Interface (CLI), in which the user can type in textual commands to interact with WEKA. More information can be found from the book "Data Mining: Practical Machine Learning Tools and Techniques" (Witten and Frank, 2005).

---

[16] Comma Separated Value files are data files used for the storage of data in tabular format.

### 2.6.2 LIBSVM - A Library for Support Vector Machines

Chang and Lin (2009), the authors of LIBSVM (Chang and Lin, 2001), describe it as a library for Support Vector Machines (SVM) with a goal of enabling users to easily use SVMs as a tool. LIBSVM is integrated software for support vector classification, regression, and distribution estimation (One-Sided SVM) that supports multi-class classification. There are sources written in both C++ and Java. There are several *view* interfaces such as WEKA, Matlab, Perl, and Python.

### 2.6.3 RapidMiner

RapidMiner, previously known as Yet Another Learning Environment (YALE)(Fischer et al. 2003), is a data mining package that has been developed in Java. There are two versions of the software currently available; *Community Edition* which is open source and freely available, and *Enterprise Edition* which is a closed-source package aimed at businesses. For an insight into the various concepts and functionality provided by RapidMiner, the reader is advised to examine a paper by Mierswa *et al.* (2006) which details YALE, the predecessor to RapidMiner. A plethora of information can also be found via the RapidMiner website[17].

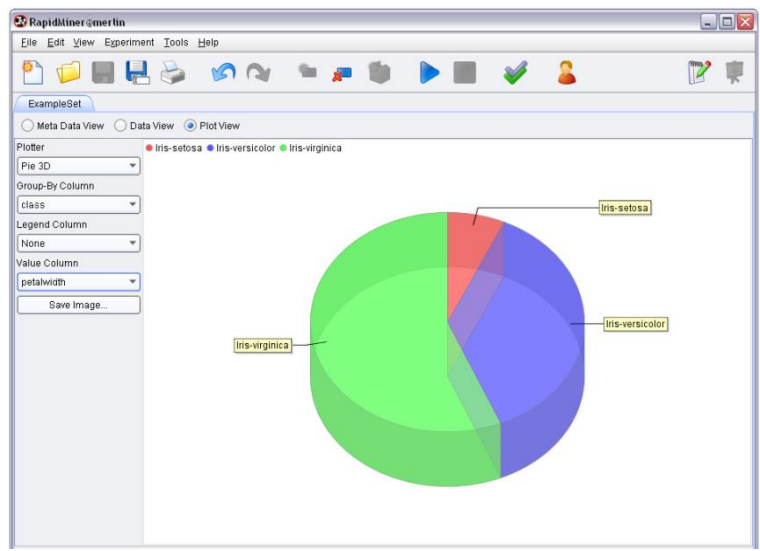

**Figure 11: RapidMiner graphical user interface[18]**

### 2.6.4 Data Description Toolbox (dd_tools)

The dd_tools Matlab toolbox is used to aid research into one-class classification (or data description) by providing the necessary tools, classifiers, and evaluation functions (Tax, 2006). It was developed by D. M. J. Tax and is an extension of PRtools[19]. The current version is 1.7.3 as of December 2009. Using the toolbox, one can define special one-class datasets and classifiers. There are also methods for generating artificial outliers, calculating error estimation, showing ROC[20] curves, area under the ROC curve error, the precision-recall curve, and mean precision[21].

---

[17] http://rapid-i.com/content/blogcategory/38/69/

[18] Source: http://rapid-i.com/content/view/9/25

## 2.7 Spectroscopy Analysis

The technique of spectroscopy is often used in both physical and analytical chemistry and involves the study of experimentally obtained spectra. These are gathered using an instrument such as a spectrometer (Hollas, 2002). This thesis is exclusively concerned with the use of Raman spectroscopy.

### 2.7.1. An Overview of Raman Spectroscopy

Raman Spectroscopy, according to Gardiner (1989), is a well-established spectroscopic technique, used in both chemistry and condensed matter physics, to study vibrational, rotational and other modes of low frequency in a system. The wavelength and intensity of light that has been scattered inelastically, by a given sample, is measured by passing a laser through the substance under analysis (Bulkin, 1991). The process, described by Howley (2007, Section 2.2) involves the interaction between the stream of photons from the laser and the substance, which can occur in many different ways. Some of the photons can pass through the substance while others can collide with molecules from the substance. Most of the colliding photons deflect and remain unchanged, though some can lose or gain energy. A Raman probe can measure the *Raman shift* of the substance: This is the change in frequency and intensity of the scattered light. Every substance will have a unique scattering on the Raman spectrum, providing a chemical fingerprint, ideal for sample identification. The spectrum can also be used to quantify a particular substance in a mixture. Every point on the spectrum is represented by the intensity at a certain frequency. The *Raman shift* normally corresponds to wave-numbers or wave-lengths. Therefore, each attribute in a Raman spectroscopy dataset is essentially a point or recorded wave-number from its own specific spectra.

---

[19] PRTools is a Matlab based toolbox for pattern recognition. It can be used freely for academic research: http://www.prtools.org

[20] Receiver Operating Characteristic (ROC) curve is a plot of the fraction of true positives against the false positives.

[21] Source: http://ict.ewi.tudelft.nl/~davidt/dd_tools.html

- 32 -

### 2.7.2. Recent Classification Work in the Area of Spectroscopy

Madden and Ryder (2002) explored the use of Machine Learning techniques, in comparison to statistical regression methods, for identifying and quantifying illicit materials using Raman spectroscopy. In their work, the Machine Learning task was broken down into data reduction and prediction subtasks. The data were reduced, to improve the dimensionality. This was done by selecting some features and discarding all others. This process of feature selection was enhanced by using a Genetic Algorithm. Predictions can then be made based on only a small number of data points. The improvements that can be achieved by using several different predictor models together were also noted. This would come at the cost of increased computation but was shown to provide better results than any one predictor by itself.

Howley *et al.* (2005; 2006) investigated how useful a Principal Component Analysis (PCA) method called Non-Linear Iterative Partial Least Squares (NIPALS) could be in reducing a high-dimensional spectral dataset for use with several well-known Machine Learning algorithms. The authors used a dataset comprising 217 different examples of Raman spectra. The goal was to correctly classify acetaminophen[22], which was present in 87 of the 217 examples, from a variety of solid mixtures. An interesting finding of the work was that although SVM and kNN were no different in their respective best results achieved with or without PCA, when PCA was present, these good results were achieved with far fewer attributes. This meant that the resulting model would be much more compact when classifying new data.

O'Connell *et al.* (2005) proposed the use of Principal Component Analysis (PCA), Support Vector Machines and Raman spectroscopy to target an analyte[23] in solid mixtures. In this case, the analyte was acetaminophen. They used near-infrared spectroscopy to analyse a total of 217 samples, some of which had the target analyte present, of mixtures with excipients[24] of varying weight. The excipients that were included were sugars, inorganic materials, and food products. The spectral data were subjected to first derivative and normalisation transformations in order to make it more suitable for analysis. After this pre-treatment, the target analyte was then discriminated using Principal Component Analysis (PCA), Principal Component Regression (PCR), and Support Vector Machines. According to the authors, the superior performance of SVM was particularly evident when raw data were used for the input. The importance and benefits of the pre-processing techniques was also emphasised.

---

[22] Acetaminophen (also known as paracetamol) is a pain-reliever drug found in many over-the-counter medications.

[23] An analyte is a substance or chemical constituent that is determined in an analytical procedure.

[24] An excipient is an inactive substance used as a carrier for the active ingredients of a medication.

Conroy *et al.* (2005) investigated the use of Raman spectroscopy with chemometric and Machine Learning techniques to differentiate between chlorinated and non-chlorinated solvents. The research also quantified the chlorinated compounds if they were present in the solvent. In a direct comparison of chemometric and Machine Learning methods, the latter showed the best classification results. The authors found that the Ripper rule-learning algorithm showed the best results in each of the individual solvent classification experiments. The Support Vector Machine (SVM) and C4.5 decision tree also produced a notably low classification error rate with the SVM performing the best on detecting whether or not *any* of the chlorinated compounds were present. The authors raised an issue with the Raman intensity variation occuring from day to day that should be addressed in future research.

Datta and DePadilla (2006) conducted a comparative study applying clustering and classification algorithms to differentiate between cancer and non-cancer protein samples. The samples were represented using mass spectroscopy data. An evaluation of five unsupervised and six supervised classification algorithms was carried out. The techniques of *t*-test–Bonferroni and *t*-test–Westfall–Young feature selection were used in individual experiments. The results found show that the best clustering algorithm was Diana whereas the best classification results were achieved by the simpler of the algorithms, namely, the Linear Discriminant Analysis (LDA) and 1-Nearest Neighbour (1-NN). The authors emphasised the importance of both variable selection and the need for consideration of multiple methods on different datasets.

Zhao *et al.* (2006) used a Support Vector Machine to identify different categories of tea represented by near infra-red spectroscopy data. Principal Component Analysis (PCA) was first used to select five principal components that served as the input to the SVM. The dataset comprised three different categories of tea. The performance of the SVM was compared to a Back Propagation Artificial Neural Network (BP-ANN). The authors concluded that the SVM provided better generalisation which in turn would lead to better overall results.

Howley (2007) uses Machine Learning techniques for the identification and quantification of materials from their corresponding spectral data. The author showed that using Principal Component Analysis (PCA) with Machine Learning methods, such as SVM, could produce better results than the chemometric technique of Principal Component Regression (PCR). The author also presented customised kernels for use with spectral analysis based on prior knowledge of the domain. A genetic programming technique for evolving kernels was also proposed for when no domain knowledge may be available.

Xu and R. G. Brereton (2007) have posited a method of analysis to identify genotypes using fluorescence excitation-emission spectral data. This approach used Piecewise Direct Standardisation (PDS) followed by the construction of two one-class classifiers based on Principal Component Analysis Data Description (Tax, 2001). Results revealed very few false positive errors and between 2% to 6% overall false negative errors. They noted that such findings indicated that the one-class classifiers may have been overly strict.

Brereton (2009) described the use of chemometrics for pattern recognition. Chemometric techniques involve the data-driven extraction of useful information from a chemical system. The author provided an in-depth analysis of several 'real world' pattern recognition case studies which have taken place where chemometric techniques were successfully applied. These case studies spanned from a variety of sources including biology, pharmaceuticals and forensics. The chemometric techniques, explained by the author, were said to be a generic description of the principles rather than a comprehensive study of all available techniques from the literature. Some of the techniques that were described include Principal Component Analysis, Partial Least Squares, Self Organising Maps and Support Vector Machines. The use of these chemometric techniques was illustrated in conjunction with analytical techniques such as Atomic Spectroscopy, Thermal Analysis, Near Infrared Spectroscopy and Nuclear Magnetic Resonance Spectroscopy, to name a few. The main emphasis of this work was on classification methods, including chapters on one-class, binary and multiclass classifiers, although information was also provided on exploratory data analysis and data pre-processing.

# 3. Developing a One-Sided Classification Toolkit

This chapter will describe the design and development of the OSCAIL toolkit. Chapter Three will begin by providing a brief overview of the toolkit itself and what it is used for. The requirements specification, defining all aspects that were to be developed, will then be described in detail, followed by a high level look at the architecture. The toolkit's main components will then be listed and explained. This will be followed by the testing and evaluation that was carried out. Finally, some possible improvements for future upgrades will then be outlined.

## 3.1 Introducing the OSCAIL Toolkit

OSCAIL, which is an acronym for **O**ne-**S**ided **C**lassification **A**nd **I**nductive **L**earning, is an experimental toolkit for carrying out one-sided classification experiments. It is a Command Line Interface[25] (CLI) driven software package that contains several one-sided algorithms that can be chosen at runtime and used to create a new classifier based on a chosen ExampleSet[26] and a variety of different options. Both experiment specific and classifier parameter options can be set by the user. The toolkit was designed to carry out comprehensive and iterative experiments with minimal input from the user. The resulting classifiers that are generated can be saved and used at a later stage to classify new examples. The toolkit contains both dataset pre-processing and splitting techniques. Experiment results are printed to the screen as they are calculated; these include the classification error, sensitivity, specificity and confusion matrix for each run and individual fold. There are currently four algorithms implemented into the toolkit. These are One-sided Support Vector Machine, One-sided k-Nearest Neighbour, the One-sided k-Means and a Multi-Cluster One-sided Support Vector Machine. The toolkit was designed with a view to facilitate the addition of further algorithms in the future.

## 3.2 Requirements Specification

This section will describe the requirements that were proposed for OSCAIL at an early stage in the project. These requirements were categorised into four distinct subgroups which were loading and organizing the dataset, choosing the classifier and configuring the settings, viewing the performance and the ability to run multiple experiments.

---

[25] This is a text-only interface in which commands are typed in order to carry out the software tasks.

[26] This is the representation of the data used by the OSCAIL experimenter. It is fully explained in section 3.4.2

### 3.2.1 Organising the Dataset

#### 3.2.1.1 Load the dataset from a file

The path to the dataset will be read in from the user and then the data will be loaded. OSCAIL will work primarily with ARFF files with possible extension capabilities to deal with other file types in the future.

#### 3.2.1.2 Optional relabelling of the dataset

The user can indicate if they wish to relabel the dataset once it has been loaded. The toolkit should detect when the currently loaded dataset is not appropriate for one-sided classification (when there is more than just a target and outlier class present) and prompt the user to either load a different dataset or relabel the current one by labelling one class as the target and all the rest as outliers. If they choose to carry out the relabelling step, they should be prompted to save the relabelled dataset as a new file. They may also use the relabelled dataset for the experiment without saving it.

#### 3.2.1.3 Optional pre-processing of the dataset

The toolkit will provide several data pre-processing steps should they be required by the user. These will include techniques such as normalisation where the attributes are normalised to lie within a fixed range, for example from zero to one. Also, techniques for dimension reduction will be included. This involves transforming the data into a reduced amount of dimensions while still capturing the variation of the data. The toolkit should also facilitate the seamless addition of other pre-processing techniques in the future.

#### 3.2.1.4 Decide how the dataset will be split

The user will decide how to split the dataset into training and testing sets during the experiments. They will have the option of either providing a percentage for where the dataset will be split or they can choose to carry out n-fold cross validation (see Appendix A.3) and indicate how many runs and folds they would like to take place. If the user does not specify how they would like to split the dataset then a default splitting procedure will be carried out automatically. The splitting procedures for OSCAIL also stratify all of the examples. Stratification involves ensuring that the splits of the training and testing sets contain the same proportion of "targets" and "others", in each, as the original data. One-sided classification requires target examples for training. Without stratification, it may happen by chance that all of the target examples end up in the testing split which would be of no use.

### 3.2.2 Classifier Configuration

#### 3.2.2.1 Choose classifier

The user will have the ability to select which classifier they would like to use for their experimentation. A default classifier can be used when one is not specified by the user.

#### 3.2.2.2 Set classifier parameters

Most classifiers will require setting certain parameters that will affect the outcome of the classification results. The user can adjust these classifier parameters in order to determine which combination produces the best results.

#### 3.2.2.3 Save classifier

When a classifier is deemed satisfactory by the user, they can save it to a file so that it can be used in the future to classify new data that the user will also provide. The file name for the saved classifier will give details about the classifier including what date it was saved.

#### 3.2.2.4 Load classifier from file

The user can provide a path to a saved classifier and load it in order to carry out their experimentation. After loading the classifier they will be asked to provide the path to a dataset of previously unseen test examples. These will then be classified and the results will be displayed.

### 3.2.3 Performance Analysis

#### 3.2.3.1 Viewing results in real time

As the experiments are being run, the results will appear on screen as they are calculated. This will be useful, especially when carrying out model selection, so that the user could notice consistently bad results and may wish to restart the experiment with different values.

#### 3.2.3.2 Summary of results

Once the experiments are completed, results will be provided at the end that will summarise the overall details. This will be useful for the user by providing a quick analysis after a large experiment has taken place.

#### 3.2.3.3 Saving log file

For every experiment, a log file will be automatically saved. The name of this file will be auto-generated and will be dependent on what time the experiment was run.

**3.2.4 Multiple Experiments Runs**

The toolkit will enable the user to carry out multiple runs of the experiments with minimal input. These runs will differ from each other based on an incrementing seed for a random number generator which shuffles the instances before they are split into training and test sets. All of the options will be set at the beginning and then the experiments will run right through to completion. Summaries of the performance of each run will be displayed on screen as they are calculated. The averaged results over all of the runs will be displayed once the experiments have finished.

## 3.3 Architecture Design

In the initial design phase for creating the OSCAIL toolkit, a high level flow control diagram was created and explained in detail. Each component of this diagram was broken down into sub diagrams to explain the individual operations. Over time, when the software coding and analysis began, these diagrams were changed to reflect the updated architecture and overall improvements to the flow of control. The toolkit was updated and improved on a continuous basis and diagrams such as these helped to communicate such changes to the rest of the OSCAIL project members. The following section provides an insight into the early development of the system architecture and how it has changed over time.

### 3.3.1 Initial Design Phase

The original design for OSCAIL involved continuously prompting the user for the required information as it was needed. This information would include the path to the dataset, if relabelling was required, how to split the dataset into training and testing etc. This made carrying out experiments quite time-consuming and tedious. Figure 12 below shows an early illustration of how the flow of control would take place in the system.

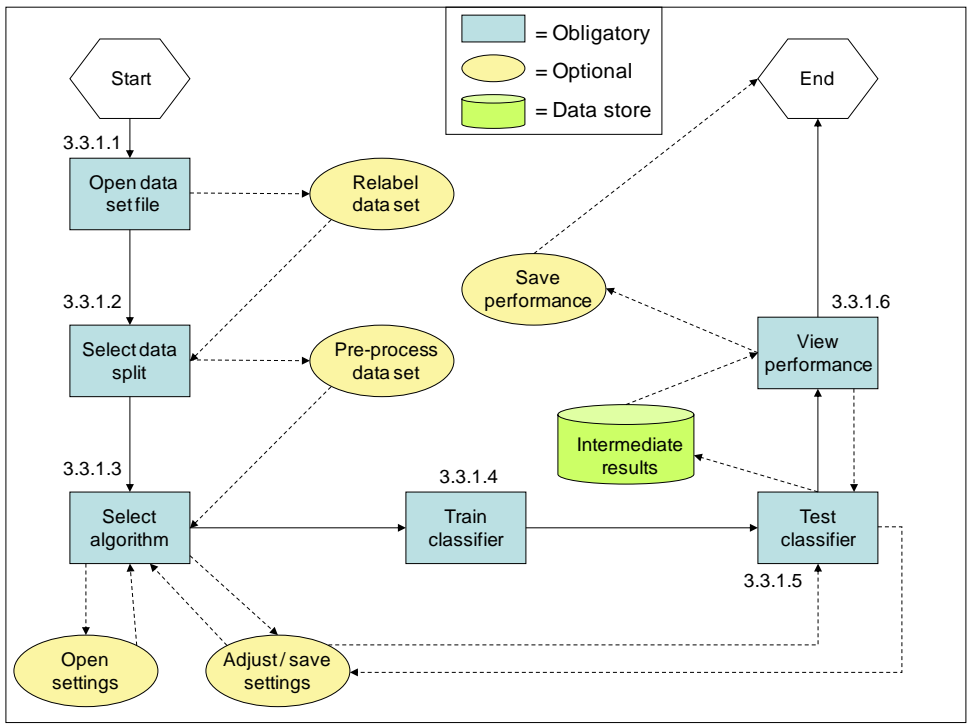

**Figure 12: Early flow of control design for OSCAIL**

<span style="background-color:cyan">Obligatory</span>: These are the basic sequence of operations that are required to run a one-sided classification experiment using OSCAIL.

<span style="background-color:orange">Optional</span>: These steps aren't always necessary but can be useful for improving the classification performance through pre-processing and tuning the classifier

<span style="background-color:green">Data Store</span>: This is used to hold the intermediate results of multiple runs

### 3.3.1.1   Open Dataset File

Once the Experimenter is run, the user is prompted to enter the path to an ARFF file that contains the dataset to be used in the experiment. The user is then asked if the dataset needs to be relabelled as a one-sided classification dataset with a single "target" class and one "other" class.

### 3.3.1.2   Select Data Split

After the dataset has been loaded, the user is asked how they would like to split the dataset into training and test examples. They are given the choice of using either a percentage split, where they enter a percentage to use for training and the remainder is used for testing, or n-fold cross validation, where they enter a value for n to split the dataset into folds. Once the split is complete, the user has the choice of using a pre-processing technique, such as normalisation, on the dataset.

### 3.3.1.3 Select Algorithm

Once the first two steps are complete, the user is asked to select an algorithm to use for the experiments. The original method to accomplish this involved checking the input against known keywords for algorithms already implemented in the toolkit. When the algorithm was chosen, the user could open a configuration file that contained previously saved parameter settings for the algorithm or they could alternatively adjust the settings directly by being prompted for the individual values.

### 3.3.1.4 Train Classifier

The classifier is trained using the chosen algorithm and the training examples resulting from the dataset split.

### 3.3.1.5 Test Classifier

The trained classifier is tested by using the test portion of the dataset. In multiple experimentation runs, the classifier can be tuned based on the results from the test data.

### 3.3.1.6 View Performance

The details of the performance of the classifier are displayed at the end of the experiment. This included showing the values for the experiment and classifier specific settings as well as the results for the different iterations of the experiment. The user is given the option to save these details to a file for future reference.

Once a beta version of this design was implemented, several alterations were proposed that would improve the overall efficiency and usability of the toolkit. This led to a different flow of control which will now be explained in the following section.

### 3.3.2 Improving the Design

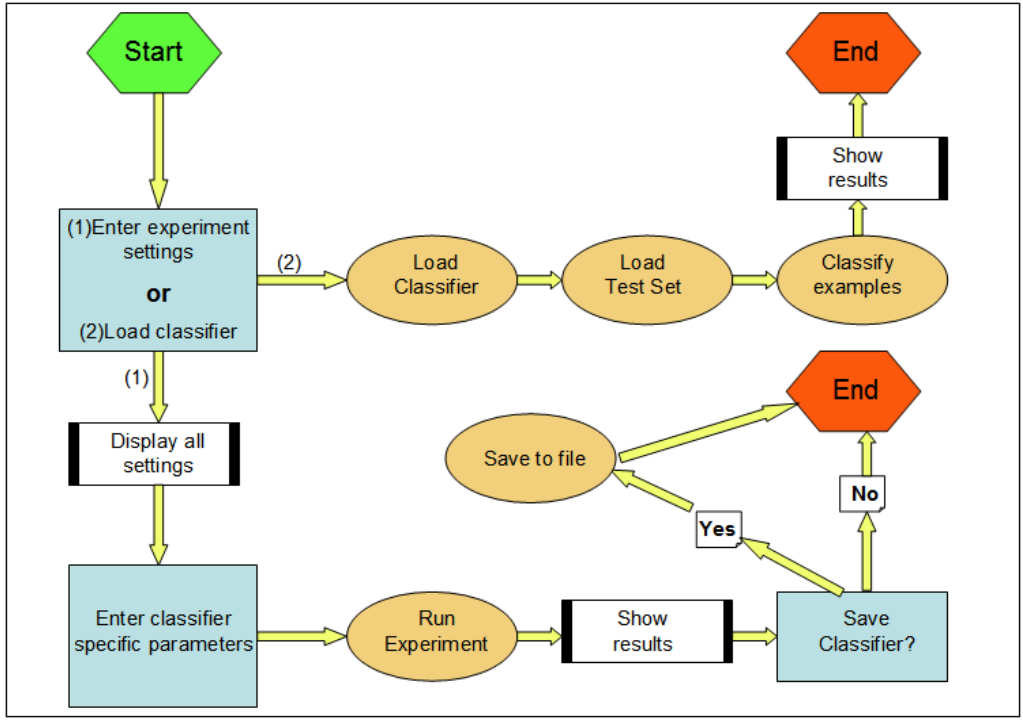

**Figure 13: Improved flow of control for OSCAIL**

The new design introduced the use of command line switches for inputting parameter values and experiment settings. All the experiment variables can be set using one string of arguments when the Experimenter is first run. If there is only a dataset path, and no other values passed, then the default settings for the experiment are used. If nothing at all is passed, the user is asked if they would like to load a classifier from a file. This process is explained in Section 3.4.1. The new flow of control is a result of simplifying the design. All of the input is taken in together for both the experiment settings and the classifier to avoid unnecessary continuous prompting of the user.

## 3.4 Toolkit Components

There are many different aspects of the OSCAIL toolkit that make it both interesting and unique. The following section will outline some of the individual components in the toolkit.

### 3.4.1 Experiment Options at Runtime

The following options for each experiment can be set by the user when the software is run. These options are selected by using command line switches and passing the values. Each option is listed and described below:

*- ExampleSet Path: (-E <fileName>)*

The path where the ARFF file to be loaded can be found. Data for the experiment is loaded from this ARFF file and stored as an ExampleSet in the toolkit. A description and explanation of an ARFF file can be found in Appendix A.1.

*- Relabelling required: (-R <yes/no>)*

This is the choice as to whether or not the ExampleSet needs to be relabelled as one-sided. This option allows the user to load an ARFF file that contains many different classes. By supplying the name of the chosen target class and a directory for the output file, the ExampleSet can be relabelled for one-sided classification and then saved. The user can use the relabelled file without saving it if they wish.

*- Normalisation required: (-N <yes/no>)*

A normalisation step of the ExampleSet can be carried out during the experiment if the user chooses this option. Attributes are often normalised to lie within a fixed range, for example, from zero to one.

*- Algorithm to use: (-A <algorithmName>)*

The user can specify which algorithm is to be used. A default algorithm is used if no particular one is chosen

*- Technique to use: (-T <techniqueName>)*

Either performance estimation or model selection can be chosen. The former estimates how well a classifier will perform in the future and the latter iterates over many user-provided algorithm parameters to select those that give the best results.

*- How to split the ExampleSet: (-S <ps/cv>)*

A percentage split (abbreviated as ps in the toolkit) or n-fold cross validation (abbreviated as cv in the toolkit) can be used for splitting the ExampleSet into training and testing Examples. For percentage split, the ExampleSet is split into training and testing sets based upon a percentage value that is entered by the user. Cross validation breaks the ExampleSet into n random folds. In turn, one fold is held out for testing and the remaining n-1 folds are used to training.

*- Amount of runs: (-r <numRuns>)*

The experiments can be run many times producing different results based on the random number seed provided by the user. For each successive run the Examples are shuffled and once again split into training and testing sets.

*- Amount of folds: (-F <numFolds>)*

The number of folds to use for cross validation.

*- Random number seed: (-s <seedNumber>)*

This seed is used for shuffling the ExampleSet examples prior to splitting. Every different seed number will give a different random pattern for the shuffled Examples. This seed value is incremented for every run.

*- Percentage for splitting: (-P <percent>)*

The percentage used for training; the remainder is used for testing. For example, if the user chooses 80 as the percentage, 80% of the Examples are then used for training the classifier and the remaining 20% are used for testing.

*- Training split used: (-t <ps/cv>)*

How to split the training set into a training and validation set. This split can be either a percentage split or a cross validation split.

*- Training split percentage: (-p <trainPercent>)*

This is the percentage used if a percentage split has been chosen.

*- Training split folds: (-f <numTrainFolds>)*

This is the amount of folds if cross validation has been chosen

All of these options have default values that are used if no others are provided. If the user does not enter any arguments when the Experimenter is first run, a list of all possible options and how they are set is displayed on screen. The user is also asked if they would like to load a previously saved classifier.

```
--- OSCAIL - One Sided Classification and Inductive Learning ---

---------------- Experimenter Version 1.0 ---------------------

OSCAIL option usage details:

-E <Path to the example set>
        The example set to use in the experiment.
-R <yes/no>
        Relabel the example set after loading? (default: no).
-N <yes/no>
        Normalize the example set? (default: no).
-A <algorithm name?>
        Which algorithm to use?: (default:KNN)
-T <technique choice?>
        Enter pe for Performance Estimation or ms for Model Selection (default: pe).
-S <ps/cv>
        Enter ps for percentage split or cv for cross validation (default: ps).
-r <Number of runs>
        The amount of runs in the experiment (default: 1).
-F <Number of folds>
        The amount of folds in the experiment (for cross validation) (default: 3).
-s <Initial random number seed>
        The initial random number generator seed for shuffling (default: 2)
-P <Percentage split percent>
        The percent used for the percentage split of the example set (default: 50)
-t <Training split used?>
        Which way the training set with be split (percentage split or cross validation) (default: 1)
-f <Training split folds?>
        Amount of folds used for the split of the training examples (default: 3)
-p <Training split percentage?>
        Percentage used for the split of the training examples (default: 50)

Would you like to load a previously saved classifier? (yes/no):
```

Figure 14: Option settings explained to the user.

Then, based on the algorithm chosen, the corresponding parameter value options, for that algorithm, can be set in a similar fashion using command line switches. Text highlighted in green is entered by the user (See Figure 15 below).

```
-D sequence <start> <increment> <end>
or
-D individual <D value1> <D value2>...
        The distance metric to use (Euclidean=e, Manhattan=m, Cosine=c) (default: e).

Example usage below;
Type: -M individual 1 2  -K individual 3 2  -T individual 9.0 7.0 -D individual c m

Please type out a single string to set the options,
as described above. Otherwise, the defaults will be chosen.
(Press Enter with no text to skip this step.)
:
-M individual 1 3 7 -K sequence 1 1 5 -T sequence 1.0 1.0 5.0 -D individual e c m

KNN Classifier Options selected:-

-M (M neighbours used)     --->  1 3 7
-K (K neighbours of M)     --->  1 2 3 4 5
-T (Threshold used)        --->  1.0 2.0 3.0 4.0 5.0
-D (Distance metric used)  --->  e c m

Experiment Options selected:-

-E (Example set path)       --->  C:\iris.arff
-R (Relabeled?)             --->  false
-N (Normalized?)            --->  false
-A (Algorithm to use?)      --->  KNN
-T (Technique to use?)      --->  Performance Estimation
-S (Example set split?)     --->  Percentage Split
-r (Number of runs?)        --->  1
-F (Number of folds?)       --->  0
-s (Random number seed?)    --->  2
-P (Percentage for split?)  --->  50.0
-t (Training set split?)    --->  Percentage Split
-f (Training split folds?)  --->  0
-p (Training split %?)      --->  50.0
```

**Figure 15: An example of how the classifier specific options are set.**

### 3.4.2 ExampleSet and Example

An OSCAIL ExampleSet is the result of reading in a file that is in ARFF format. After the file has been read in, instances are represented as the type Example. An ExampleSet can also be used to reproduce an ARFF file or to re-label the Examples and then create the corresponding ARFF file (see Figure 16 below). If the user attempts to load an ARFF file containing many different classes, they are prompted to either provide the class name that is to be set as the target (the rest are considered to be outliers) or else load a different dataset that contains only one target class and one class representing outlier examples.

```
%################ O S C A I L ##################
%#One-Sided Classification and Inductive Learning#
%################################################
%
%The iris example set has been relabeled to
%only contain one Target class and one Other class.
%
%[Target Class = "Iris-setosa"], [Other Class = All others]
%
%The old class options were written as follows:
%@attribute class { Iris-setosa, Iris-versicolor, Iris-virginica, }
%
%
%
@relation iris
@attribute sepallength real
@attribute sepalwidth real
@attribute petallength real
@attribute petalwidth real
@attribute class {"Other", "Target"}
@data
%
%
5.1, 3.5, 1.4, 0.2, Target
4.9, 3.0, 1.4, 0.2, Target
4.7, 3.2, 1.3, 0.2, Target
4.6, 3.1, 1.5, 0.2, Target
5.0, 3.6, 1.4, 0.2, Target
5.4, 3.9, 1.7, 0.4, Target
4.6, 3.4, 1.4, 0.3, Target
5.0  3.4  1.5  0.2  Target
```

**Figure 16: ARFF file produced from a relabelled ExampleSet.**

The ExampleSet class from the toolkit has several different methods to access all of the vital information about the dataset that has been loaded. The class diagram, in Figure 17 below, lists all of these methods that are available.

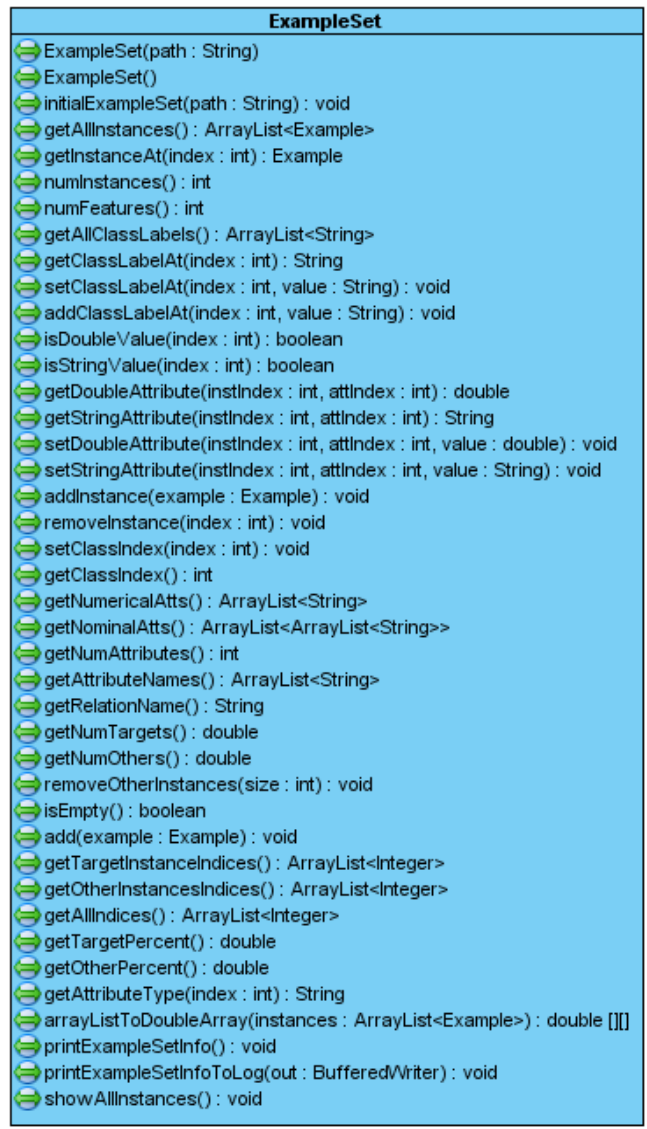

**Figure 17: Class diagram for the ExampleSet class from the OSCAIL toolkit**

### 3.4.3 Performance Estimation and Model Selection

When the user is running an experiment, they have a choice of either Performance Estimation or Model Selection. The purpose of the former is to create an estimate of how the classifier will perform on future unseen data. The latter selects the best model by testing a variety of parameter options and recording how well they perform.

### 3.4.4 Keywords "individual" and "sequence"

During an experiment, individual classifier parameter options can be given several different values by the user. Then, all the different combinations of these values are tested and their performance is recorded. This ensures that comprehensive and thorough experiments can be run with minimal input from the user. When setting the values for these classifier options, one of two keywords can be chosen.

*individual:* The individual keyword is followed by any amount of individual values for the option in question. These are all separated by a space and the keyword must be used after the switch for every option.

```
Please type out a single string to set the options,
as described above. Otherwise, the defaults will be chosen.
(Press Enter with no text to skip this step.)
:
-M individual 1 2 3 4 5 -K individual 1 2 3 4 5 -T individual 1.0 2.0 3.0 -D individual e c m
```

**Figure 18: An example of the user setting the options using the individual keyword**

*sequence*: The sequence keyword is used for entering a sequential pattern of values for a particular parameter value. Three arguments are provided directly after this keyword. The first is the starting value for the parameter in question. The next value is how the pattern is to be incremented. Finally, the third value specifies the last parameter value, or stopping condition of the pattern.

```
Please type out a single string to set the options,
as described above. Otherwise, the defaults will be chosen.
(Press Enter with no text to skip this step.)
:
-M sequence 1 1 7 -K sequence 1 1 7 -T sequence 1.0 1.0 5.0
```

**Figure 19: An example of the user setting the options using the sequence keyword**

From Figure 19 above, the parameters for the classifier would be set as follows:

| M | Values = (1, 2, 3, 4, 5, 6, 7) |
|---|---|
| K | Values = (1, 2, 3, 4, 5, 6, 7) |
| T | Values = (1.0, 2.0, 3.0, 4.0, 5.0.) |

**Table 1: Algorithm parameter values set using the "sequence" keyword**

### 3.4.5 Arrays of index references

Once an ARFF file has been loaded and is stored as an ExampleSet, from then on, any splits of this ExampleSet are carried out by using references to it and not by creating new ExampleSets. The original ExampleSet always stays intact and, for splitting, referencing the individual Examples index is used. This makes the process for recreating experiments much more efficient.

### 3.4.6 Default and Automatic Experiment Settings

Every option that can be set for both the experiment itself and the individual classifier options all have default values that are used should no options be selected by the user. Also, while setting the options for an experiment, if an unknown input value is read in for an option, it is automatically set to the default and the user is notified.

### 3.4.7 Log files

When an experiment is run, all the details are stored in a text log file.
This contains the following information:

- All experiment and classifier specific settings used
- Training and testing instances for each run
- Best parameters and error estimate for each run.
- Matrices describing classification accuracy for each run.

### 3.4.8 Output of Results

A summary of the results for the experiments are displayed on screen as they are calculated. All these results and settings are also stored in the log files for future reference. Once the results have been displayed on the screen, if the run was for Model Selection, the users are asked if they would like to save the classifier to a file.

```
|------------------|--------------------|------------------|
|     Target       |         17         |        4         |
|------------------|--------------------|------------------|
|     Other        |         2          |        5         |
|------------------|--------------------|------------------|

M: 4 K: 4 Threshold: 4.0 Dist metric: c Error estimate: 0.21428571428571427
--------------------------

-----------------------------------------------------------
|---------------------------------------------------------|
|                  | Target predictions | Other Predictions |
|------------------|--------------------|------------------|
|     Target       |         21         |        0         |
|------------------|--------------------|------------------|
|     Other        |         7          |        0         |
|---------------------------------------------------------|
-----------------------------------------------------------

M: 4 K: 4 Threshold: 4.0 Dist metric: m Error estimate: 0.25
____________________________________________
Model Selection Results:
--------------------------------------------
Smallest Error Estimate -> 0.03571428571428571
Best m              ---------> 1
Best k              ---------> 2
Best threshold      ---------> 4.0
Best distance metric ----> c
--------------------------------------------

Would you like to save this classifier?(yes/no):
```

**Figure 20: An example of the output for model selection**

### 3.4.9 Saving Classifiers and Reusability

Once the user has decided upon the best model to use, as a result of running the model selection process, they can save the resulting classifier to a file and reload it at a later time. The filename for the saved classifier is automatically generated with the name of the algorithm used, a system timestamp and the extension ".oscail". All classifiers that are saved can then be reloaded by running the experimenter without any arguments and then providing the path to the required classifier. The user will then be asked to load a test set to analyse how well the saved classifier performs.

## 3.5 Testing and Verification

This section will provide a summary overview of how testing was carried out during the development of the toolkit. The procedure used for testing and evaluating each of the toolkit's components will now be outlined briefly in the following subsections.

### 3.5.1 Reading ARFF files and Producing ExampleSets

The toolkit contains individual modules that were created to access ARFF file information such as the relation name, number of features, types of features, number of instances etc. This information is then used to create corresponding ExampleSets. Once these modules were created, they were all tested as individual units and verified manually using the original ARFF file as a reference. Early testing involved using small datasets in which manual inspection could be easily carried out to ensure that the header file information and data was being read in correctly. Much larger, high dimensional, datasets were then tested to ensure that the toolkit could handle all relevant datasets that had been found in the literature and online. All of the results were examined using a text editor to verify the functionality of the code.

### 3.5.2 Relabelling and Saving ExampleSets as ARFF files

The first set of tests carried out for relabelling ExampleSets involved the correct detection of unsuitable datasets. This involved iterating through the labels and setting a Boolean value to false if a label was encountered that was neither "Target" nor "Other". When this was verified, the correct relabelling, based on the users choice of target, was then tested. This is obviously a very important step, so a wide variety of different datasets, of varying size and number of labels, were tested. When the verification of the relabelling was complete, through manual inspection, the data had to be written out correctly to be saved as an ARFF file (as this is an option that is presented to the user). This newly saved file and the original file were compared using a file-compare program, called UltraCompare, to make sure that the structure remained the same and that the labels in the saved file had been changed correctly. Figure 16, from Section 3.4.2 above, shows an extract from an ARFF file that was produced and saved from a relabelled ExampleSet.

### 3.5.3 Splitting the Datasets

Testing was carried out for both the percentage split and cross validation modules of the toolkit. For the percentage split, the main testing involved running checks to see if a valid percentage was entered by the user and that, following the split, there were examples present in both the training and testing sets. A check was carried out to ensure that the percentage was rounded up or down according to the number of examples present. Datasets with just a few examples were tested, with a variety of percentages, to note the behaviour in extreme cases. Cross validation was tested by writing out all the folds to a text file and then manually inspecting each of the folds. The extreme cases were also checked such as selecting 5-fold cross validation when there are only 4 examples present to catch and handle the errors in an appropriate manner.

### 3.5.4 Stratification of the Examples

The stratification procedure was again checked using small amounts of examples and also when the class labels were severely imbalanced. The stratified splits, similar to the cross validation testing, were written out to a file and verified.

### 3.5.5 Normalising the Examples

Calculations from the normalisation module were carried out on a variety of different datasets and then the resulting values were written out to a text file. The same calculations were then carried out using Microsoft Excel so that they could be directly compared with the text file.

### 3.5.6 Adding and Testing New Classifiers

The most important aspect of the toolkit is the ability to add new classifiers and to ensure that they are comprehensively tested in collaboration with all of the other modules such as interacting with the dataset and the splits etc. Adding new classifiers took place after all of the other modules were deemed to be operationally sound through testing. All the classifiers were first tested individually before being integrated into the toolkit. An abstract class called "Classifier" was then created which defined several abstract methods that each new classifier had to implement.

### 3.5.7 Index References to the ExampleSets

When multiple runs of an experiment are taking place there can be many splits of the data. In the beginning, several smaller ExampleSets were created for this reason. This worked well for small datasets but, through testing, it was realised that this consumed a lot of system resources as the datasets grew in size. An improvement to the toolkit was decided upon which involved just storing references to one original ExampleSet. This was a large scale upgrade that involved changes to a lot of classes; the most important being the splitting classes. Tests were first carried out separate to the toolkit to see if the implementation idea was feasible. ExampleSets were created and then split using the referencing system. These splits were written to a file and validated against the original version splits. Once these external tests were complete, the whole toolkit was updated, one class at a time, and tests were carried out at each stage to verify the functionality.

### 3.5.8 Parsing Command Line Input

The testing of the parsed input module had to be very extensive given the importance of this step and the variety of ways in which errors could occur. Firstly, it was important to check that each switch was correctly saving the supplied list of arguments from the user. This testing was carried out externally first and then implemented in the toolkit. A list of various circumstances that could result in errors was drawn up and code was written to warn the user in each case and re-prompt for new correct input if necessary.

**Error Scenarios:**

- Unknown switch provided / No switch provided (values only)
- Incorrect, misspelt or missing keyword
- No values for a specific switch that was provided
- Not enough / Too many arguments for the "sequence" switch
- Incorrect input format: string/numeric
- Ensure default values are used when nothing is provided

### 3.5.9 Internal Validation Splits

The biggest issue that was noticed as a result of testing this feature was that the overall splits of the data were conflicting with the internal ones. In order to prevent this, new methods had to be added to the splitting classes. Once the splits were completely independent, the same procedures for testing the overall splits, as described earlier, were used on the internal ones.

### 3.5.10 Saving Classifiers to a File

The testing for this component of the toolkit involved first checking to see that the serialised files were being correctly written to and read from the file. All the settings and parameters were listed and then the classifier was saved. When the classifier was reloaded from the file, all of the settings were checked against those that had been set. The next step was to test the integration of this component into the toolkit. To ensure that all of the classifiers that were saved had different names, a system timestamp was used as part of the file name. The most important testing carried for saving the classifiers was to make sure that the whole classifier was encapsulated correctly when saved and that it could be reproduced, exactly as it was before, from the saved file. This involved trial and error tests of the classifiers to compare and verify the settings before and after they were saved to a file.

### 3.5.11 Output of Results

The classification error, sensitivity, specificity, Balanced Accuracy Rate and the confusion matrix for all runs is displayed on screen and recorded in a corresponding log file. All of these calculations in the code were first verified using Microsoft Excel before being added. These results are all based on the prediction mechanism of the classifier and therefore all of the predictions being made were written to a file and the same calculations were carried out externally to ensure they were the same as those being calculated by the toolkit.

### 3.5.12 Testing All Modules and Settings

All the components of the toolkit were developed and tested incrementally as described. Once this was completed it was important to thoroughly test the toolkit as a whole. A list of different scenarios and settings were drawn up and all aspects of the toolkit were carefully tested. Any bugs that were encountered in the code were taken note of and later addressed in further updates. The following table, Table 2, shows a high level example of a testing checklist.

|  | | Suitable | Not Suitable |  | Check |
|---|---|---|---|---|---|
| **Load dataset** | | Continue.. | Warning and display options | | OK |
|  | | | | | |
|  | | *Relabel* | *Load different dataset* | | |
| **User options** | | Target class? | Dataset path? | | OK |
|  | | | | | |
|  | | *Target OK* | *Target Not OK* | | |
| **Relabel** | | Continue.. | Warn, ask to re-enter | | OK |
|  | | | | | |
|  | | *Path OK* | *Path Not OK* | | |
| **Other dataset** | | Check dataset | Warn, ask to re-enter | | OK |
|  | | | | | |
|  | | *Yes* | *No* | | |
| **Normalization** | | Apply | Continue.. | | OK |
|  | | | | | |
|  | | *Valid* | *Invalid* | | |
| **Experiment settings** | | Continue.. | Warn, use defaults | | OK |
|  | | | | | |
|  | | *Correct* | *Incorrect* | | |
| **Check display (dev)** | | Continue.. | Abort, fix problems (code) | | OK |
|  | | | | | |
|  | | *Valid path* | *Invalid path* | | |
| **Load Classifier** | | Continue.. | Warn, ask to re-enter | | OK |
|  | | | | | |
|  | | *Valid path* | *Invalid path* | | |
| **Load test set** | | Continue.. | Warn, ask to re-enter | | OK |
|  | | | | | |
|  | | *Valid* | *Invalid* | | |
| **Classifier Parameters** | | Continue.. | Warn, ask to re-enter | | OK |
|  | | | | | |
|  | | *Save OK* | *Error* | | |
| **Save Classifier (dev)** | | Continue.. | Abort, fix problems (code) | | OK |
|  | | | | | |

**Table 2: A high level example of the overall testing checklist**

### 3.5.13 Removing Test Code

When all of the testing was complete it was important to read back over all the code and remove excess code that had been added only for the purposes of testing. This would help to avoid confusion for anyone reading the code in the future. When this was complete it was necessary to re-run the overall tests to ensure that the functionality of the toolkit was not affected in any way by the removal of such test code.

### 3.5.14 Verifying Code Descriptions

The next step involved a further iteration over all of the code, this time to ensure that the descriptions and comments were clear and concise throughout. This step would help to simplify the understanding of the current code when carrying out any further updates.

## 3.6 Possible Improvements for the Toolkit

Since the beginning of its development, the OSCAIL toolkit has been constantly updated with new features and improved functionality. This section describes a few further improvements that may be possible in its future development.

- **Graphical User Interface (GUI)**

    At an early stage of the toolkits development it was decided that a GUI was unnecessary and that the focus of the software would be applied through textual input and output from the command line. Now that this implementation has been completed, it may be worthwhile to add a graphical interface to make the toolkit more accessible to users. While it is still not completely necessary it could still be worthwhile.

- **Extension of Existing Software**

    There are many existing applications, see Sections 2.6, for carrying out classification tasks. The OSCAIL toolkit could work as an add-on to classification software that does not currently support one-sided classification. To facilitate this upgrade, suitable software would have to be identified and the authors would be contacted. The toolkit would also have to have better support for other dataset representations and file types.

- **Comprehensive User Guide**

    Although there is a large amount of information documented which describes both the development and implementation of the toolkit, it would be useful to have a user guide directed specifically at end-users. This could provide example usage and directions on how to use individual features as well as listing the merits and drawbacks of particular algorithms and techniques.

- **Larger Variety of Algorithms and Sample Datasets**

    Adding other one-sided classifiers to the toolkit will greatly improve the options that the user will have at their disposal. Also, over the course of the software development many datasets have been collected. Packaging these and organising them cohesively could also be very useful. It will also be very important to correctly reference each of them giving credit to their respective authors.

# 4. On the Classification of Out-Of-Sample Data

This chapter will first consider the implications of working with closed datasets and the potential drawbacks of the standard assumption that the test examples are drawn from the same distribution as the training examples. The effect of adding "unexpected" outliers to the test set, that is, test examples belonging to a completely different distribution than the training examples, is then described. The applicability of transfer learning to "unexpected" outliers and concept drift will then be addressed. Chapter Four will conclude with some experimentation involving hand-written digit recognition data where "unexpected" outliers are incrementally added to the test sets.

## 4.1 Working with Standard 'Closed' Datasets

Classification algorithms are evaluated in a straightforward manner, for example, by dividing a given dataset into training and testing subsets, or by using a cross validation procedure. The outcome of such an approach is that the test datasets are drawn from the same distribution as the training datasets: This corresponds to what is known as the 'closed-set' assumption made by binary classification algorithms. In such a scenario, binary classification algorithms tend to perform better than one-sided classification algorithms. It is for this reason that multi-class classification algorithms are often found to be suitable for a task, even though a one-sided classification algorithm might be a better choice from a theoretical perspective. After all, in some applications, such test datasets do not reflect the true 'open' nature of the data that such classifiers will be exposed to in deployment. If it is known *a priori* that the outliers are not sufficiently characterised in the training set, then it is inevitable that the classifier will be eventually exposed to such outliers. When this is the case, the performance of multi-class classifiers begins to deteriorate and can become both erratic and unpredictable.

## 4.2 Introducing the Concept of "Unexpected" Outliers

One-sided classifiers are generally chosen in practice when outlier training examples are either extremely rare or entirely absent. However, they are also suitable when there are many outlier examples but when such examples do not form a complete statistical representation of all possible outliers. Although several algorithms for one-sided classification have been proposed over the past decade, standard multi-class algorithms still tend to be employed, even in situations where they may not be ideal from a theoretical perspective. Sometimes one-sided classifiers can be more reliable than multi-class ones, especially when they are presented with "unexpected" outliers that differ significantly from the outlier examples in the training set.

### 4.2.1 Definition of "Unexpected" Outliers

These are essentially examples which violate the Identical Distribution assumption that is made by conventional multi-class classifiers. We define this term as:

> *"Test examples which are **not** targets and which are **not** taken from the same distribution as the training examples."*

When it is not feasible to fully characterise the negative concept in the training data then it is inevitable that these out-of-sample examples will be encountered in practice. Their existence can pose a significant challenge to multi-class classifiers.

### 4.2.2 A Schematic Illustration

In order to fully understand how these "unexpected" outliers come about, we can study the following illustration, Figure 21 below, which represents two different scenarios. The first of these, Scenario 1, depicts in-sample test instances from a closed set of examples. The classifier has no problem generating a separating boundary between the target and outlier examples.

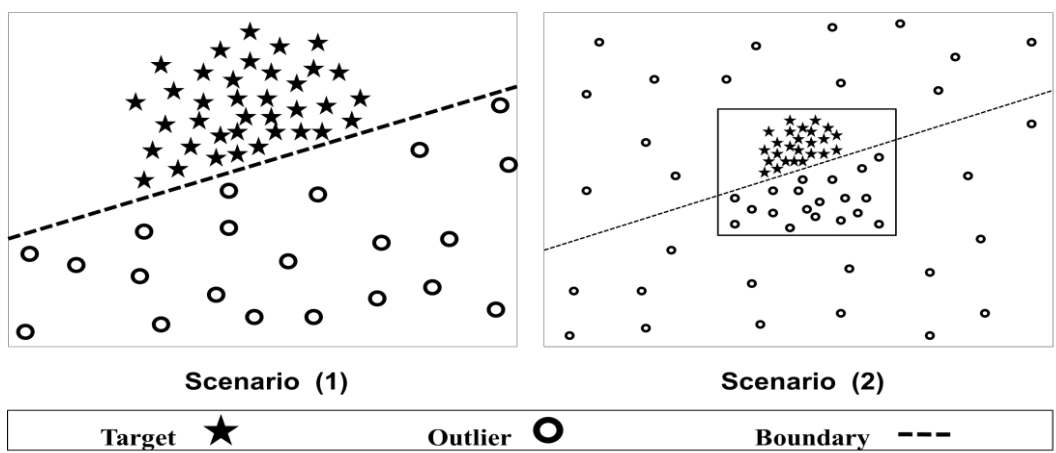

**Figure 21: Schematic illustration of in-sample and out-of-sample test data**

Scenario 2 considers the numerously dispersed out-of sample examples that surround the closed set in all directions. These represent all examples that are not targets and that fall into the "*everything else"* category. In such a scenario, it would be expected that the classification ability of multi-class classifiers would become erratic as the distinguishing boundary would no longer be effective given that "unexpected" outliers appear on both sides.

### 4.2.3 Some Practical Examples

There are a variety of applications in which it is not feasible to sufficiently characterise the outlier concept in the training set. A selection of these examples will now be described briefly.

The task of identifying text examples written by a particular author is an interesting problem. If we take Shakespearean literature as an illustrative example, we can easily collect many negative examples but it is very difficult, if not impossible, to ensure that these are representative of everything *not* written by him. If a binary classifier is trained using excerpts from Shakespeare's work as positive examples and excerpts from the works of other English playwrights and poets as negative examples, the performance is likely to be unreliable if presented with text from today's newspaper or from a Russian novel. Conversely, if a one-sided classifier is trained to distinguish Shakespeare's work from any other texts, its performance is likely to be more robust when presented with such "unexpected" outliers.

As another clear example, we can look at the classification task of detecting machine faults in industrial process control. There would be an abundance of data describing how the machine operates correctly. A problem arises, however, when we try to fully characterise all the possible ways in which the machine could fail to operate correctly. Once again, it is rather naïve to think that such a sufficient characterisation of the outlier concept is possible. The scope of such a characterisation is much too vast.

Similarly, as a final example, we could attempt to distinguish between chlorinated and non-chlorinated solvents based on their chemical spectra. This process will be fully described in Section 5.1.1 but for now, we will consider the main points. Target solvents comprise those that are chlorinated in any way. The outliers consist of all solvents that are not chlorinated. It is very straightforward to provide plentiful examples of non-chlorinated solvents for training. However, these will not even come close to ever statistically representing all possible non-chlorinated solvents that could be met in practical deployment of the classifier.

## 4.3 Transfer Learning and Concept Drift

Transfer learning (Thrun, 1996; Baxter, 1997; Caruana, 1997) involves using knowledge from one set of tasks to improve the performance of a related set of tasks. In the context of classification, it involves using labelled data from "similar" classification tasks to improve the performance of the task at hand. It is also suggested in the literature to be a solution when the test set is taken from a different distribution to the training set (Dai *et al.*, 2007). Could this be a possible solution to dealing with "unexpected" outliers that have been previously discussed? As an example of transfer learning, Pan and Yang (2008) look at Web document classification which is concerned with classifying Web documents based on a set of predefined categories. They note that, for instance, one could have labelled examples of university Web pages which are linked to specific category information based on a manual labelling procedure. For newly created Web sites, however, the features and distribution of the data may be very different. As a result, there could be a lack of labelled training data and it may be impossible to apply the Web-page classifier, learned using the university Web site data, to the new Web site. When this is the case, transferring the classification knowledge into the new domain could prove to be very useful. Rosenstein *et al.* (2005) note that the challenge of transfer learning is to know *what* knowledge to transfer over and *how* it is going to be transferred. They also emphasise the point that an important decision has to be made as to whether or not transfer learning is worthwhile as, in some cases, its use can actually hinder the performance The approach of transfer learning does not apply to "unexpected" outliers for the following reasons: These outlier examples do not form part of a different or parallel task, they have just not been well characterised in the training set and, therefore, the possibility of being encountered in practice remains. The reason that these examples are termed "unexpected" is because, while their existence remains part of the initial task, the concept that they form part of is much too vast to fully represent them as training examples. Therefore, it is not possible to transfer any knowledge that will aid in the classification of these particular types of outliers.

The statistical properties of many real-world concepts can change over time. Examples of this can include customer preferences and rules for predicting the weather or climate change. When a target concept is changed over time, this is known as the problem of *concept drift*. For a formal definition of concept drift see Stanley (2003). Learning models, which were trained on older data, become inconsistent and less effective as the concept changes. The problem of "unexpected" outliers differs from concept drift in that, with the former, we have a static concept, but over time the weaknesses of the training data are exposed. Of course, re-training might be possible, if problem cases were identified and labelled correctly, but this research is concerned with classifiers that have to maintain a robust performance without the need for re-training.

Solutions for transfer learning and concept drift can, at first glance, appear to be applicable to the problem of "unexpected" outliers. However, they still remain as two very different scenarios.

## 4.4 Experimentation: Introducing Unexpected Outliers

In this section, we will look at a set of experiments that introduce "unexpected" outliers to the test sets so that the trends in performance can be analysed for both multi-class and one-sided classifiers. A description of the experiments and the dataset used will first be described and this will be followed by the results and a brief analysis.

### 4.4.1 Multi-Feature Digit Dataset

This dataset, compiled by Duin (1999), consists of features of handwritten digits from '0' to '9' that were extracted from a collection of Dutch utility maps. There are 200 patterns per class (for a total of 2,000 patterns) which have been digitized into binary images. These digits are represented in terms of six feature sets (individual files). The specific feature set used for these experiments is the profile correlation representations of the data which consists of 216 features.

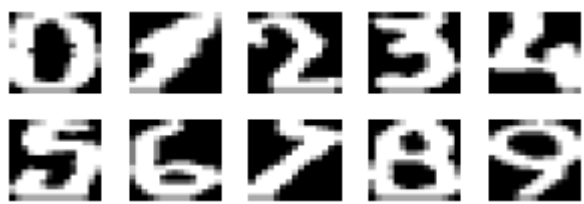

**Figure 22: Some examples of the handwritten digits**
**Source: (Tax, 2001)**

The dataset was modified as follows: The original files were converted to ARFF files and relabelled to only contain a "target" and an "other" class. The primary dataset set used (a subset of the original) comprised all 200 instances of the digit '2' relabelled as the target class with all of the instances of digit '3' (200 also) being used as the "expected" (same distribution) outliers. A secondary dataset was compiled to represent the "unexpected" (different distribution) outliers. This was made up of 200 instances (consisting of 25 of each remaining digit) that were not used for the primary dataset ('0', '1', '4', '5', '6', '7', '8', '9').

### 4.4.2 Differentiating Between "Expected" and "Unexpected" Outliers

The reason that the digit '3' is termed "expected" is due to the fact that there are examples of these outliers available to the classification algorithm at the time of training. Therefore, if the classifier runs into test examples of the digit '3' when it is deployed, its classification mechanism will be better equipped to distinguish these examples, as outliers, from the target class. This is certainly true in the case of multi-class classifiers which assume that all classes are well characterised at the time of training. One-sided classifiers, on the other hand, do not require these "expected" outliers in the training set in order to perform well. While they can sometimes make use of such outliers to tighten the classification boundary, the one-sided classification algorithms' primary concern, at the time of training, is to have a statistically representative set of the target examples.

The "unexpected" outliers in these experiments consisted of legitimate outlier digits that have not been characterised at all in the training set. This creates a significant challenge for the classifiers. It could be argued that a multi-class classifier could be easily trained with some examples from each of these eight digits but the whole point of these experiments was to mimic the effect that would be caused by actual "unexpected" outliers in a real world scenario, that is, genuine outlier examples that are not known to the classification algorithm at the time of training.

### 4.4.3 Experiment Details

The first experiment was conducted on the primary dataset only with a percentage split of 67% for training and 33% for testing. 100 runs of this experiment were carried out with the dataset being randomly shuffled each time. The results reported are the averages over all of the runs. This mirrors the situation where the testing data are sampled from the same distribution as the training data.

The next experiment introduced the "unexpected" outliers (secondary dataset) in increments of 25 up until and including all 200 of them. This was essentially adding one new "unexpected" digit at a time until they were all present in the test set.

Two one-sided algorithms were used from the OSCAIL toolkit, namely the One-Sided k-Nearest Neighbour and One-Sided k-Means algorithms. Two multi-class classifiers were used from the Weka Machine Learning software. These were the Two-Class Support Vector Machine and the Two-Class k-Nearest Neighbour algorithm.

### 4.4.4 Algorithm Parameters

The main goal was to observe the trends in each algorithm *individually* and not directly compare algorithms against each other. Single values for the parameters that produced a reasonable performance were used. These corresponded to the defaults in Weka.

### 4.4.4.1 OSCAIL Toolkit Parameters:

*One-Sided k-Nearest Neighbour Parameters*:     M = 3,     K = 3,     T = 1.5

M is the number of nearest neighbours to the test example. K is the number of nearest neighbours to the M nearest neighbours. T is the threshold value used to classify the examples. The distance metric used was Euclidean.

*One-Sided k-Means Parameters*:     C=10,     T = 2000

C is the number of clusters in the training examples. T is the threshold value used for classification. The distance metric used was Euclidean.

### 4.4.4.2 Weka 3.5.7 Parameters:

*Two-Class k-Nearest Neighbour Parameters*:     K=1,     Alg = LinearNNSearch

K is the number of neighbours. Alg is the search algorithm used for carrying out the nearest neighbour algorithm search. LinearNNSearch is a simple, brute-force linear search of the test examples which, in this case, used Euclidean distance as the distance metric. The algorithm compares each test example to all of the training examples, in linear fashion, in order to identify the nearest neighbour which is then use to classify the test example.

*Two-Class Support Vector Machine*:     C = 1, Exponent = 1, kernel = polyKernel

C is the complexity parameter. Exponent is the value of the exponent of the polynomial function. The kernel chosen is the polynomial kernel:

$$K(x, y) = <x, y>^p$$

x and y in this formula correspond to two vectors in low dimensional space. p is the exponent value. The < > parentheses indicate the use of the dot product. The dot product takes two vectors, in this case x and y, over the real numbers R and returns a real-valued scalar quantity. The training attributes are normalised by default in this algorithm implementation.

### 4.4.5 Performance Measures

Performance measures are used to indicate how well a classification algorithm has performed in classifying previously unseen examples from the test set. The two measures that were used in this thesis are the overall classification error estimate and the Balanced Error Rate.

### 4.4.5.1 Overall Error Estimate

The estimation of error is recorded for every experiment run. This value is the percentage of instances from the test set that are classified incorrectly. The value is averaged over all of the experiment runs giving the total overall error estimate.

### 4.4.5.2 Balanced Error Rate (BER)

The Balanced Accuracy Rate (BAR) is measured as the average of the true positive rate (sensitivity) and the true negative rate (specificity). See Appendix A.4. The Balanced Error Rate which is what we report in our experiment results, is calculated as one minus the BAR value. This is calculated for each run and the values are then averaged over all of the runs.

### 4.4.6 Results

| Classification Error | One-Sided kNN | One-Sided kMeans | Multi-Class kNN | Multi-Class SVM |
|---|---|---|---|---|
| # Unexpected Outliers | % Error | % Error | % Error | % Error |
| 0 | 9.21 | 9.48 | 0.62 | 1.20 |
| 25 | 7.75 | 7.97 | 2.92 | 3.82 |
| 50 | 6.73 | 7.53 | 7.01 | 8.57 |
| 75 | 5.96 | 6.95 | 11.16 | 11.66 |
| 100 | 5.35 | 7.05 | 14.34 | 13.59 |
| 125 | 4.88 | 6.43 | 16.61 | 15.51 |
| 150 | 4.46 | 5.86 | 17.80 | 16.46 |
| 175 | 4.09 | 5.38 | 18.81 | 17.19 |
| 200 | 3.89 | 5.20 | 19.42 | 17.51 |

**Table 3: Classification error results for all algorithms on digit recognition data**

| Balanced Error Rate | One-Sided kNN | One-Sided kMeans | Multi-Class kNN | Multi-Class SVM |
|---|---|---|---|---|
| # Unexpected Outliers | % BER | % BER | % BER | %BER |
| 0 | 9.21 | 9.48 | 0.62 | 1.20 |
| 25 | 8.72 | 7.46 | 2.74 | 3.62 |
| 50 | 8.48 | 6.83 | 6.15 | 7.57 |
| 75 | 8.32 | 6.24 | 9.27 | 9.78 |
| 100 | 8.21 | 6.21 | 11.40 | 10.94 |
| 125 | 8.14 | 5.72 | 12.74 | 12.06 |
| 150 | 8.07 | 5.31 | 13.26 | 12.43 |
| 175 | 8.00 | 4.98 | 13.65 | 12.66 |
| 200 | 8.01 | 4.85 | 13.79 | 12.63 |

**Table 4: Balanced Error Rate results for all algorithms on digit recognition data**

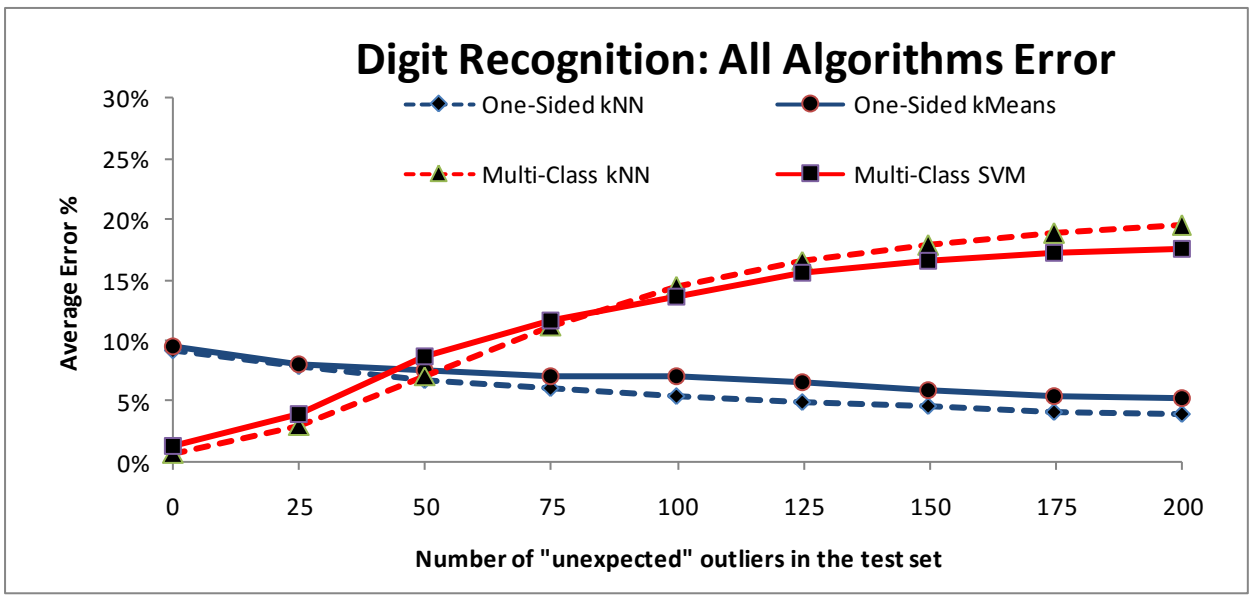

**Figure 23: Classification error when 0 to 200 outliers are present in the test set**

### 4.4.7 Analysis

It is quite clear to see that the multi-class classifiers outperformed the one-sided ones in the first scenario, that is, when there were no "unexpected" outliers in the test set. This was because the decision boundary of the multi-class classifiers had the benefit of being well supported from both sides with representative training examples from each class. In this scenario, the multi-class algorithms essentially had more information to aid the classification mechanism and, therefore, would be expected to out-perform the one-sided approach in such circumstances. An interesting trend occurred, however, when the "unexpected" outliers were added to the test sets. The Identical Distribution assumption no longer held and therefore the error rates originally achieved by the multi-class classifiers began to increase. Conversely, the one-sided classifiers retained a more consistent and decreasing error rate as the outliers were added. This was because they were trained to distinguish the target class from everything else and were therefore more suited to this problem.

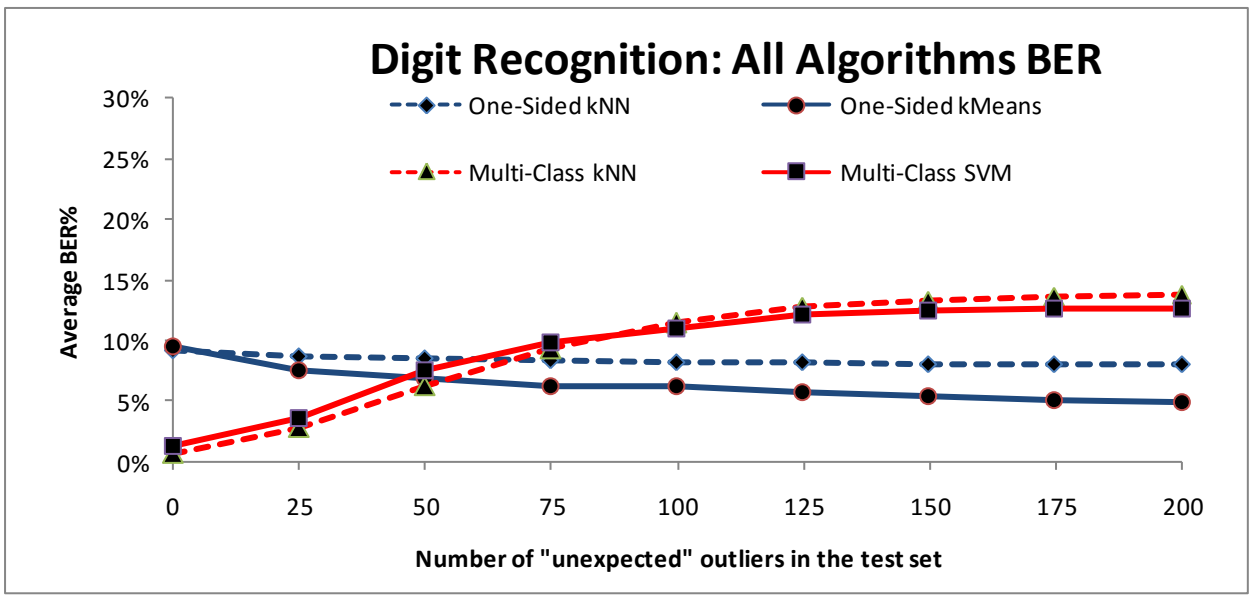

**Figure 24: Balanced Error Rate when 0 to 200 outliers are present in the test set**

From Figure 23 and Figure 24 above, it can be seen that the cross-over in performance of the multi-class and one-sided algorithms occurred after approximately 50 to 75 "unexpected" outliers had been added. This exact amount of outliers needed becomes irrelevant when we consider that none of the classifiers were tuned to maximise performance before the experiments were run. It is the individual trend that each classifier followed that becomes noteworthy. The ability of the one-sided approach to deal with "unexpected" outliers was clearly demonstrated by these results. The classification error of the One-Sided k-NN and k-Means algorithms *decreased* by 5.32% and 4.28% respectively when all 200 outliers had been added to the test set. The multi-class algorithm's error rate *increased* by 18.8% (SVM) and 16.31% (kNN) in the same circumstances.

# 5. Application to the Analysis of Spectroscopy Data

Chapter Five will begin by introducing the problem domain that is of interest to this thesis. The next section will then provide information about the dataset that was used to carry out the experimentation. The procedure followed for each individual experiment will then be described in detail followed by a presentation of the results that were found. The final section will both explain and critically analyse the experimentation results.

## 5.1 A Summary of the Problem Domain

This problem domain is concerned with identifying materials from their respective chemical spectra. The spectral data for the materials can be either gathered in pure form or as a mixture of several different materials. The goal is to identify the presence or absence of a particular material of interest from test example spectrums. What is of particular interest to this thesis is when the distribution of counter-examples does not statistically represent all the possible counter-examples. The separation of chemical laboratory waste is one specific application within this problem domain that will now be described.

### 5.1.1 An Application for Chemical Waste Disposal

The correct procedure of waste disposal is a very important aspect of running a chemical laboratory. Chemical waste that is potentially hazardous to the environment should be identified and disposed of in the correct manner. Laboratories generally have strict guidelines in place as well as legal requirements for such procedures. Organic solvents can create a major disposal problem in organic laboratories as they are usually water immiscible and can be highly flammable (Harwood et al. 1999, Section 1.3). The authors also noted that such solvents are created in abundance each day in busy laboratories. Differentiating between chlorinated and non-chlorinated solvents is of particular importance. Depending on whether a solvent is chlorinated or not will dictate how it is transported from the laboratory and, more importantly, what method is used for its disposal (Conroy *et al.* 2005). Indentifying and labelling such solvents is a routine laboratory procedure which usually makes the disposal a straightforward process. However, Conroy *et al.* (2005) pointed out that the solvents could be accidentally contaminated or inadvertently mislabelled. In such circumstances, it would be beneficial to have an analysis method that would correctly identify whether or not a particular solvent was chlorinated.

## 5.2 Spectroscopy Dataset Used

### 5.2.1 Primary Dataset

The *primary* dataset used for the waste disposal experimentation was collected by Jennifer Conroy, with the guidance of her supervisor Dr. Alan Ryder of the School of Chemistry at the National University of Ireland, Galway (NUI Galway). This was carried out during an earlier research project led by Dr. Michael Madden of the College of Engineering and Informatics, NUI Galway, in collaboration with Dr. Ryder. The sources of information that provide details about this dataset are Conroy *et al.* (2005) and Howley (2007, Section A.1.2). The authors state that the dataset comprises 230 spectral samples that contain both chlorinated and non-chlorinated mixtures. The compilation of the data involved keeping the concentrations of the mixtures as close as possible to real life scenarios from industrial laboratories. Twenty-five solvents, some chlorinated and some not, were included and these are listed in Table 5 below. Table 6 contains a summary of the mixtures.

| Solvents | Grade | Solvents | Grade |
|---|---|---|---|
| Acetone | *HPLC** | Acetophenol* | *Analytical* |
| Toluene | *Spectroscopic* | n-Pentane | *Analytical* |
| Cyclohexane | *Analytical & Spectroscopic* | Xylene | *Analytical* |
| Acetonitrile | *Spectroscopic* | Nitromethane | *Analytical* |
| 2-Propanol | *Spectroscopic* | Dimethylformanide | *Analytical* |
| 1,4-Dioxane | *Analytical & Spectroscopic* | NitroBenzene* | *Analytical* |
| Hexane | *Analytical* | Tetrahydrofuran | *Analytical* |
| 1-Butanol | *Analytical & Spectroscopic* | Diethyl Ether | *Analytical* |
| Methyl Alcohol | *Analytical* | Petroleum Acetate | *Analytical* |
| Benzene | *Analytical* | Chlorform | *Analytical & Spectroscopic* |
| Ethyl Acetate | *Analytical* | Dichloromethane | *Analytical & Spectroscopic* |
| Ethanol | *Analytical* | 1,1,1-Trichloroethane | *Analytical & Spectroscopic* |
| Cyclopentane | *Analytical* | | |

**Table 5: Chlorinated and non-chlorinated solvents used and their respective grades**

*****Solvents containing fluorescent impurities**

******High Performance Liquid Chromatography**

**(Source: Conroy et al. 2005)**

|  | Chlorinated | Non-chlorinated | Total |
|---|---|---|---|
| Pure Solvents | 6 | 24 | 30 |
| Binary Mixtures | 96 | 23 | 119 |
| Ternary Mixtures | 40 | 12 | 52 |
| Quaternary Mixtures | 12 | 10 | 22 |
| Quintary Mixtures | 0 | 7 | 7 |
| **Total** | 154 | 76 | 230 |

**Table 6: A summary of the chlorinated and non-chlorinated solvent mixtures used (Source: Howley, 2007)**

### 5.2.2 Secondary Dataset

For our Scenario 2 experiments (see Section 5.3), we introduced 50 additional spectra that represented outliers that were taken from a different distribution to those that were present in the primary dataset. These samples were the Raman spectra of various laboratory chemicals, and none of them were chlorinated solvents nor were they the other materials that are listed in Table 5 above. They included materials such as sugars, salts and acids in solid or liquid state, including Sucrose, Sodium, Sorbitol, Sodium Chloride, Pimelic Acid, Acetic Acid, Phthalic Acid and Quinine.

### 5.2.3 Preparing the Datasets for the Toolkit

Four variants of the primary dataset were created, which differed only by the labelling of the solvent that was currently assigned as the target class.

1. *Chlorinated Or Not*: All chlorinated solvents are labelled as targets.
2. *Chloroform*: Only examples containing Chloroform are labelled as targets.
3. *Dichloromethane*: Only examples containing Dichloromethane labelled as targets.
4. *Trichloroethane*: Only examples containing Trichloroethane are labelled as targets

In each of these variants, all instances not labelled as targets were labelled as outliers. The most important dataset created was for carrying out experiments to detect whether a given mixture was chlorinated or not. For this dataset, all of the chlorinated solvents were labelled as targets and everything else was labelled as other. A further three datasets were created for the detection of the specific chlorinated solvents: *Chloroform*, *Dichloromethane*, and *Trichloroethane*.

## 5.3 Experiment Overview

These experiments were designed to assess the robustness of both one-sided and multi-class classifiers in a situation in which the test sets were exposed to outliers that were not drawn from the same distribution as training outliers. This involved analysing the effect of adding out-of-sample examples to the test sets of one-sided and multi-class classifiers. The OSCAIL toolkit and Weka (version 3.5.7) were used for conducting these experiments. The following algorithms were used:

Multi-Class Classification Algorithms (from Weka):
- Two-Class k-Nearest Neighbour
- Two-Class Support Vector Machine

One-Sided Classification Algorithms (from the OSCAIL toolkit):
- One-Sided k-Nearest Neighbour
- One-Sided k-Means
- One-Sided Support Vector Machine

Two different scenarios were used in these experiments.

*Scenario 1: "Expected" Test Data Only.* In this scenario the test data were sampled from the same distribution as the training data. The primary dataset (Section 5.2.1) was divided repeatedly into training sets and test sets, with the proportions of targets and outliers held constant at all times, and these internal test sets were used to test the classifiers that were built with the training datasets.

*Scenario 2: "Unexpected" and "Expected" Test Data.* In this scenario, the dataset was augmented with the 50 examples of a secondary dataset (Section 5.2.2) that were not drawn from the same distribution as the training dataset. Therefore, a classifier trained to recognise any chlorinated solvent should reject these as outliers. However, these samples represented a significant challenge to the classifiers, since they violated the standard assumption that the test data would be drawn from the same distribution as the training data: It is for this reason that they are termed "unexpected". These outliers were introduced to the test sets in increments of 10 up to, and including, 50. This was done to examine the trends that emerged as a result of the addition of these outliers.

### 5.3.1 Splitting into Training and Testing

For each experiment, 100 runs were carried out with the data being randomly split each time into 67% for training and 33% for testing. The splitting procedure from our toolkit ensured that there was the same proportion of targets and outliers in the training sets as there was in the test sets. The same dataset splits were used for the one-class classifier algorithms and the Weka-based algorithms to facilitate direct comparisons. For each separate run, the data were shuffled by using an incremented random number seed.

### 5.3.2 Parameter Selection

Good parameter selection plays an important role in determining how well a classification algorithm will perform. Different datasets require different sets of parameters in order to optimise the classification performance. In these particular experiments, we are not concerned with tuning parameters and identifying the best algorithms based on their overall performance. Our goal was to analyse the trend that occurred when "unexpected" outliers were added to the test sets. For this reason, a parameter selection step was omitted from the experimentation procedure. Single values for the algorithm parameters were used and these values were the same for all variants (which are explained in Section 5.2.3) of the dataset.

#### 5.3.2.1 OSCAIL Toolkit Parameters

*One-Sided k-Nearest Neighbour Parameters*:     M = 1,     K = 3,     T = 1.5

M is the number of nearest neighbours to the test example. K is the number of nearest neighbours to the M nearest neighbours. T is the threshold value used to classify the examples. The distance metric used was Euclidean.

*One-Sided k-Means Parameters*:     C=14,     T = 1.5

C is the number of clusters in the training examples. T is the threshold value used for classification. The distance metric used was Euclidean.

*One-Sided Support Vector Machine*     S=1,     N=0.01,     kernel = Gaussian

S is the kernel width and N is regularisation parameter for the SVM. The Gaussian kernel was used which is defined as:

$$K(x, y) = \exp\left(-\frac{\|x - y\|^2}{2\sigma^2}\right)$$

exp(·) is the exponential function. ‖ · ‖ is the Euclidean norm for the vectors and $\sigma$ is the kernel parameter.

#### 5.3.2.2 Weka Parameters

*Two-Class k-Nearest Neighbour Parameters*:   K=1,   Alg = LinearNNSearch

K is the amount of neighbours. Alg is the search algorithm used for carrying out the nearest neighbour algorithm search. LinearNNSearch is a simple, brute-force linear search of the test examples which, in this case, used Euclidean distance as the distance metric. The algorithm compares each test example to all of the training examples, in linear fashion, in order to identify the nearest neighbour which is then use to classify the test example.

*Two-Class Support Vector Machine*:   C = 1, Exponent = 1, kernel = polyKernel

C is the complexity parameter. Exponent is the exponent of the polynomial function. The kernel chosen is the polynomial kernel:

$$K(x, y) = <x, y>^p$$

x and y in this formula correspond to two vectors in low dimensional space. p is the exponent value. The < > parentheses indicate the use of the dot product. The dot product takes two vectors, in this case x and y, over the real numbers R and returns a real-valued scalar quantity. Training attributes are normalised by default in this implementation.

### 5.3.3   Normalisation of the Data

A common method for normalising a dataset is to recalculate the values of the attributes to fall within the range of zero to one. This is usually carried out on an attribute-by-attribute basis and ensures that certain attribute values, which differ radically in size from the rest, do not dominate the prediction calculations. The normalisation carried out on the spectral data is different to this in that it is carried out on an instance-by-instance basis. This means that the attributes in each instance are normalised based only on the remaining attributes in that instance and not the entire dataset. Considering that every attribute of a spectral instance is a point on the spectrum, this process is essentially rescaling the spectrum into the range of zero to one.

### 5.3.4   Individual Experiments

The main experiment involved detecting the presence of chlorinated solvents from examples that were either in pure form or that contained several solvents as part of a mixture. The training sets contained some examples of chlorinated mixtures labelled as target. There were also non-chlorinated examples labelled as other in the training set. Once the classifiers were trained, they were then presented with an unlabelled test set containing both chlorinated and non-chlorinated examples and the goal was to correctly identify their labels. Further experiments attempted to detect the individual chlorinated compounds of *Chloroform*, *Dichloromethane*, and *Trichloroethane*. This was a more challenging task, considering that some of the sample mixtures could contain one, two or even all three of these compounds. For example, when *Chloroform* is the target class, a target example could consist of both *Chloroform* and *Trichloroethane* in a mixture. However, pure *Trichloroethane* would still be labelled as an outlier as it would not contain any *Chloroform*. It is certainly a difficult task for the classifier to distinguish between a target and an outlier example when both of them contain the same compound (*Trichloroethane* from the example above). Perhaps this could also be said of the *Chlorinated or Not* dataset; however, the proportions of non-chlorinated compounds in the target chlorinated mixtures that reappear as part of some outlier examples would be much smaller.

## 5.4 Results

This section will provide tabled results, both average classification error and Balanced Error Rate, explained in Section 4.4.5, for all of the algorithms considered. The corresponding graphs for these tables will also be provided to observe the trends in performance when the "unexpected" outliers were added to the test sets. Each of the algorithm results will first be discussed individually and then there will be a discussion of the overall results later in Section 5.5. An explanation of the layout of the tabled and graphed results will be provided in the next section (Two-Class k-Nearest Neighbour Algorithm Results). This layout applies to the tables and graphs in all of the subsequent result sections.

### 5.4.1 Two-Class k-Nearest Neighbour Algorithm Results

The first table of results, Table 7 below, represents the overall classification error (averaged from one hundred runs) for each of the four variants of the dataset that were previously discussed. The #UO column, on the left hand side of the table, corresponds to the number of "unexpected" outliers that are currently in the test set. The amount of "unexpected" outliers was added in increments of ten to the test set. Classification results for each variant of the dataset are represented in the corresponding columns, from left to right, that are adjacent to the amount of "unexpected" outliers.

| #UO | Chlorinated or Not | Chloroform | Dichloromethane | Trichloroethane |
|---|---|---|---|---|
| 0 | 7.48 | 21.34 | 11.66 | 21.90 |
| 10 | 11.24 | 23.35 | 12.02 | 21.26 |
| 20 | 15.32 | 26.09 | 12.39 | 22.04 |
| 30 | 19.03 | 28.79 | 12.66 | 22.22 |
| 40 | 22.30 | 31.02 | 13.01 | 22.38 |
| 50 | 25.35 | 32.89 | 13.46 | 22.99 |

**Table 7: Classification error results of Two-Class k-Nearest Neighbour algorithm on the four variants of the Chlorinated Solvents dataset, with varying numbers of unexpected outliers.**

The first of the two-class algorithms that were tested was the Two-Class k-Nearest algorithm. We can see from both Table 7 and Table 8 that the performances of the algorithm on the *Chlorinated or Not* and *Dichloromethane* datasets were quite good in the first scenario. However, the error rates begin to increase as the "unexpected" outliers are added. The most severe dip in performance can be seen in both the *Chlorinated or Not* and the *Chloroform* dataset variants.

The error rates for *Dichloromethane* and *Trichloroethane* did increase somewhat when all 50 of the "unexpected" outliers had been added. However, these results were much more consistent than the other two variants of the dataset, whose classification performance became erratic.

| #UO | Chlorinated or Not | Chloroform | Dichloromethane | Trichloroethane |
|---|---|---|---|---|
| 0 | 10.61 | 23.14 | 13.74 | 23.35 |
| 10 | 14.57 | 24.62 | 14.02 | 22.96 |
| 20 | 18.29 | 26.90 | 13.73 | 23.60 |
| 30 | 21.22 | 29.10 | 13.66 | 23.65 |
| 40 | 23.45 | 30.85 | 13.73 | 23.69 |
| 50 | 25.35 | 32.28 | 13.93 | 24.07 |

**Table 8: Balanced Error Rate results of Two-Class k-Nearest Neighbour algorithm on the four variants of the Chlorinated Solvents dataset, with varying numbers of unexpected outliers.**

Table 8 differs from the previous table (Table 7) in that it presents the Balanced Error Rate results that were achieved by the algorithm. This performance measure takes into consideration the sensitivity and specificity of the classification results. From both tables, we can see that the same trend in performance occurs for each of the variant datasets.

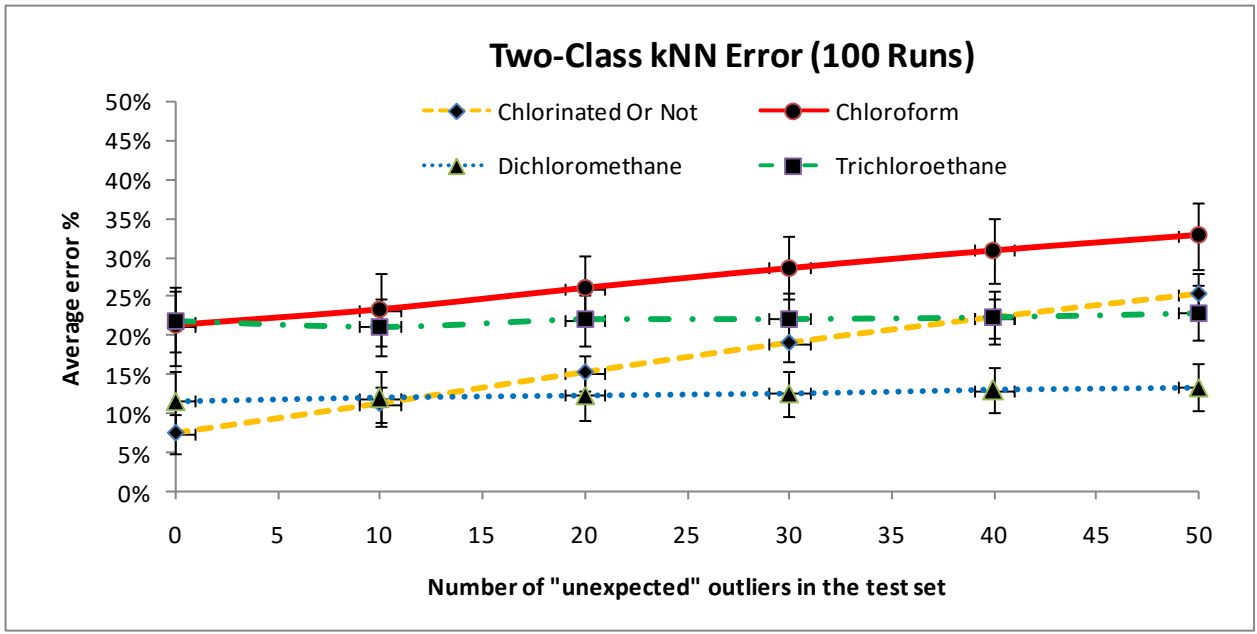

**Figure 25: Error graph for Two-Class k-Nearest Neighbour algorithm**

Figure 25, above, shows a graphical representation of the classification error results for the Two-Class k-Nearest Neighbour algorithm (from Table 7). The number of "unexpected" outliers present in the test set is plotted against the percentage error result achieved by the algorithm on each of the variant datasets. The vertical whiskers on the graph indicate the standard deviation from the error at each point.

Figure 26 shows a graphical presentation of the Balanced Error Rate for the Two-Class k-Nearest Neighbour algorithm. Once again, the data that is plotted in this graph has been taken from the corresponding table (Table 8) for the Balanced Error Rate results. The graphing of such data facilitates a straightforward observation of the trends in performance that have taken place.

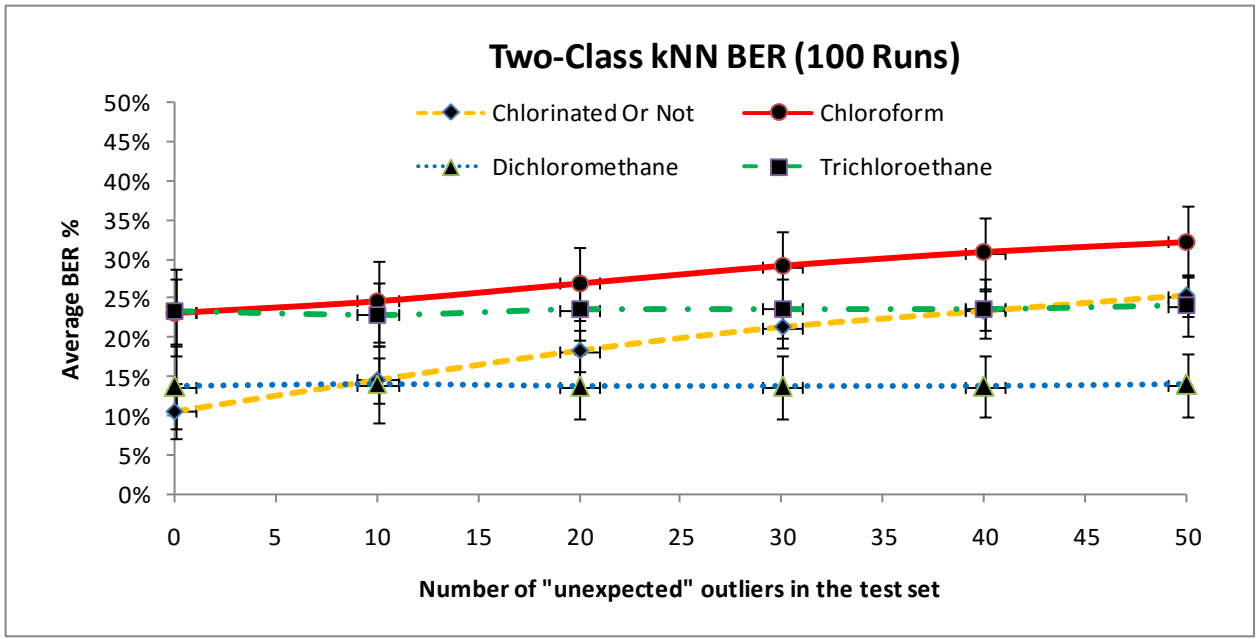

Figure 26: Balanced Error Rate graph for Two-Class k-Nearest Neighbour algorithm

We can see from both Figure 25 and Figure 26 that the performance of Two-Class k-NN on the *Chlorinated or Not* dataset became much worse when all of the "unexpected" outliers were added. In fact, it goes from having the best results of all four (in the first scenario) to having the second worst results overall in Scenario 2. The *Chloroform* variant shows a similar trend whereas the other two retain more consistent error rates.

### 5.4.2 Two-Class Support Vector Machine Algorithm Results

The Two-Class Support Vector Machine achieved the best results, in Scenario 1, than any of the other algorithms that were tested (one-sided and two-class). Good performances can be seen right across both Table 9 and Table 10, below, when there are no "unexpected" outliers present in the test set.

| #UO | Chlorinated or Not | Chloroform | Dichloromethane | Trichloroethane |
|---|---|---|---|---|
| 0  | 5.65  | 7.26  | 6.49 | 6.77  |
| 10 | 11.17 | 9.30  | 6.45 | 10.66 |
| 20 | 16.08 | 10.18 | 6.77 | 15.10 |
| 30 | 20.34 | 10.49 | 6.97 | 18.91 |
| 40 | 24.26 | 10.58 | 7.10 | 22.44 |
| 50 | 27.86 | 10.81 | 7.18 | 25.82 |

Table 9: Classification error of the Two-Class Support Vector Machine on the four variants of the Chlorinated Solvents dataset, with varying numbers of unexpected outliers.

Similar to the Two-Class k-NN, however, the performance on the *Chlorinated or Not* dataset became severely degraded following the introduction of the "unexpected" outliers. The same trend in performance, that is, a large increase in the error rates, can be observed in the *Trichloroethane* dataset also. The change in error rate is not as severe in the other two variants of the dataset but it is still evidently becoming worse.

| #UO | Chlorinated or Not | Chloroform | Dichloromethane | Trichloroethane |
|---|---|---|---|---|
| 0 | 7.89 | 9.24 | 7.76 | 8.58 |
| 10 | 14.46 | 10.98 | 7.78 | 11.68 |
| 20 | 19.23 | 11.75 | 8.04 | 14.98 |
| 30 | 22.70 | 12.07 | 8.21 | 17.67 |
| 40 | 25.53 | 12.21 | 8.32 | 20.05 |
| 50 | 27.86 | 12.44 | 8.40 | 22.25 |

**Table 10: Classification Balanced Error Rate of the Two-Class Support Vector Machine on the four variants of the Chlorinated Solvents dataset, with varying numbers of unexpected outliers.**

Once again, we can observe the same trend of performance in the table of Balanced Error Rate results, above. The largest increase of the error rate is seen with the *Chlorinated or Not* data which rises by almost 20 % after all of the "unexpected" outliers have been added.

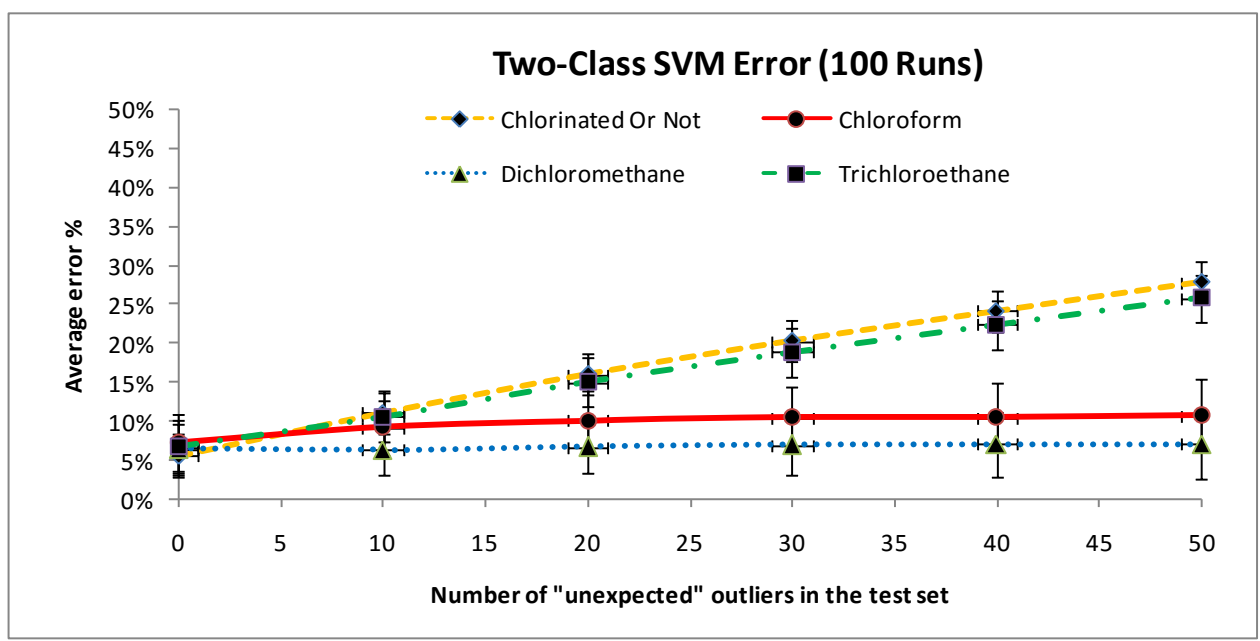

**Figure 27: Error graph for Two-Class Support Vector Machine**

Figure 27 and Figure 28 illustrate the trends that occur from the results that are listed in Table 9 and Table 10. We can observe that the error rates for both *Chloroform* and *Dichlormethane* remain consistent across the graph where as the other two get progressively worse as the outliers are added.

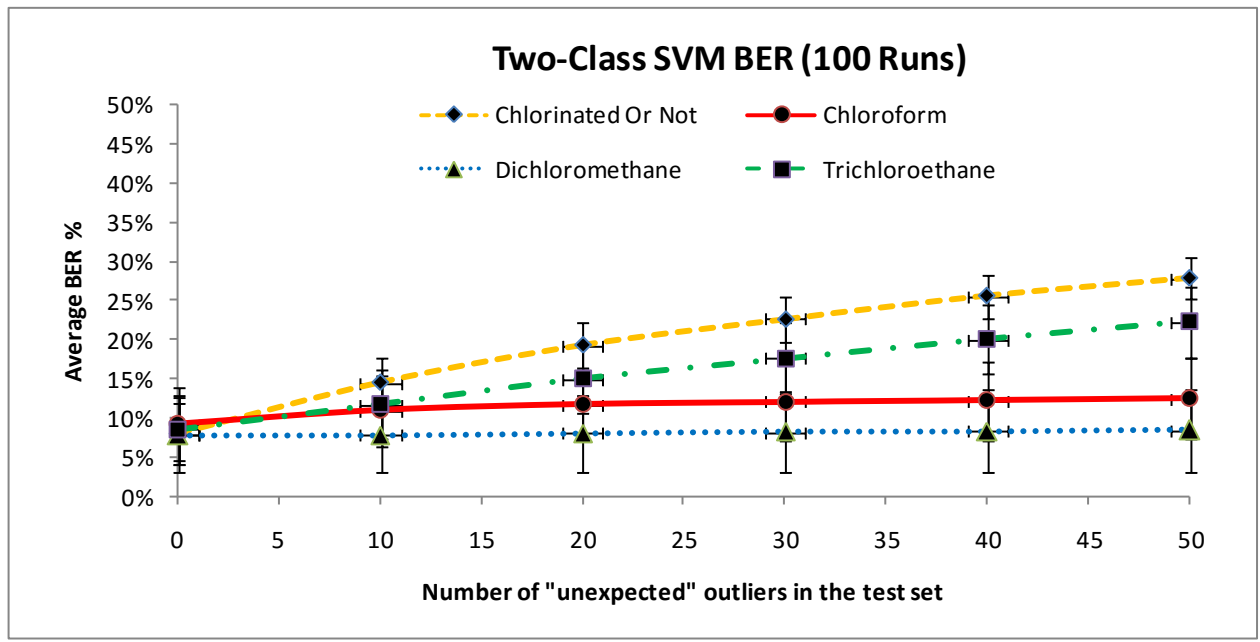

**Figure 28: Balanced Error Rate graph for Two-Class Support Vector Machine**

### 5.4.3  One-Sided k-Nearest Neighbour Algorithm Results

The first of the one-sided classification algorithms to be tested was the One-Sided k-Nearest Neighbour algorithm. We can observe that the classification error decreases, in all of the variant cases from Table 11 below, as the number of "unexpected" outliers are increased in the test set.

| #UO | Chlorinated or Not | Chloroform | Dichloromethane | Trichloroethane |
|---|---|---|---|---|
| 0  | 9.47 | 33.38 | 18.81 | 27.17 |
| 10 | 8.39 | 29.64 | 16.66 | 23.24 |
| 20 | 7.53 | 26.67 | 14.94 | 20.44 |
| 30 | 6.82 | 24.19 | 13.54 | 18.29 |
| 40 | 6.63 | 22.13 | 12.74 | 16.61 |
| 50 | 6.88 | 21.20 | 11.74 | 15.79 |

**Table 11: Classification error results of One-Sided k-Nearest Neighbour algorithm on the four variants of the Chlorinated Solvents dataset, with varying numbers of unexpected outliers.**

The first scenario, in which none of these outliers are present, clearly produces somewhat poor results. This is especially true when trying to detect the *Chloroform* compound from the test examples. Detecting the individual compounds appears to be a more difficult task than just predicting whether or not a test sample is chlorinated. A positive aspect which can be taken from these results is the fact that the classifier robustly rejected the out-of-sample examples as they were introduced.

| #UO | Chlorinated or Not | Chloroform | Dichloromethane | Trichloroethane |
|---|---|---|---|---|
| 0 | 10.10 | 27.17 | 14.32 | 22.84 |
| 10 | 8.44 | 23.24 | 12.57 | 19.35 |
| 20 | 7.49 | 20.44 | 11.26 | 16.89 |
| 30 | 6.88 | 18.29 | 10.25 | 15.04 |
| 40 | 6.81 | 16.61 | 9.67 | 13.58 |
| 50 | 7.09 | 15.79 | 9.00 | 12.41 |

Table 12: Balanced Error Rate results of One-Sided k-Nearest Neighbour algorithm on the four variants of the Chlorinated Solvents dataset, with varying numbers of unexpected outliers.

The Balanced Error Rate results for this algorithm are shown in Table 12, above. Although the figures achieved are somewhat different, the same promising trend in performance can be seen across the table. In all cases, a robust rejection of the "unexpected" outliers is evident.

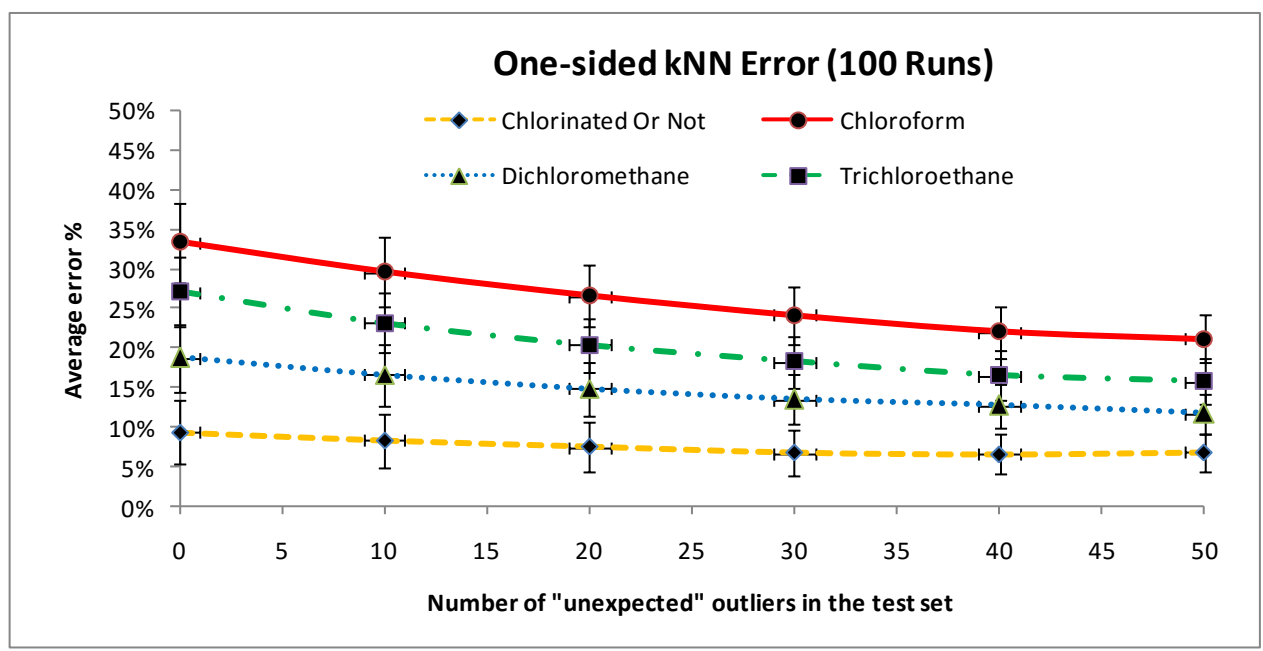

Figure 29: Error graph for One-Sided k-Nearest Neighbour algorithm

From Figure 29 and Figure 30, we can observe that the same trend has occurred for all of the variants of the dataset when using the One-Sided k-Nearest Neighbour algorithm. That is, the error rate was reduced for each of them, in some cases quite significantly. It is also notable that the performances of the four variants retain the same ranking after all of the "unexpected" outliers have been added. This is a direct result of them all following the same trend.

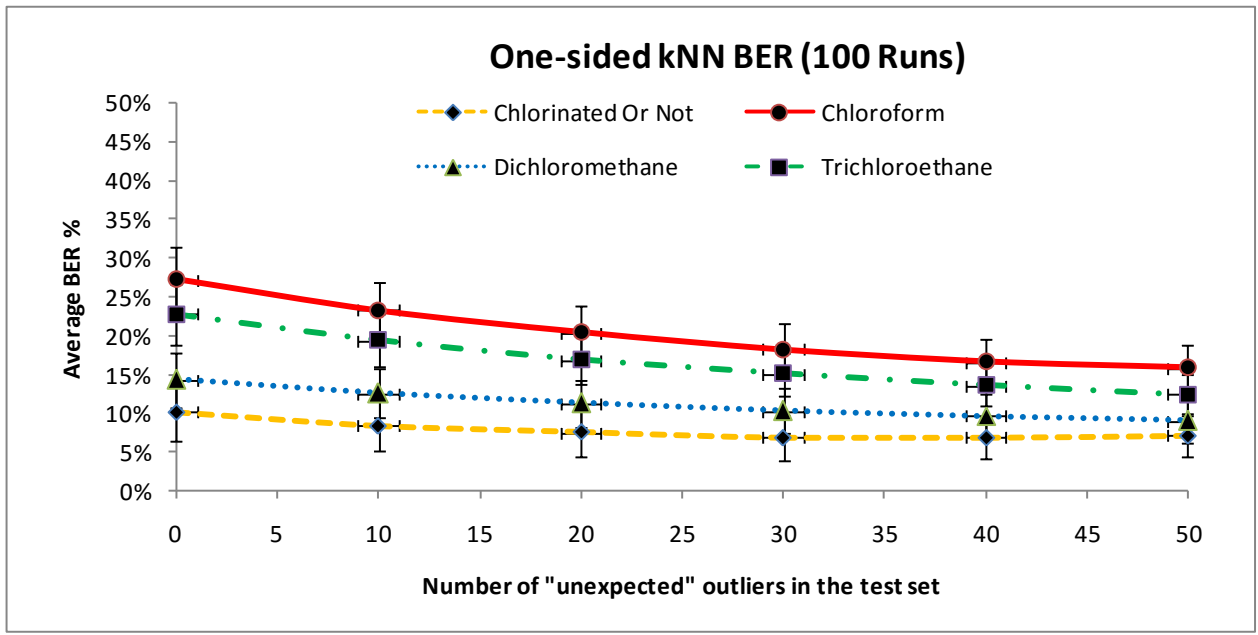

**Figure 30: Balanced Error Rate graph for One-Sided k-Nearest Neighbour algorithm**

### 5.4.4 One-Sided k-Means Algorithm Results

The first noticeable aspect of the k-Means algorithm results, from Table 13 and Table 14 below, is that the performance in the first scenario was certainly quite poor. The use of the same parameters for each variant of the dataset is the most-likely cause for this. Tuning the parameters to the specific tasks would provide better results but we are only concerned with performance trends.

| #UO | Chlorinated or Not | Chloroform | Dichloromethane | Trichloroethane |
|---|---|---|---|---|
| 0 | 25.17 | 45.60 | 37.73 | 44.45 |
| 10 | 22.39 | 40.47 | 33.39 | 39.46 |
| 20 | 20.19 | 36.40 | 29.95 | 35.49 |
| 30 | 18.39 | 33.09 | 27.15 | 32.27 |
| 40 | 16.91 | 30.35 | 24.83 | 29.60 |
| 50 | 15.65 | 28.04 | 22.87 | 27.35 |

Table 13: Classification error results of One-Sided k-Means algorithm on the four variants of the Chlorinated Solvents dataset, with varying numbers of "unexpected" outliers.

As the "unexpected" outliers were added to the test sets, both the classification error and Balanced Error Rate began to reduce for all of the datasets. This demonstrated the ability of the classifier to reject these outliers.

| #UO | Chlorinated or Not | Chloroform | Dichloromethane | Trichloroethane |
|---|---|---|---|---|
| 0 | 30.40 | 37.86 | 27.01 | 35.99 |
| 10 | 24.08 | 32.63 | 23.33 | 30.72 |
| 20 | 20.51 | 28.90 | 20.60 | 26.95 |
| 30 | 18.21 | 26.09 | 18.51 | 24.12 |
| 40 | 16.61 | 23.91 | 16.84 | 21.92 |
| 50 | 15.43 | 22.17 | 15.49 | 20.16 |

Table 14: Balanced Error Rate results of One-Sided k-Means algorithm on the four variants of the Chlorinated Solvents dataset, with varying numbers of "unexpected" outliers.

This trend in performance is similar to that which was observed for the One-Sided k-Nearest Neighbour algorithm above.

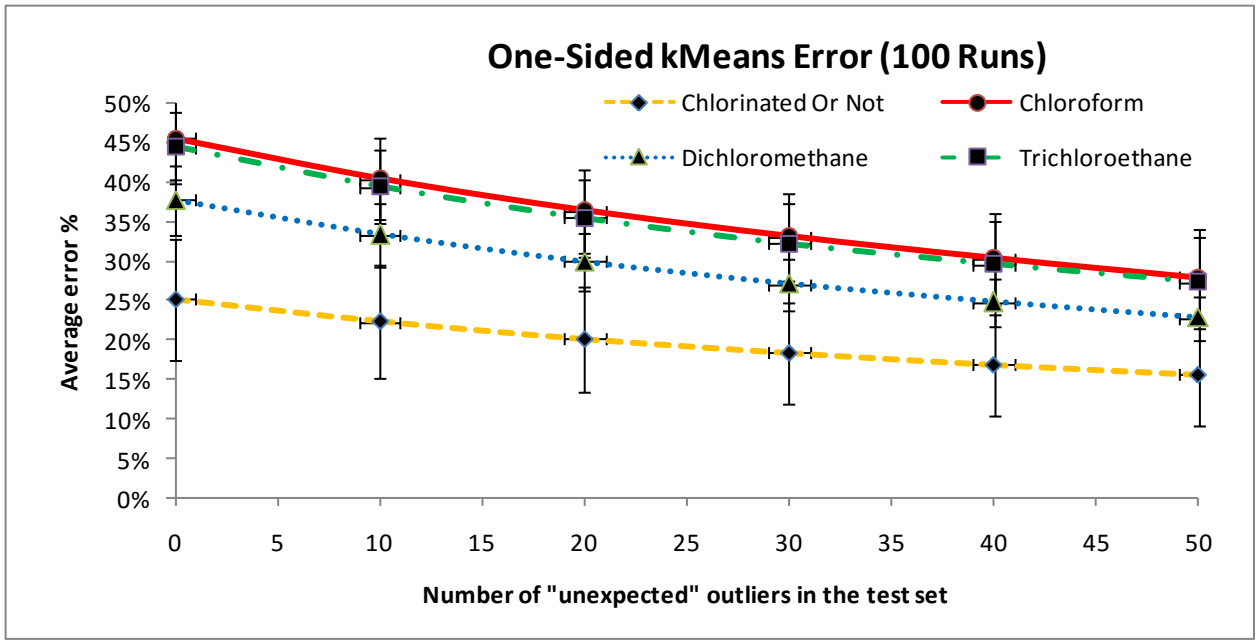

Figure 31: Error graph for One-Sided k-Means algorithm

We can observe from Figure 31 and Figure 32 that the standard deviation of the results can be quite large at times. This is most notable in the "Chlorinated or not" classification error results in Figure 31 above and shows how great the variance in results can be over the one hundred runs. The error rate for all of the variants of the dataset becomes considerably reduced as a result of the introduction of the "unexpected" outliers to the test set.

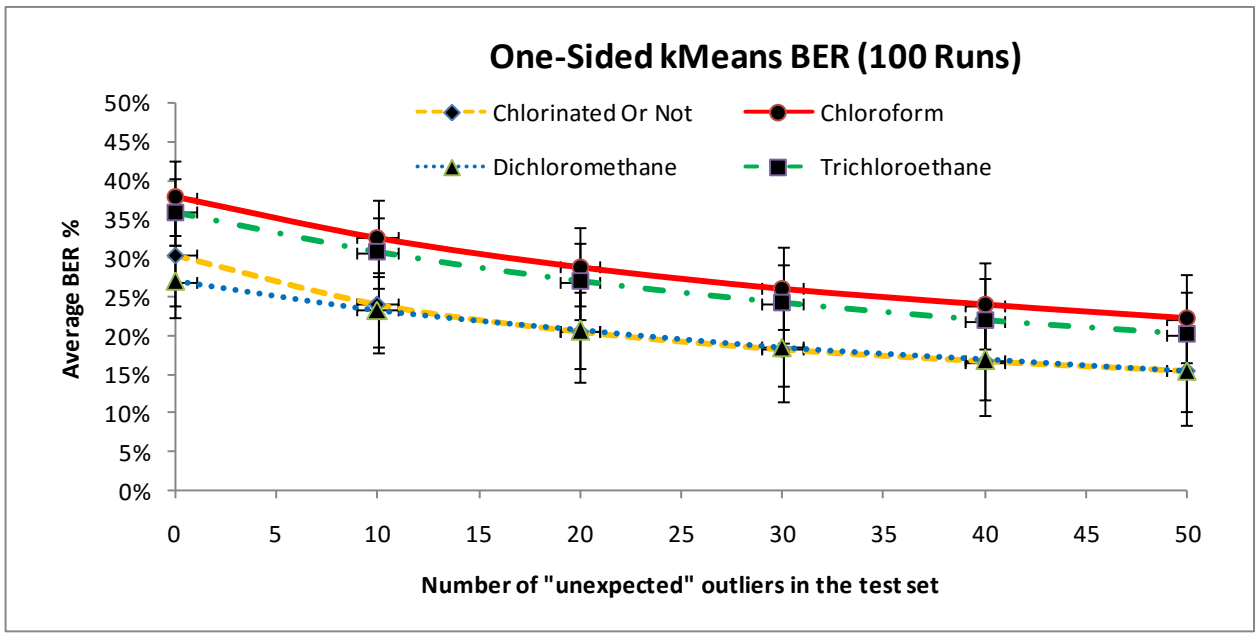

Figure 32: Balanced Error Rate graph for One-Sided k-Means algorithm

### 5.4.5 One-Sided Support Vector Machine Algorithm Results

The results obtained for the One-Sided Support Vector Machine were undoubtedly disappointing. The trend in performance was very different to the one-sided algorithms that have been discussed above. The One-Sided Support Vector Machine failed to robustly deal with the "unexpected" outliers as they were added to the test sets. This is most evident from the Balanced Error Rate results from Table 16 below. From this table, we can see that the results were very poor in Scenario 1 and become even worse when all fifty of the "unexpected" outliers had been added.

| #UO | Chlorinated or Not | Chloroform | Dichloromethane | Trichloroethane |
|---|---|---|---|---|
| 0 | 19.18 | 32.92 | 17.84 | 29.71 |
| 10 | 27.92 | 29.48 | 16.11 | 30.78 |
| 20 | 35.24 | 26.89 | 15.39 | 29.86 |
| 30 | 41.01 | 24.75 | 17.25 | 27.32 |
| 40 | 45.76 | 23.13 | 16.75 | 25.03 |
| 50 | 49.84 | 22.54 | 17.41 | 23.95 |

**Table 15: Classification error results of One-Sided Support Vector Machine algorithm on the four variants of the Chlorinated Solvents dataset, with varying numbers of unexpected outliers**

These unsatisfactory results could be due to a variety of factors. For instance, parameter selection has a major influence over the performance of One-Sided Support Vector Machines. Since we were only concerned with the trend in performance, an extensive parameter search did not take place and all variants of the dataset used the same parameters.

| #UO | Chlorinated or Not | Chloroform | Dichloromethane | Trichloroethane |
|---|---|---|---|---|
| 0 | 25.35 | 45.62 | 32.44 | 39.01 |
| 10 | 32.41 | 45.62 | 32.49 | 41.60 |
| 20 | 36.81 | 45.70 | 32.94 | 42.22 |
| 30 | 39.46 | 45.75 | 34.79 | 41.49 |
| 40 | 41.28 | 45.89 | 35.03 | 40.81 |
| 50 | 42.69 | 46.48 | 35.86 | 40.80 |

**Table 16: Balanced Error Rate results of One-Sided Support Vector Machine algorithm on the four variants of the Chlorinated Solvents dataset, with varying numbers of unexpected outliers**

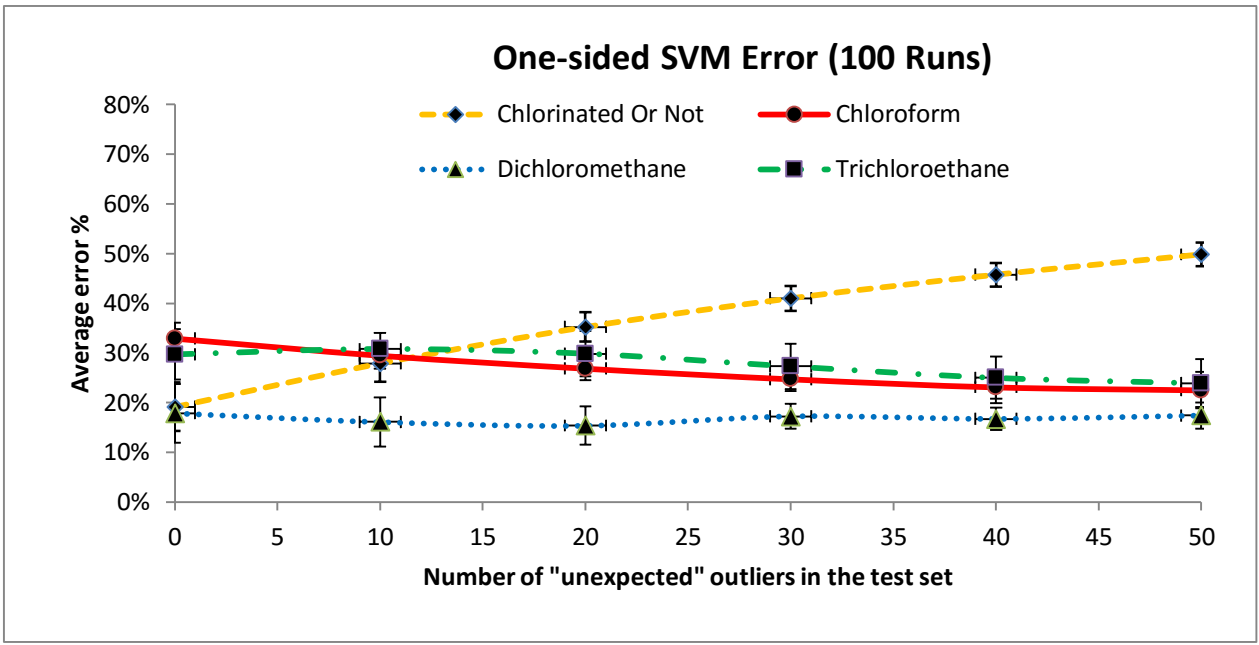

**Figure 33: Error graph for One-Sided Support Vector Machine**

The reader should note that the Y-axes in both Figure 33 and Figure 34 have been rescaled to fall between zero and eighty percent (all other graphs are between zero and fifty percent). The purpose of this rescaling is to centralise the data shown in the graphs. From Figure 33, above, we can see the abrupt rise in classification error when using the "Chlorinated or not" dataset. All of the other variants remain somewhat consistent; however their overall performances remain poor.

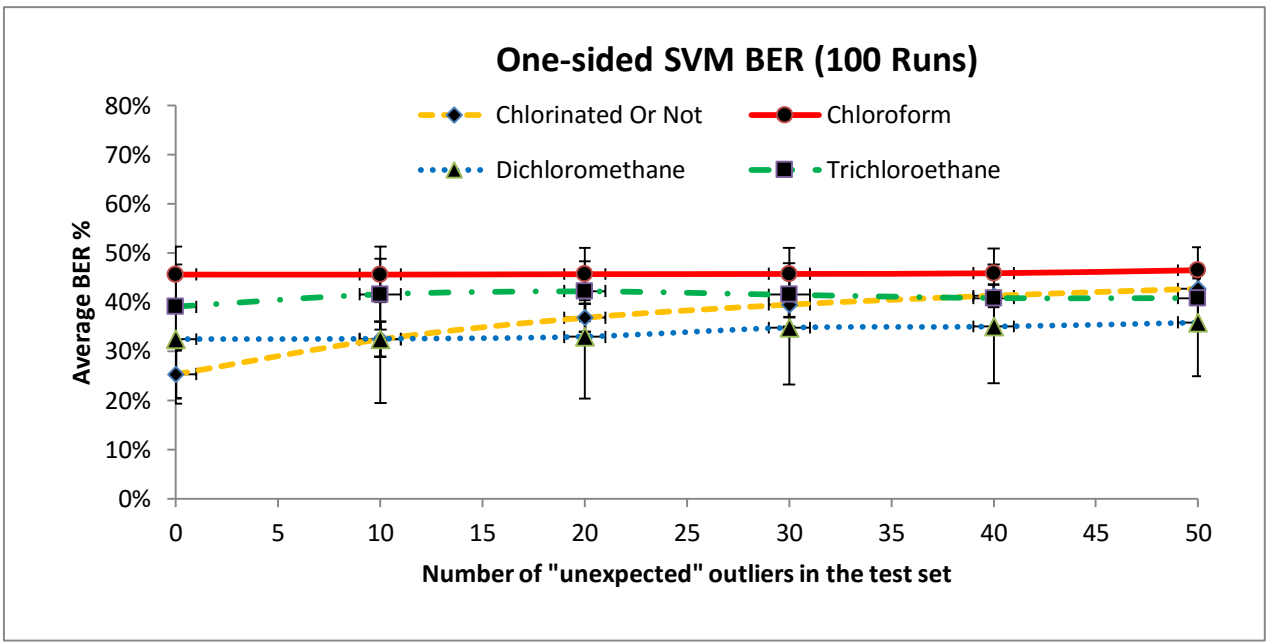

**Figure 34: Balanced Error Rate graph for One-Sided Support Vector Machine**

## 5.5 Analysis of Experiments

Upon examination of these results, it can be observed that while the conventional multi-class classifiers perform quite well in the first scenario (no "unexpected" outliers); their performance begins to deteriorate once the outliers have been introduced in Scenario 2. Given the limitations of the multi-class approach (discussed earlier in Section 2.2.2), these results were expected. The Two-Class kNN and Two-Class SVM algorithms had a tendency to perform well in the "closed set" scenario because they had adequate training data to help them make an informed classification prediction. As the "unexpected" outliers were introduced, both algorithms no longer had a reliable means to aid their classification mechanism which led to the unpredictable and often poor results. Conversely, the one-sided algorithms are trained to distinguish the target class from *everything else* and not just the negative examples in the training set. In fact, negative training examples are ignored by some one-sided classification algorithms so whether or not they are statistically representative, in this case, is irrelevant.

An interesting question arises from observing the one-sided classification experiment results: Why does the overall error rate decrease as opposed to just holding constant? Our explanation is that one-sided classifiers usually have the ability to robustly reject "unexpected" outliers, that is, examples which differ significantly from the target class. Outlier examples that are expected to be encountered in the training set, which we can term "expected" outliers, can often be quite similar to the target class which can lead to a greater number of misclassifications with one-sided classifiers. It is for this reason that the overall error rate decreases as these easier to classify examples ("unexpected" outliers) begin to form a larger part of the test set.

In detecting whether or not a sample is chlorinated, the average error rate of the Two-Class kNN increased by 17.87 % (see Table 7) and the Two-Class SVM increased by 22.21 % (see Table 9) in Scenario 2 when all 50 outliers were added. In contrast with the multi-class classifiers, the One-Sided kNN is seen to retain a consistent performance with the error rate actually decreasing by 2.59 % (see Table 11). While the performance of One-Sided k-Means is quite poor in the first scenario, the classification error rate is decreased by 9.52 % (see Table 13) when all of the "unexpected" outliers have been added. When each algorithm is detecting the individual solvents, the same pattern in performance can generally be seen. The multi-class algorithms error rate increases, in some cases quite radically, in the second scenario. The One-Sided kNN and One-Sided k-Means manage to remain at a more consistent error rate, and in all cases the overall error rate is reduced somewhat. Such results were expected to be observed from the One-Sided Support Vector Machine; however, this was not the case.

Figure 35 and Figure 36 are two graphs which show the *Chlorinated or Not* experiment results plotted together for all of the algorithms.

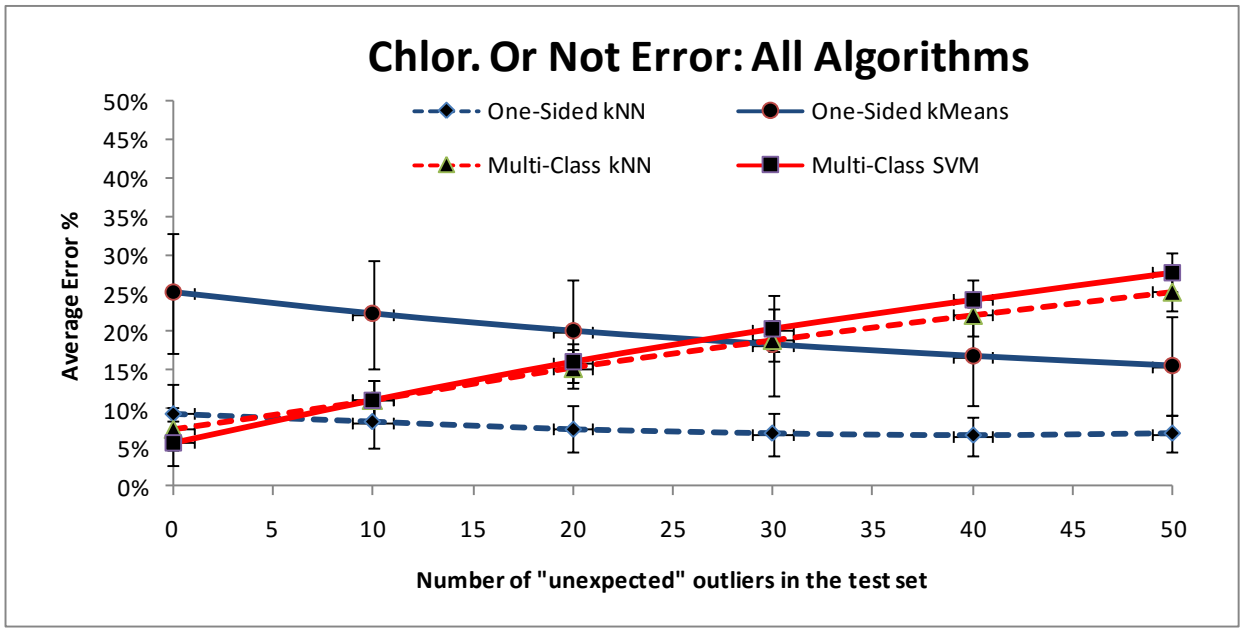

**Figure 35: Chlorinated or Not error graphed for all algorithms**

The One-Sided Support Vector Machine results have been omitted from these graphs due to its overall performance being poor, as discussed earlier.

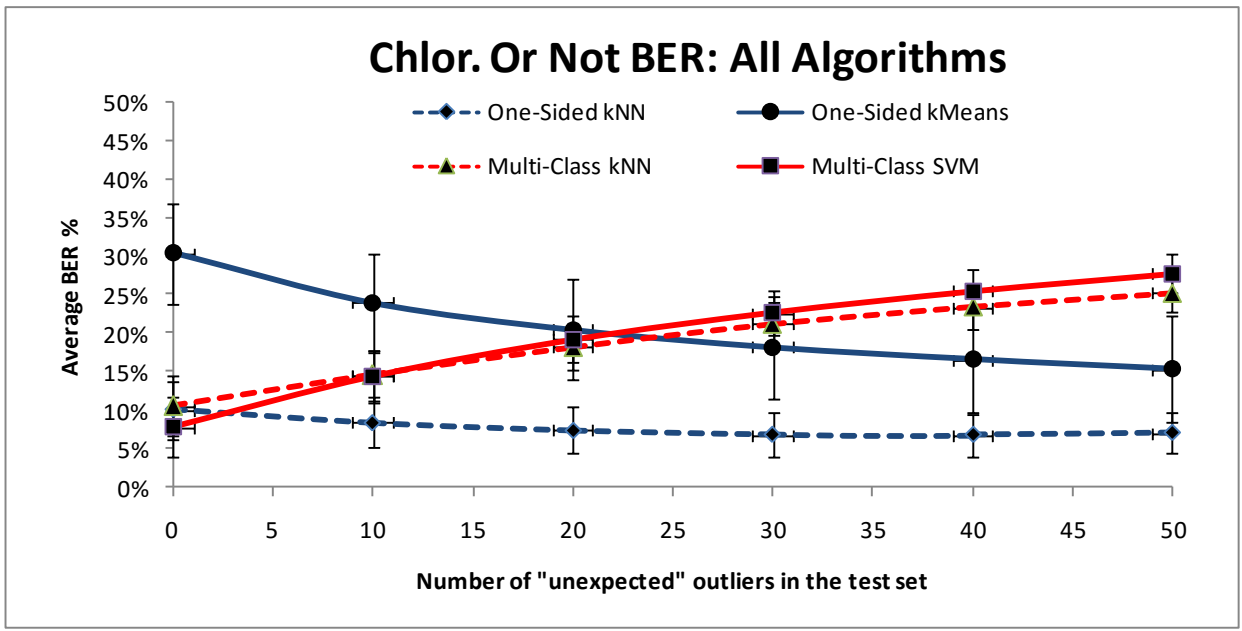

**Figure 36: Chlorinated or Not Balanced Error Rate graphed for algorithms**

From the graphs above, it can be seen that the One-Sided kNN algorithm starts to out-perform the multi-class approaches after just 5 outliers have been added. The k-Means algorithm begins to out-perform them when between 20 and 30 of these outliers are added. These results show the one-sided classifier's ability to robustly reject "unexpected" outliers. Also, based on the results from Chapter Four, it would be expected that these trends would continue in the same directions as more and more "unexpected" outliers were encountered in the test set.

# 6. Conclusions

This chapter will discuss the conclusions that were drawn from this research work. All of the findings will be first summarised and this will be followed by detailing the key research contributions that have taken place. The directions for future work, based on the overall findings, will then be discussed. This chapter, and the thesis as a whole, will then be brought to a close with some concluding remarks.

## 6.1 Work Summary

An extensive literature review was carried out at an early stage of this research. One-sided classification was described in detail and compared with the conventional multi-class approach to classification. Some drawbacks and pitfalls of multi-class classification were identified and a discussion of how one-sided classifiers could overcome these was provided. Next, a selection of one-sided algorithms, which were described in the literature, were presented and explained in detail. This was followed by a review of previous research that was carried out in the literature using one-sided classification in a variety of different application domains. This work summarised the diverse range of practical problems in which one-sided classification had shown promising results. A selection of existing software packages for classification were then identified and described. This was an important step to discover what software was currently available before the development of the OSCAIL toolkit began. The analytical chemistry technique of Raman spectroscopy was then introduced and some previous work involving Machine Learning and spectroscopy data was outlined. It was concluded that while multi-class classification techniques had been used with Ramon spectroscopy data in the past, the use of one-sided algorithms with such data had yet to be explored.

When the literature review had been completed, the design phase of the OSCAIL toolkit began. Once the design had been agreed upon by all project members, coding of the implementation then commenced. The early development of the toolkit involved a simple layout with just a single one-sided algorithm, the One-Sided k-Nearest Neighbour. As the development continued, the toolkit was improved and refined with further algorithms being added. All of the work that was carried out for the OSCAIL toolkit is fully described in Chapter Three.

## 6.2 Key Research Contributions

While the design and development of the OSCAIL toolkit could be regarded as a research contribution, it is through its application to novel experimentation, using Raman spectroscopy data that is most noteworthy. Once the development of the toolkit had been completed, the focus changed from coding the software to using its functionality to carry out some original experimentation. It was decided that the toolkit would be used in the classification of Raman spectroscopy data. Promising results for the use of multi-class algorithms with such data had been reported from the literature and, to the best of this author's knowledge, no one-sided classification experiments using Raman spectroscopy data had been previously published. The early results, which were the initial experiments that led to the results in Chapter Five of this thesis, were published in a peer-reviewed conference.

Another key contribution of this research has been to consider scenarios that involve out-of-sample data and analyse the effect of this data on both one-sided and multi-class classification algorithms. Although there are a variety of domains where such data would be commonplace in practical deployment of the classifiers, multi-class algorithms still tend to be used although they may not be the most suitable from a theoretical point of view. These experiments were set up to test, what we introduced and defined as "unexpected" outliers in the test set. This term was coined to represent the out-of-sample cases that we believe will be inevitably met in practice when the negative concept cannot be fully characterised in the training set. This has been fully explained in Chapter Four.

## 6.3  Directions for Future Work

There are several directions in which future work could be carried out. These include:

- **Improving the toolkit:** The toolkit could be upgraded to include a larger variety of one-sided algorithms and pre-processing techniques. Further upgrades, as explained in Section 3.6, could include developing a user-friendly graphical interface and using the toolkits Application Programming Interface (API) to extend existing software packages thus enabling them to carry out one-sided experiments.

- **Further analysis of Ramon spectroscopy data:** Considering this thesis deals primarily with a single application of Ramon spectroscopy data (chlorinated and non-chlorinated solvents), it would be important to extend these experiments by obtaining the Raman spectra for other compounds and mixtures to ensure that the same underlying trends in performance can be seen.

- **Further "Unexpected" outlier experiments:** To consolidate the findings in this thesis, other algorithms and datasets should be used in further experimentation. Some of the practical examples that have been discussed in Section 4.2.3 could be a good starting point.

- **Investigate poor One-Sided Support Vector Machine results:** The surprisingly poor results that were achieved for the One-Sided SVM in the "unexpected" outlier experiments are, in themselves, an interesting research area to delve into. It would be interesting to discover the cause of such results. Possible reasons could include an implementation error, this particular data being unsuitable for One-Sided SVM or simply that this algorithm is susceptible to the same drawbacks as the multi-class approach in such circumstances.

## 6.4 Concluding Remarks

The repercussions that are caused by "unexpected" outliers are certainly a very interesting area of research. As a result of such scenarios not being explicitly considered in the past, the scope for further research is enormous. The results reported in this dissertation are promising in that they support the theoretical standpoint that for some real-world practical problems, one-sided classifiers will tend to be more robust than comparable multi-class classifiers in practice, as it is not always feasible to sufficiently characterise the outlier concept in the training set.

Now that a one-sided classification toolkit has been developed, it can be used and improved upon to extend this research.

# Appendix A: Additional Information

## A.1 Attribute Relation File Format (ARFF) files

ARFF files are ASCII text files that describe a list of instances that have related attributes. This type of file was created by the Machine Learning Project at the Department of Computer Science in the University of Waikato, for use with WEKA Machine Learning software[27]. An ARFF file is broken up into a header section and a data section. A small example, shown in Figure 37 below, explains how the elements of these sections are denoted.

```
Example1.arff
%This is a comment
@relation example1
@attribute height numeric
@attribute width numeric
@attribute class {standard, large}
@data
50, 20, standard
150, 70, large
         ↑    ↑
       height width
```

- The `%` sign is used to write comments in .arff files.
- `@relation <relation-name>` where <relation-name> is a string.
- `@attribute <attribute-name> <data-type>` where <data-type> can be: numeric, nominal-specification, string, date
- Nominal values are defined by providing a list of possible values: {nominal-name1, nominal-name2......}
  For example: `@attribute class {standard, large}` describes 2 possible class values.
- The `@data` declaration is a single line denoting the start of the data segment in the file.
- Each instance is represented on a single line, with carriage returns denoting the end of the instance.
- Attribute values for each instance are delimited by commas.
- They must appear in the order that they were declared in the header section. See: height, width above.

**Figure 37: The different components of an ARFF file[28]**

---

[27] http://www.cs.waikato.ac.nz/~ml/weka/arff.html

[28] This explanation was created for OSCAIL documentation based on information from the above link.

- 102 -

## A.2 Distance Metrics

A distance metric is a function which defines the distance between elements in a given set. There are many different distance metrics available. The following three are implemented in the OSCAIL toolkit. More information about these distance metrics can be found in Priddy and Keller (2005, Section 4.2.3).

### A.2.1 Euclidean Distance

The Euclidean distance is the "ordinary" distance between two points that could be measured with a ruler. It can be proven with a repeated application of the Pythagorean Theorem.

The Euclidean distance between points
$P = (p_1, p_2,....,p_n)$ and $Q = (q_1, q_2,....,q_n)$
in Euclidean space is defined as:

$$\sqrt{\sum_{i=1}^{N}(p_i - q_i)^2}$$

### A.2.2 Manhattan Distance

The Manhattan, also known as Taxicab, distance between two points is measured by calculating the sum of the absolute differences of their coordinates.

The Manhattan distance between points
$P_1$ with coordinates $(x_1, y_1)$ and $P_2$ with coordinates $(x_2, y_2)$
is $|x_1 - x_2| + |y_1 - y_2|$

### A.2.3 Cosine Similarity

Cosine similarity uses the cosine of the angle between two vectors to measure the similarity of them. Given two vectors with attributes A and B, the cosine similarity, $\theta$, is represented as follows:

$$Similarity = \cos(\theta) = \frac{A \bullet B}{\|A\|\|B\|}$$

## A.3 Cross Validation

In cross validation, a number of folds or partitions of the data is selected and each fold, in turn, is used for testing while the remainder is used for training. For example, if we choose five as the number of folds, the following procedure will occur:

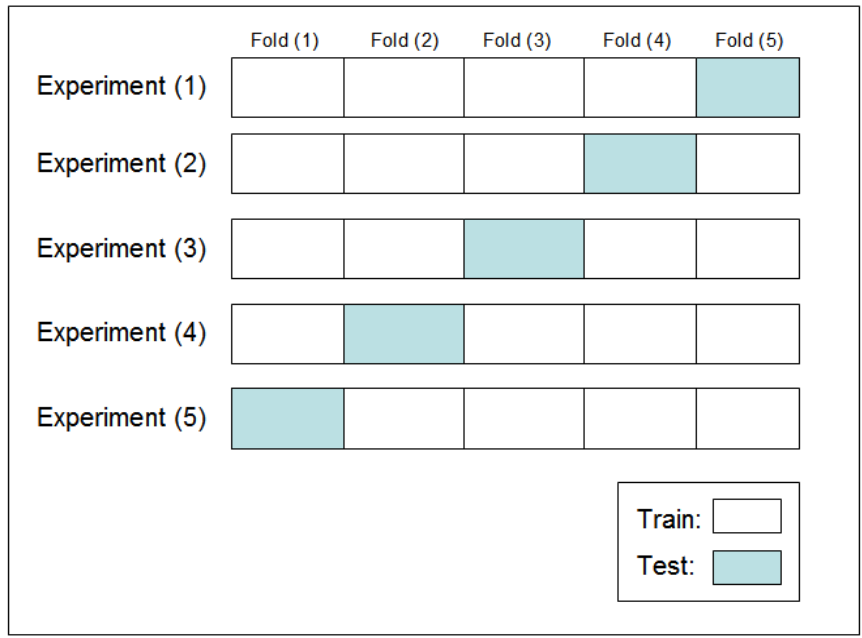

Figure 38: An example of 5-fold cross validation

This ensures that every instance in the dataset has been used exactly once for testing. The most common amount of folds to use in cross validation is ten. Based on extensive tests on a series of different datasets and learning techniques, using ten folds has shown to produce the best estimate of error. It is also a standard procedure to carry out this process ten times, giving *10 times 10-fold cross validation*. More information about cross validation can be found in Witten and Frank (2005, Section 5.3)

## A.4 Sensitivity and Specificity

In terms of classification, sensitivity refers to the proportion of positives that have been classified as positives, that is, the true positive rate. The specificity refers to the proportion of negatives which have actually been classified as negatives. Witten and Frank (2005, Section 5.7) provide an example of the use of the terms sensitivity and specificity in relation to medical diagnostic tests. The sensitivity corresponds to patients who have the disease and return a positive test result. The specificity corresponds to patients who do not have the disease and who return a negative test result after testing.

# Appendix B: Publications

# Analysis of the Effect of Unexpected Outliers in the Classification of Spectroscopy Data

Frank G. Glavin and Michael G. Madden

College of Engineering and Informatics,
National University of Ireland, Galway, Ireland.
`frank.glavin@gmail.com`, `michael.madden@nuigalway.ie`

**Abstract.** Multi-class classification algorithms are very widely used, but we argue that they are not always ideal from a theoretical perspective, because they assume all classes are characterized by the data, whereas in many applications, training data for some classes may be entirely absent, rare, or statistically unrepresentative. We evaluate one-sided classifiers as an alternative, since they assume that only one class (the target) is well characterized. We consider a task of identifying whether a substance contains a chlorinated solvent, based on its chemical spectrum. For this application, it is not really feasible to collect a statistically representative set of outliers, since that group may contain *anything* apart from the target chlorinated solvents. Using a new one-sided classification toolkit, we compare a One-Sided k-NN algorithm with two well-known binary classification algorithms, and conclude that the one-sided classifier is more robust to unexpected outliers.

**Key words:** One-Sided, One-Class, Classification, Support Vector Machine, k-Nearest Neighbour, Spectroscopy Analysis

## 1 Introduction

### 1.1 One-Sided Classification

One-sided classification (OSC) algorithms are an alternative to conventional multi-class classification algorithms. They are also referred to as single-class or one-class classification algorithms, and differ in one vital aspect from multi-class algorithms, in that they are only concerned with a single, well-characterized class, known as the target or positive class. Objects of this class are distinguished from all others, referred to as outliers, that consist of *all* the other objects that are not targets. In one-sided classification, training data for the outliers may be either rare, entirely unavailable or statistically unrepresentative.

Over the past decade, several well-known algorithms have been adapted to work with the one-sided paradigm. Tax [1] describes many of these one-sided algorithms and notes that the problem of one-sided classification is generally more difficult than that of conventional classification. The decision boundary in the multi-class case has the benefit of being well described from both sides with appropriate examples from each class being available, whereas the single-class case can only support one side of the decision boundary fully, in the absence of a comprehensive set of counter-examples. While multi-class (including binary or two-class) algorithms are very widely used in many different application domains, we argue that they are not always the best choice from a theoretical perspective, because they assume that all classes are appropriately characterized by the training data. We propose that one-sided classifiers are more appropriate in these cases, since they assume that only the target class is well characterized, and seek to distinguish it from any others. Such problem domains include industrial process control, document author identification and the analysis of chemical spectra.

### 1.2 Spectroscopic Analysis

Raman spectroscopy, which is a form of molecular spectroscopy, is used in physical and analytical chemistry. It involves the study of experimentally-obtained spectra by using an instrument such as a spectrometer [2]. According to Gardiner [3], Raman spectroscopy is a well-established spectroscopic technique which involves the study of vibrational and rotational frequencies in a system. Spectra are gathered by illuminating a laser beam onto a substance under analysis and are based on the vibrational motion of the molecules which create the equivalent of a chemical fingerprint. This unique pattern can then be used in the identification of a variety of different materials [4].

### 1.3 Machine Learning Task

In this work, we consider the task of identifying materials from their Raman spectra, through the application of both one-sided and multi-class classification algorithms. Our primary focus is to analyse the performance of the classifiers when "unexpected" outliers are added to the test sets. The spectra are gathered from materials in pure form and in mixtures. The goal is to identify the presence or absence of a particular material of interest from its spectrum. This task can be seen as an "open-ended" problem, as having a statistically representative set of counter-examples for training is not feasible, as has been discussed already.

In particular, we consider the application of separating materials to enable the safe disposal of harmful solvents. Chemical waste that is potentially hazardous to the environment should be identified and disposed of in the correct manner. Laboratories generally have strict guidelines in place, as well as following legal requirements, for such procedures. Organic solvents can create a major disposal problem in organic laboratories as they are usually water-immiscible and can be highly flammable [5]. Such solvents are generally created in abundance each day in busy laboratories. Differentiating between chlorinated and non-chlorinated organic solvents is of particular importance. Depending on whether a solvent is chlorinated or not will dictate how it is transported from the laboratory and, more importantly, what method is used for its disposal [6]. Identifying and labeling such solvents is a routine laboratory procedure which usually makes the disposal a straightforward process. However, it is not unlikely that the solvents could be accidentally contaminated or inadvertently mislabeled. In such circumstances it would be beneficial to have an analysis method that would correctly identify whether or not a particular solvent was chlorinated.

We have carried out several experiments for this identification using both one-sided and multi-class classification algorithms in order to analyse the effect of adding "unexpected" outliers to the test sets.

## 2   Related Research

Madden and Ryder [7] explore the use of standard multi-class classification techniques, in comparison to statistical regression methods, for identifying and quantifying illicit materials using Raman spectroscopy. Their research involves using dimension reduction techniques to select some features of the spectral data and discard all others. This feature selection process is performed by using a Genetic Algorithm. The predictions can then be made based only on a small number of data points. The improvements that can be achieved by using several different predictor models together were also noted. This would come at the cost of increased computation but was shown to provide better results than using just one predictor by itself.

O'Connell *et al.* [8] propose the use of Principal Component Analysis (PCA), support vector machines (SVM) and Raman spectroscopy to identify an analyte[1] in solid mixtures. In this case, the analyte is acetaminophen, which is a pain reliever used for aches and fevers. They used near-infrared Raman spectroscopy to analyse a total of 217 samples, some of which had the target analyte present, of mixtures with excipients[2] of varying weight. The excipients that were included were sugars, inorganic materials and food products. The spectral data was subjected to first derivative and normalization transformations in order to make it more suitable for analysis. After this pre-treatment, the target analyte was then discriminated using Principal Component Analysis (PCA), Principal Component Regression (PCR) and Support Vector Machines. According to the authors, the superior performance of SVM was particularly evident when raw data was used for the input. The importance and benefits of the pre-processing techniques was also emphasized.

Howley [9] uses machine learning techniques for the identification and quantification of materials from their corresponding spectral data. He shows how using Principal Component Analysis (PCA) with machine learning methods, such as SVM, could produce better results than the chemometric technique of Principal Component Regression (PCR). He also presents customized kernels for use with spectral analysis based on prior knowledge of the domain. A genetic programming technique for evolving kernels is also proposed for when no domain knowledge is available.

---

[1] An analyte is a substance or chemical constituent that is determined in an analytical procedure.
[2] An excipient is an inactive substance used as a carrier for the active ingredients of a medication.

## 3 A Toolkit for One-sided Classification

In the course of our research, we have developed a one-sided classification toolkit written in Java. It is a command line interface (CLI) driven software package that contains one-sided algorithms that may be chosen by the user at runtime and used to create a new classifier based on a loaded data set and a variety of different adjustable options. Both experiment-specific and classifier parameter options can be set. The toolkit was designed to carry out comprehensive and iterative experiments with minimal input from the user. The resulting classifiers that are generated can be saved and used at a later stage to classify new examples. The user can set up many different runs of an experiment, each differing by an incremented random number seed that shuffles the data for every run before it is broken up into training and testing sets. Results are printed to the screen as they are calculated; these include the classification error, sensitivity, specificity and confusion matrix for each run or individual folds.

## 4 Data Sets and Algorithms Used

### 4.1 Primary Data Set

The primary data set that we used for these experiments was compiled in earlier research, as described by Conroy *et al.* [6]. It comprises of 230 spectral samples that contain both chlorinated and non-chlorinated mixtures. According to the authors, the compilation of the data involved keeping the concentrations of the mixtures as close as possible to real life scenarios from industrial laboratories. Twenty five solvents, some chlorinated and some not, were included; these are listed in Table 1.

Several variants of the data set were created, which differed only by the labeling of the solvent that was currently assigned as the target class. In each of these variants, all instances not labeled as targets were labeled as outliers. These relabeled data sets were used in the detection of the specific chlorinated solvents: Chloroform, Dichloromethane and Trichloroethane. As an example of the data, the Raman spectrum of pure Chloroform, a chlorinated solvent, is shown in Fig. 1. Other samples from the data set consist of several different solvents in a mixture which makes the classification task more challenging. A final separate data set was created such that all of the chlorinated solvents were labeled as targets. This is for carrying out experiments to simply detect whether a given mixture is chlorinated or not.

### 4.2 Secondary Data Set

For our Scenario 2 experiments (see Section 5.1), we introduce 48 additional spectra that represent outliers that are taken from a different distribution to those that are present in the primary dataset. These samples are the Raman spectra of various laboratory chemicals, and none of them are chlorinated solvents nor are they the other materials that are listed in Table 1. They include materials such as sugars, salts and acids in solid or liquid state, including Sucrose, Sodium, Sorbitol, Sodium Chloride, Pimelic Acid, Acetic Acid, Phthalic Acid and Quinine.

Table 1. A list of the various chlorinated and non-chlorinated solvents used in the primary data set and their grades.(Source: Conroy *et al.* [6])

| Solvent | Grade | Solvent | Grade |
|---|---|---|---|
| Acetone | HPLC | Cyclopentane | Analytical |
| Toluene | Spectroscopic | Acetophenol | Analytical |
| Cyclohexane | Analytical & Spect. | n-Pentane | Analytical |
| Acetonitrile | Spectroscopic | Xylene | Analytical |
| 2-Propanol | Spectroscopic | Dimethylformanide | Analytical |
| 1,4-Dioxane | Analytical & Spect. | Nitrobenzene | Analytical |
| Hexane | Spectroscopic | Tetrahydrofuran | Analytical |
| 1-Butanol | Analytical & Spect. | Diethyl Ether | Analytical |
| Methyl Alcohol | Analytical | Petroleum Acetate | Analytical |
| Benzene | Analytical | Chloroform | Analytical & Spect. |
| Ethyl Acetate | Analytical | Dichloromethane | Analytical & Spect. |
| Ethanol | Analytical | 1,1,1-trichloroethane | Analytical & Spect. |

Table 2. Summary of chlorinated and non-chlorinated mixtures in the primary data set. (Source: Howley [9])

| | Chlorinated | Non-chlorinated | Total |
|---|---|---|---|
| Pure Solvents | 6 | 24 | 30 |
| Binary Mixtures | 96 | 23 | 119 |
| Ternary Mixtures | 40 | 12 | 52 |
| Quaternary Mixtures | 12 | 10 | 22 |
| Quintary Mixtures | 0 | 7 | 7 |
| Total | 154 | 76 | 230 |

### 4.3  Algorithms Used

We carried out the one-sided classification experiments using our toolkit. The conventional classification experiments were carried out using the Weka [10] machine learning software.

We have chosen a One-Sided k-Nearest Neighbour (OkNN) algorithm and two conventional classification algorithms; namely, k-Nearest Neighbour, that we refer to as Two-Class KNN, and a Support Vector Machine (SVM) that we will refer to as Two-Class SVM.

The OkNN algorithm we use is based on one described by Munroe and Madden [11]. The method involves choosing an appropriate threshold and number of neighbours to use. The average distance from a test example 'A' to its m nearest neighbours is found and this is called 'D1'. Then, the average distance of these neighbours to their own respective k nearest neighbours is found and called 'D2'. If 'D1' divided by 'D2' is greater than the threshold value, the test example 'A' is rejected as being an outlier. If it is less than the threshold, then it is accepted as being part of the target class.

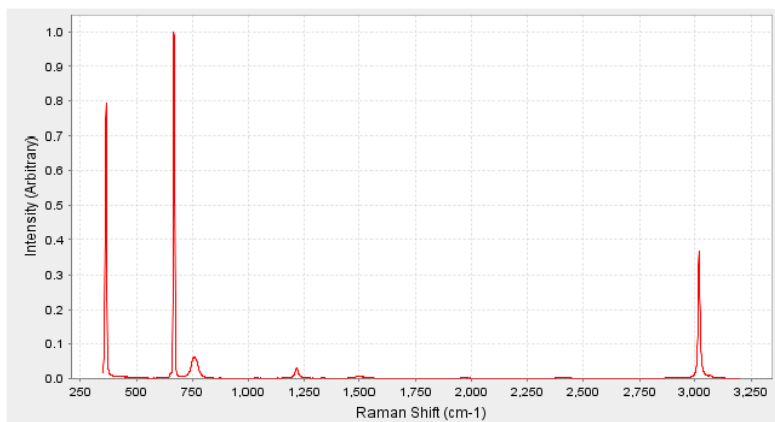

**Fig. 1.** *The Raman Spectrum of a sample of 100% pure Chloroform*

## 5 Description of Experiments

### 5.1 Scenarios Considered

Two scenarios are described in our experiments, as described next.

**Scenario 1: "Expected" Test Data Only.** In this scenario, the test data is sampled from the same distribution as the training data. The primary dataset is divided repeatedly into training sets and test sets, with the proportions of targets and outliers held constant at all times, and these internal test sets are used to test the classifiers that are built with the training datasets.

**Scenario 2: "Unexpected" and "Expected" Test Data.** In this scenario, we augment each test dataset with the 48 examples from the secondary data set that are *not* drawn from the same distribution as the training dataset. Therefore, a classifier trained to recognise any chlorinated solvent should reject them as outliers. However, these samples represent a significant challenge to the classifiers, since they violate the standard assumption that the test data will be drawn from the same distribution as the training data; it is for this reason that we term them "unexpected".

This second scenario is designed to assess the robustness of the classifiers in a situation that has been discussed earlier, whereby in practical deployments of classifiers in many situations, the classifiers are likely to be exposed to outliers that are not drawn from the same distribution as training outliers. In fact, we contend that over the long term, this is inevitable: if we know *a priori* that the outlier class distribution is not well characterized in the training data, then we must accept that sooner or later, the classifier will be exposed to data that falls outside the distribution of the outlier training data.

It should be noted that this is different from *concept drift*, where a target concept may change over time; here, we have a static concept, but over time the weaknesses of the training data are exposed. Of course, re-training might be possible, if problem cases can be identified and labeled correctly, but we concern ourselves with classifiers that have to maintain robust performance without re-training.

### 5.2 Experimental Procedure

The data sets, as described earlier, were used to test the ability of each algorithm in detecting the individual chlorinated compounds. This involved three separate experiments for each algorithm, to detect Chloroform, Dichloromethane and Trichloroethane. A fourth experiment involved detecting the presence of any chlorinated compound in the mixture.

All spectra were first normalized. A common method for normalizing a dataset is to recalculate the values of the attributes to fall within the range of zero to one. This is usually carried out on an attribute-by-attribute basis and ensures that certain attribute values, which differ radically in size from the rest, don't dominate in the prediction calculations. The normalization carried out on the spectral data is different to this in that it is carried out on an instance-by-instance basis. Since each attribute in an instance is a point on the spectrum, this process is essentially rescaling the height of the spectrum into the range of zero to one.

For each experiment, 10 runs were carried out with the data being randomly split each time into 67% for training and 33% for testing. The splitting procedure from our toolkit ensured that there was the same proportion of targets and outliers in the training sets as there was in the test sets. The same data set splits were used for the one-sided classifier algorithms and the Weka-based algorithms, to facilitate direct comparisons.

A 3-fold internal cross validation step was used with all the training sets, to carry out parameter tuning. A list of parameter values was passed to each classification algorithm and each, in turn, was used on the training sets, in order to find the best combination that produced the smallest error estimate. It must be emphasized that we only supplied a small amount of different parameters for each algorithm and that these parameters used were the same for all of the four variants of the data set. The reason for this was that our goal was not to tune and identify the classifier with the best results overall but to notice the change in performance when "unexpected" outliers were added to the test set.

For the One-Sided kNN algorithm, the amount of nearest neighbours (m) and the amount of their nearest neighbours (k) was varied between 1 and 2. The threshold values tried were 1, 1.5 and 2. The distance metric used was Cosine

Similarity. For the Weka experiments, the Two-Class kNN approach tried the values 1,2 and 3 for the amount of nearest neighbours. The Two-Class SVM varied the complexity parameter C with the values of 1,3 and 5. The default values were used for all of the other Weka parameters.

### 5.3   Performance Metric

The error rate of a classification algorithm is the percentage of examples from the test set that are incorrectly classified. We measure the average error rate of each algorithm over the 10 runs to give an overall error estimate of its performance. With such a performance measure being used, it is important to know what percentage of target examples were present in each variant of the data set. This information is listed in Table 3 below.

Table 3. Percentage of target examples in each variant of the *primary* data set

| Dataset | Targets | "Expected" Outliers | Target Percent |
|---|---|---|---|
| Chlorinated or not | 154 | 76 | 66.95% |
| Chloroform | 79 | 151 | 34.34% |
| Dichloromethane | 60 | 170 | 26.08% |
| Trichloroethane | 79 | 151 | 34.34% |

## 6   Results and Analysis

The results of the experiments carried out are listed in Table 4, Table 5, and Table 6 below. Each table shows the overall classification error rate and standard deviation (computed over 10 runs) for each algorithm, for both of the scenarios that were tested.

It can be seen that while the conventional multi-class classifiers perform quite well in the first scenario, their performance quickly begins to deteriorate once the "unexpected" outliers are introduced in Scenario 2. The One-Sided kNN's performance is generally worse than the multi-class approach in Scenario 1. As described in Section 1.1, the decision boundary for the multi-class classifiers have the benefit of being well supported from both sides with representative training examples from each class. In such a scenario, the multi-class algorithms essentially have more information to aid the classification mechanism and, therefore, would be expected to out-perform the one-sided approach.

In detecting whether or not a sample is chlorinated, the average error rate of the Two-Class kNN increased by 28.87% and the Two-Class SVM increased by 33.57% in Scenario 2. In contrast with the two-class classifiers, the One-Sided kNN is seen to retain a consistent performance and the error is only increased by 0.14%. When the algorithms are detecting the individual chlorinated solvents, the same pattern in performance can be seen. The multi-class algorithms' error rates increase, in some cases quite radically, in the second scenario. The One-Sided kNN manages to remain at a more consistent error rate and, in the case of Chloroform and Dichloromethane, the overall error rate is reduced somewhat.

It should be noted that our experiments are not concerned with comparing the relative performances of a one-sided classifier versus the multi-class classifiers. Rather, we analyse the variance between the two scenarios for each individual classifier and demonstrate the short-comings of the multi-class approach when it is presented with "unexpected" outliers. Our results demonstrate the one-sided classifier's ability to robustly reject these outliers in the same circumstances.

Table 4. Overall average error rate for two-class kNN in both scenarios

| Two-Class kNN | Scenario 1. | Scenario 2. |
| --- | --- | --- |
|  | Error % (std. dev.) | Error % (std. dev.) |
| Chlorinated or not | 6.49 (2.03) | 35.36 (3.65) |
| Chloroform | 22.59 (6.93) | 39.44 (7.37) |
| Dichloromethane | 11.94 (4.89) | 16.24 (3.49) |
| Trichloroethane | 23.24 (5.10) | 25.68 (4.27) |

Table 5. Overall average error rate for two-class SVM in both scenarios

| Two-Class SVM | Scenario 1. | Scenario 2. |
| --- | --- | --- |
|  | Error % (std. dev.) | Error % (std. dev.) |
| Chlorinated or not | 4.67 (1.95) | 38.24 (2.19) |
| Chloroform | 11.68 (4.01) | 37.2 (2.39) |
| Dichloromethane | 8.70 (4.37) | 11.68 (3.52) |
| Trichloroethane | 11.03 (3.47) | 30.08 (2.50) |

Table 6. Overall average error rate for one-sided kNN in both scenarios

| One-Sided kNN | Scenario 1. | Scenario 2. |
| --- | --- | --- |
|  | Error % (std. dev.) | Error % (std. dev.) |
| Chlorinated or not | 10.90 (4.5) | 11.04 (4.7) |
| Chloroform | 26.10 (3.43) | 18.32 (3.16) |
| Dichloromethane | 12.98 (3.23) | 9.36 (2.84) |
| Trichloroethane | 20.77 (3.46) | 21.04 (5.07) |

## 7  Conclusions and Future Work

Our research demonstrates the potential drawbacks of using conventional multi-class classification algorithms when the test data is taken from a different distribution to that of the training samples. We believe that for a large number of real-world practical problems, one-sided classifiers should be more robust than multi-class classifiers, as it is not feasible to sufficiently characterize the outlier concept in the training set. We have introduced the term "unexpected outliers" to signify outliers that violate the standard underlying assumption made by multi-class classifiers, which is that the test set instances are sampled from the same distribution as the training set instances. We have shown that, in such circumstances, a one-sided classifier can prove to be a more capable and robust alternative. Our future work will introduce new datasets from different domains and also analyse other one-sided and multi-class algorithms.

**Acknowledgments.** The authors are grateful for the support of Enterprise Ireland under Project CFTD/05/222a. The authors would also like to thank Dr. Abdenour Bounsiar for his help and valuable discussions, and Analyze IQ Limited for supplying some of the Raman spectral data.